%% file: main.tex
\documentclass{article}

\usepackage{hello-paper}

\usepackage[utf8]{inputenc}
\usepackage[T1]{fontenc}
\usepackage[square,numbers,sort&compress]{natbib}
\usepackage{hyperref}
\hypersetup{hidelinks}
\usepackage{url}
\usepackage{booktabs}
\usepackage{amsmath,amssymb}
\usepackage{nicefrac}
\usepackage{microtype}
\usepackage{graphicx}
\usepackage{tikz}
\usepackage{float}
\usepackage{placeins}
\usepackage{multirow}
\usepackage{makecell}
\usepackage{xcolor}
\usepackage{enumitem}
\usepackage{caption}
\usepackage{array}
\usepackage{tabularx}
\usepackage{pifont}
\usepackage{cleveref}
\usepackage{colortbl}
\usepackage{subcaption}
\newcommand{\tbd}[1]{xx}

\newcommand{\cmark}{\ding{51}}
\newcommand{\xmark}{\ding{55}}
\newcommand{\pmark}{$\triangle$}

\newcommand{\SevenViewFlatten}{7-View Flatten}
\newcommand{\MetricScaleFlag}{Metric Scale Flag}
\newcommand{\ZeroInitTransfer}{Zero-Init Cross-View Transfer}
\newcommand{\SelfRollout}{Self-Rollout Alignment}

\title{HelloWorld: Towards Practical Applications of Generative Driving World Models}

\author{
  HelloWorld Team
}

\date{}

\begin{document}
\maketitle

\begin{abstract}
Driving world models provide a promising route toward scalable counterfactual data generation and interactive simulation beyond recorded driving logs. Realizing this potential requires a system that can generalize across diverse scenes, respond faithfully to prescribed controls, generate coherent multi-sensor observations, and operate efficiently under repeated inference. We present \textbf{HelloWorld}, a 2B driving world model system designed around these requirements. HelloWorld progressively specializes broad visual and motion priors from heterogeneous video data into controllable driving generation using ego pose, HD maps, and 3D boxes. A block-causal generation interface, together with adaptation to self-generated context, aligns the model with sequential simulation. The system further supports synchronized seven-camera RGB generation and conditional LiDAR synthesis, and is distilled toward few-step inference for efficient deployment. Experiments evaluate visual quality, control fidelity, cross-view consistency, robustness under repeated generation, inference efficiency, and LiDAR synthesis. Together, HelloWorld provides a unified framework for scalable driving data generation and interactive simulation.

\textbf{Website:} \href{https://helloworld-4d.github.io}{\textcolor{blue}{https://helloworld-4d.github.io}}
\end{abstract}


\input{sections/overview}
\input{sections/related}
\input{sections/data}
\input{sections/method}
\input{sections/experiments}
\input{sections/applications}
\input{sections/limitations}
\input{sections/conclusion}

\bibliographystyle{unsrtnat}
\bibliography{refs}

\appendix
\input{sections/authors}

\end{document}

%% file: sections/overview.tex
\section{Introduction}
\label{sec:overview}

Driving logs provide rich observations of the physical world, but they are inherently limited to trajectories and traffic configurations that were actually encountered. They cannot directly reveal how the same scene would evolve under a different ego trajectory, how observations would change under an alternative traffic arrangement, or what rare situations might look like before sufficient real-world examples have been collected. Generative driving world models offer a complementary capability: they can synthesize counterfactual observations beyond recorded data for scalable data generation and, more broadly, provide generative environments for interactive driving simulation. Realizing this potential requires moving beyond isolated video generation toward a system that can generalize across diverse scenes, respond faithfully to prescribed controls, produce coherent observations of the evolving world, and operate efficiently enough for repeated interaction.

Turning such a generative model into a practical driving world model is therefore not a matter of improving visual fidelity alone. The model must represent sufficiently broad driving scenarios, expose interfaces through which the simulated world can be intentionally altered, and support repeated generation under the information flow available during simulation. It must also produce the observations required by downstream driving systems at a tractable computational cost. We view these requirements jointly as a system-level problem, encompassing \textit{generalization}, \textit{controllability}, \textit{multi-sensor observation generation}, and \textit{efficient inference}, with temporal causality providing the generation interface that connects them to sequential simulation.

To address this problem, we present \textbf{HelloWorld}, a 2B driving world model system for scalable data generation and interactive simulation. HelloWorld is built around a progressive learning strategy that first acquires broad visual and motion priors from heterogeneous video data and then specializes these priors into controllable driving dynamics using increasingly structured supervision. Explicit trajectory and scene conditions allow the model to synthesize observations beyond recorded driving logs, while causal generation and adaptation to self-generated context align its temporal behavior with sequential simulation. The framework further extends the generated world state across synchronized camera views and geometric sensing, and employs few-step generation to reduce the cost of repeated inference. Together, these designs turn a general video prior into a controllable and efficient generative model of driving observations.

\begin{figure}[t]
\includegraphics[width=\linewidth]{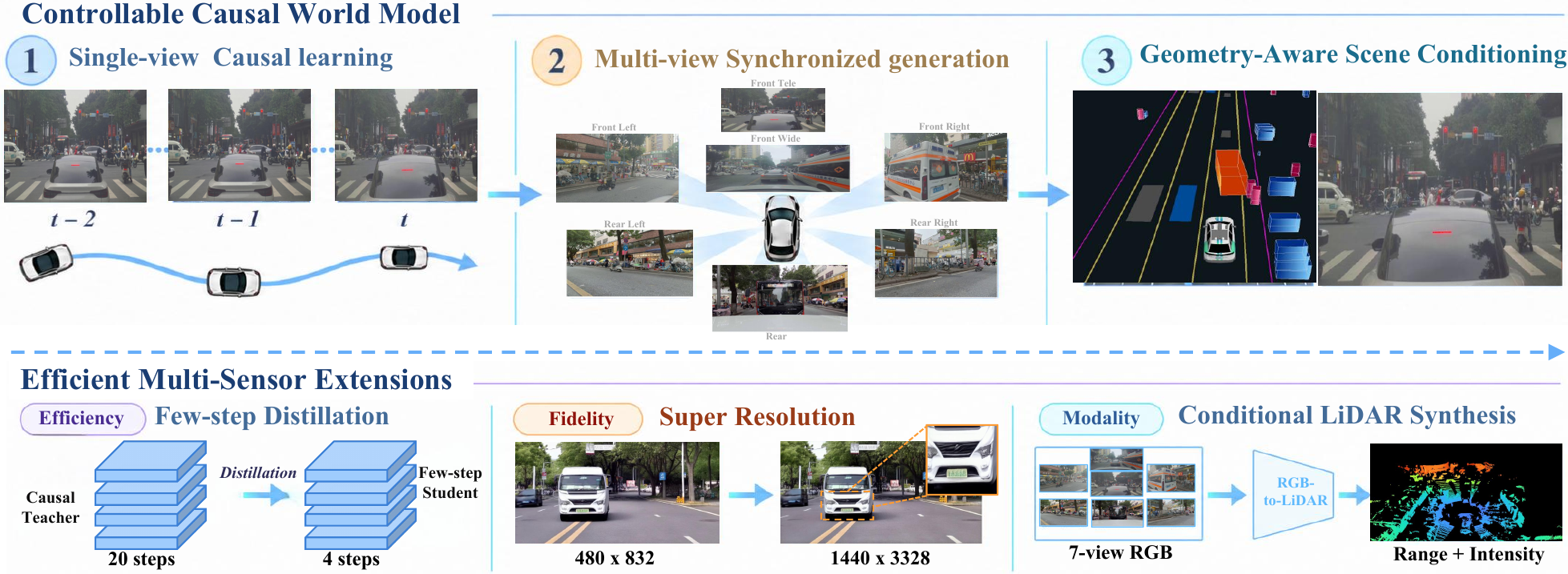}
\caption{\textbf{Overview of HelloWorld.} \textit{Top:} A block-causal RGB generator is progressively specialized from pose-conditioned single-view modeling to synchronized seven-view synthesis and geometry-aware scene conditioning with ego pose, HD maps, and 3D boxes. The causal observation interface is retained across all stages. \textit{Bottom:} A 20-step causal teacher is distilled into a four-step student, a decoupled super-resolution module enhances spatial fidelity, and an RGB-conditioned LiDAR branch synthesizes range and return intensity from synchronized seven-view observations.}
\label{fig:overview}
\end{figure}

A central challenge is to combine broad visual knowledge with the structured supervision required for controllable driving generation. Training only on richly annotated driving data limits scene diversity, whereas general video provides much broader appearance, motion, and interaction patterns but lacks the geometric and semantic interfaces needed for driving simulation. HelloWorld therefore adopts a progressive specialization strategy: it first learns pose-conditioned visual dynamics from heterogeneous general and fleet video, while accounting for their different pose-scale conventions, and subsequently introduces increasingly structured driving supervision. Calibrated multi-view data establish consistent surround observations, while HD maps and 3D boxes provide explicit scene-level conditions. Together with ego-pose control, these structured signals allow the model to move beyond replaying recorded trajectories and generate counterfactual observations under altered ego motion and scene configurations. This progression forms the upper path of Figure~\ref{fig:overview}, while the lower path connects the resulting RGB model to few-step generation and conditional LiDAR synthesis.

Once the model can be actively controlled, its temporal formulation must also match the way it is used during simulation. In an interactive setting, future observations are unavailable: each new observation must be generated from the history produced so far together with the supplied conditions. We therefore adopt block-causal observation modeling from the beginning of driving adaptation, generating frames jointly within each block while restricting later blocks to previously available observations. This causal interface aligns the model with sequential simulation, but it does not eliminate the discrepancy between training on recorded histories and inference on model-generated ones. We therefore further expose the model to imperfect context through context corruption and self rollout. These mechanisms improve robustness to the self-generated histories encountered during repeated simulation.

With the spatial controls and temporal interface established, HelloWorld further broadens the observation space produced at each simulation step. Calibrated cross-view communication enables synchronized seven-camera generation that reflects a shared scene state across viewpoints, while a conditional LiDAR branch extends the system from appearance synthesis to geometric sensing. The LiDAR component uses geometry-aware tokenization and cross-modal interaction with RGB features to preserve valid returns and metric structure without forcing the two modalities into a shared native representation. Finally, because both large-scale data synthesis and interactive simulation repeatedly invoke the generative model, inference cost becomes part of the system design rather than a standalone acceleration problem. We therefore distill HelloWorld toward few-step generation using consistency training and distribution matching, reducing repeated sampling cost while preserving the same control and sequential-generation interfaces.

Taken together, these components form a unified driving world model system that connects broad data priors, explicit control, sequential generation, multi-sensor observation synthesis, and efficient inference. Rather than treating these capabilities as independent extensions, HelloWorld organizes them around a common objective: generating controllable driving observations beyond recorded trajectories and scene configurations, while retaining the temporal and computational properties required for repeated simulation.

Our contributions are summarized as follows:
\begin{itemize}[leftmargin=2em]
\item We present \textbf{HelloWorld}, a scalable and controllable driving world model system for data generation and interactive simulation, integrating broad video priors with structured driving supervision, synchronized multi-view generation, multi-sensor observation synthesis, and efficient inference.

\item We develop a progressive specialization strategy that transfers general visual and motion knowledge into controllable driving generation. Ego pose, HD maps, and 3D boxes provide complementary interfaces for modifying camera motion and scene configuration, enabling counterfactual synthesis beyond recorded driving logs.

\item We introduce a temporally causal generation interface together with adaptation to self-generated context, reducing the gap between offline training and sequential simulation.

\item We extend the system to synchronized seven-camera and conditional LiDAR generation, and further distill the generative process to few-step inference using consistency training and distribution matching, improving its suitability for repeated data synthesis and interactive use.

\end{itemize}

%% file: sections/related.tex
\section{Related Work}
\label{sec:related}

\subsection{Video Foundation and Interactive World Models}

Large-scale video foundation models established the architectural and data-scaling basis for modern world generation. Wan~\cite{wan2025} combines a causal video autoencoder, a diffusion-transformer backbone, flow matching, large-scale image--video curation, and task-specific post-training in an open model suite. Cosmos~\cite{nvidia2025cosmos} extends the foundation-model view toward physical AI, organizing tokenization, pretraining, conditional post-training, deployment, and safety tooling as one platform. These systems show that reusable visual priors can support many downstream generators more efficiently than training each domain from scratch. HelloWorld follows this foundation-to-specialization principle: its main 2B configuration starts from Cosmos-Predict2.5, but introduces a curriculum designed around the supervision and camera geometry of autonomous-driving data rather than immediately fitting a driving-only adapter.

A related line turns video models into interactive environments controlled by camera motion or user actions. Genie~3~\cite{genie3} emphasizes promptable real-time worlds, while HY-World~1.5 / WorldPlay~\cite{hyworld2025} describes a pipeline spanning pretraining, interaction-oriented post-training, and streaming distillation. LingBot-World~\cite{lingbotworld2026} and Matrix-Game~2.0~\cite{matrixgame2} emphasize long-horizon interaction and open deployment. Publicly described systems in this family mainly expose a first-person visual stream and scale-free motion or learned action spaces. They are not designed around the fixed extrinsics, heterogeneous fields of view, and simultaneous outputs of a production vehicle rig. HelloWorld uses these broad-domain data priors in Stage~1, then explicitly transfers them to metric pose control and synchronized seven-view generation. We distinguish interactive observation generation from a validated driving closed loop: the latter additionally requires action-to-dynamics propagation, reactive traffic, and policy-in-the-loop metrics.

\subsection{Driving Video Generation and Conditioning}

Controllable driving generation has developed along two overlapping routes. \emph{Layout-driven} methods condition synthesis on BEV sketches, HD maps, occupancy, or projected 3D boxes. BEVGen~\cite{swerdlow2024bevgen} and BEVControl~\cite{yang2023bevcontrol} establish layout-to-street-view generation with multi-perspective constraints. DriveDreamer~\cite{wang2023drivedreamer}, MagicDrive~\cite{gao2023magicdrive}, Panacea~\cite{wen2024panacea}, and UniScene~\cite{li2024uniscene} extend structured conditioning to driving video and multi-view settings. MagicDrive-V2~\cite{gao2024magicdrivev2} combines a multi-view DiT block with spatial-temporal encoders for camera, trajectory, map, and box conditions, and uses progressive training to increase resolution and duration. DiVE~\cite{jiang2024dive} likewise studies stronger control within a DiT video generator. Cosmos-Transfer~\cite{nvidia2025cosmostransfer} generalizes this idea to adaptive combinations of control modalities, including depth, segmentation, LiDAR, and HD maps.

Layout conditioning is effective because it supplies dense geometric evidence at every frame. It also creates an important modeling choice: when layouts are reprojected under the target camera trajectory, ego motion is partly represented by the layout sequence itself. Such a model can edit geometry precisely, but using it for pose-only pretraining requires dense annotations, and independently changing ``where the vehicle moves'' and ``what exists in the scene'' becomes less direct. HelloWorld retains dense layout control while assigning ego motion to a separate metric 6-DoF Plücker pathway. The distinction is architectural rather than semantic: pose and layout may describe the same future, but neither must pass through the other's encoder.

\emph{Action-driven and predictive} models expose motion more directly. GAIA-1~\cite{hu2023gaia1} formulates driving as autoregressive token prediction conditioned on text and vehicle actions. GAIA-2~\cite{russell2025gaia2} moves to latent flow matching, generates consistent multi-camera video, accepts ego dynamics, agent state, road semantics, and scene metadata, and supports sliding-window autoregressive prediction from previous latent context. Vista~\cite{gao2024vista} studies a generalizable single-view world model with several driving controls. GenAD~\cite{yang2024genad} learns a generalized predictive representation. DrivingWorld~\cite{hu2024drivingworld} returns to a Video-GPT formulation, and Drive-WM~\cite{wang2024drivewm} connects multi-view visual forecasting to planning. These works demonstrate that explicit actions and multi-view generation are compatible. HelloWorld's narrower distinction is the simultaneous use of dense metric camera pose, a separately switchable layout tower, a calibrated seven-camera rig, and one block-causal interface across all training stages.

\subsection{Multi-View Consistency and Single-to-Multi-View Transfer}

Synchronized driving cameras introduce constraints absent from ordinary video generation: views have fixed but different intrinsics and extrinsics, only selected pairs overlap, and an object may leave one view while entering another. Existing systems address these constraints through joint generation, camera-aware embeddings, shared BEV or projected conditions, and explicit cross-view attention~\cite{gao2023magicdrive,wen2024panacea,gao2024magicdrivev2,russell2025gaia2}. These mechanisms improve consistency, but most methods train the multi-view generator as a target architecture from the beginning or initialize it from a generic video model before driving-specific training. This makes synchronized data the main route by which both appearance and cross-view correspondence are learned.

HelloWorld separates those learning problems. Stage~1 learns appearance, motion, and pose response from the union of single-view internet video and flattened fleet cameras. Stage~2 then inserts adjacency-restricted cross-view attention as a zero-output residual. Copying its query, key, value, and normalization weights from self-attention provides a non-random projection, while the zero output projection makes the added residual contribution zero at initialization. The intended benefit is a controlled initialization for learning cross-view interactions from synchronized data. Zero initialization preserves the added branch's input at insertion. Whole-model parity and retention after full-model fine-tuning are separate empirical questions.

\subsection{Causal and Long-Horizon Video Generation}

Long-horizon generation is not synonymous with causality. A model may synthesize a long clip with bidirectional attention over the full sequence, recurrently extend a fixed denoising window, generate sparse anchors and interpolate between them, or maintain a causal state that can be updated online. These choices have different implications for latency, memory, and train--test mismatch. GAIA-1 and DrivingWorld use autoregressive token models~\cite{hu2023gaia1,hu2024drivingworld}. GAIA-2 predicts future latent windows from a sliding context~\cite{russell2025gaia2}, and MiLA~\cite{mila2025} combines low-frame-rate anchor generation with a coarse-to-refine process to produce videos up to one minute. Recent driving work continues to investigate causal rollout and temporal error correction~\cite{horizondrive2026}.

For diffusion models, a central difficulty is the distribution of the history. Teacher-forced training observes real or clean previous frames, whereas autoregressive inference conditions on samples containing the model's own errors. Diffusion Forcing~\cite{chen2024diffusionforcing} assigns noise levels across a sequence so generation and prediction tasks can share a diffusion formulation. Self Forcing~\cite{huang2025selfforcing} instead trains with self-generated autoregressive context to address exposure bias more directly. HelloWorld combines these ideas at chunk granularity: per-chunk noise schedules, context corruption, and \SelfRollout{} train the same block-causal network that is executed sequentially with a KV cache. Unlike overlap-and-regenerate extension, completed chunks remain causal context rather than being jointly denoised again. Our distinction is not the causal mask alone, but the role of the causal interface across learning: it is established on broad Stage-1 data, preserved while multi-view and structured-control capabilities are attached under scarcer supervision, and explicitly aligned to generated history. 

LingBot-World-Infinity~\cite{lingbotworldinfinity2026} begins from the observation that an interactive world is a temporal state process: each state is generated from past observations and currently available inputs, rather than from future frames or instructions. Its reason for causal pretraining is therefore the temporal factorization of deployment, not the availability of a particular supervision subset. Training the deployable transition first produces a causal world-model teacher whose dynamics can subsequently be accelerated without changing the state definition. Its MoBA mask is not purely autoregressive: a bidirectional component regularizes teacher forcing to preserve visual fidelity and flexible-length behavior, while a lower-triangular text mask prevents future-prompt leakage on the autoregressive component. Consistency distillation then reduces denoising cost, and self-rollout DMD targets the state distribution induced by the student itself. HelloWorld adopts the same high-level sequence of causal state definition, teacher training, deployment-state alignment, and few-step distillation. It instantiates this sequence with chunk-causal driving observations, nested geometric supervision, and explicit cache-boundary training. The benefits of LingBot's complete recipe do not by themselves establish that HelloWorld's early causal adaptation is superior to a matched bidirectional-to-causal retrofit.

\subsection{Structured Control Architectures}

ControlNet~\cite{zhang2023controlnet} popularized a trainable side branch with zero-initialized connections for adding spatial control while protecting a pretrained diffusion backbone. VACE~\cite{jiang2025vace} extends adapter-style conditioning to a broad family of video creation and editing tasks. Driving generators instantiate these ideas with maps, occupancy, boxes, depth, or LiDAR~\cite{yang2023bevcontrol,gao2024magicdrivev2,nvidia2025cosmostransfer}. The main design axes are where controls are encoded, whether multiple modalities share a branch, and how control strength varies across space and time.

HelloWorld adopts a VACE-style video control tower for rendered HD maps and 3D boxes, but does not ask that tower to represent ego pose. Pose modulates the main network through Plücker features and AdaLN, while structured control enters through VAE latents and periodic tower hints. Stage~3 explicitly trains joint, pose-only, and control-only batches. Consequently, dropping either input at inference is a trained operating mode rather than an architectural modification or an assumption that the remaining condition reconstructs the missing one.

\subsection{Few-Step Generation}

Consistency models~\cite{song2023consistency} and their continuous-time scaling variants~\cite{lu2024scm} learn mappings that reduce the number of numerical sampling steps. Distribution matching distillation~\cite{yin2024dmd} and DMD2~\cite{yin2024dmd2} instead train a fast generator against distribution-level score differences. Most formulations are evaluated on images or self-contained clips. In an autoregressive world model, however, a small per-window distribution shift can become the next window's context and accumulate over time. HelloWorld combines packed teacher-forced consistency training with self-forcing distribution matching. The distinction is the role of these objectives within a progressively trained driving generator: capability acquisition, generated-history alignment and compression share the same causal observation interface. Teacher/student comparisons assess retained geometry and control together with execution cost.

\input{sections/lidar_related}

\subsection{Capability Positioning}
\label{sec:positioning}

Table~\ref{tab:capability} summarizes the capability gap discussed above. It focuses on the properties of the core generator. LiDAR representation and conditional generation are developed in Sections~\ref{sec:method-lidar-vae} and~\ref{sec:method-lidar-dit}, with generated-video applications in Section~\ref{sec:app-lidar}. The relevant distinction is not whether a model accepts any motion input, but whether metric ego pose is independently controllable rather than represented by a low-dimensional action or embedded in a projected layout sequence.

\begin{table}[t]
\centering
\caption{Capability comparison with representative world models. ``Pose form'' distinguishes independent metric 6-DoF ego pose, low-dimensional action, and ego motion implicit in layout conditioning. \cmark{} = supported, \pmark{} = partial/preliminary, \xmark{} = not reported. Entries reflect public reports as of Aug.\ 2026.}
\label{tab:capability}
\small
\setlength{\tabcolsep}{5pt}
\begin{tabular}{lccccc}
\toprule
 & \makecell{Pose\\form} & \makecell{Multi-\\view} & \makecell{Structured\\control} & \makecell{Causal\\AR} & \makecell{Few-\\step} \\
\midrule
MagicDrive-V2~\cite{gao2024magicdrivev2} & layout & \cmark & \cmark & \xmark & \xmark \\
GAIA-2~\cite{russell2025gaia2} & action & \cmark & \cmark & \cmark & \xmark \\
Cosmos-Transfer~\cite{nvidia2025cosmostransfer} & layout & \pmark & \cmark & \pmark & \pmark \\
Cosmos WFM~\cite{nvidia2025cosmos} & action & \xmark & \pmark & \cmark & \cmark \\
LingBot-World~\cite{lingbotworld2026} & action & \xmark & \xmark & \cmark & \cmark \\
HY-World 1.5~\cite{hyworld2025} & action & \xmark & \xmark & \cmark & \cmark \\
Genie 3~\cite{genie3} & action & \xmark & \xmark & \cmark & \cmark \\
\midrule
\textbf{HelloWorld (ours)} & \textbf{6-DoF} & \cmark & \cmark & \cmark & \cmark \\
\bottomrule
\end{tabular}
\end{table}

%% file: sections/lidar_related.tex
\subsection{LiDAR Representation, Conditional Generation, and Rectification}
\label{sec:related-lidar}

\paragraph{Learning a geometric video representation.}
Video autoencoders such as Wan~\cite{wan2025} provide a temporally compressed latent interface, but range images are not ordinary color images. Their values represent metric surfaces, while absent returns require a separate interpretation. We retain a pretrained video-VAE architecture and adapt its representation and objectives: explicit return validity separates existence from range, and noman and Haar constraints address unsupported intermediate depths and high-frequency structure. The contribution is this geometry-specific training design, not a new video-VAE backbone or a claim that BCE and Haar transforms are new.

\paragraph{From visual conditions to sensor observations.}
Cosmos~\cite{nvidia2025cosmos} provides a foundation-model setting for physical-AI generation; its RGB-to-LiDAR generator serves as a quantitative reference here. UniScene~\cite{li2024uniscene} organizes driving-scene generation around occupancy. UniDriveDreamer~\cite{zhao2026unidrivedreamer} introduces modality-specific autoencoders and Unified Latent Anchoring for multimodal diffusion. Sensor2Sensor~\cite{wang2026sensor2sensor} translates monocular dashcam video into a multi-camera and LiDAR sensor suite. Our task is narrower: given synchronized RGB windows, synthesize only LiDAR. We combine native-grid token packing, azimuth/coverage ray features, latent anchoring, and geometric supervision through a frozen decoder. Packing and latent alignment adapt the pretrained interface; the key training link is that generated latents must also satisfy output-space geometric criteria.

\paragraph{Geometry after range-image decoding.}
L3DR~\cite{liu2026l3dr} addresses depth bleeding and wavy surfaces with three-dimensional residual regression and robust Welsch supervision. We adopt this rectification principle with radial updates that retain each original ray, and study single-frame and motion-compensated temporal variants. Because both VAE reconstruction and DiT generation end in the same range representation, the rectifier interface is reusable; this does not imply that a gain measured on reconstruction transfers unchanged to generation.

%% file: sections/data.tex
\section{Data Construction}
\label{sec:data}

\begin{figure}[htbp]
\centering
\includegraphics[width=\linewidth]{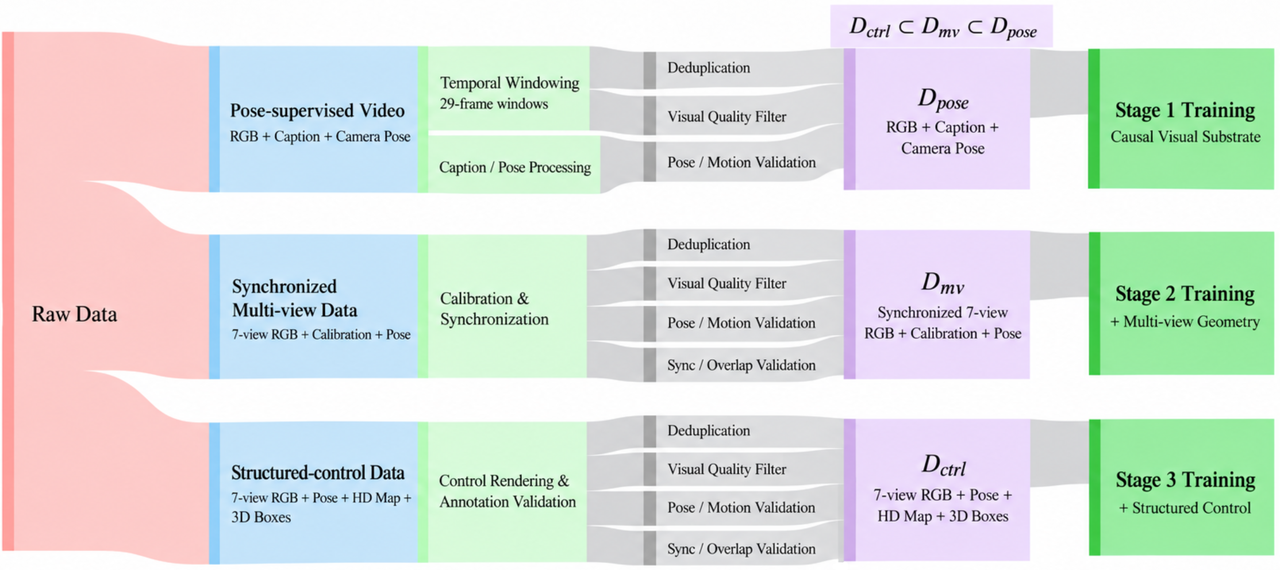}
\caption{Data construction under nested supervision. Pose-supervised video provides the broad pool for learning causal visual dynamics. Synchronized camera observations add multi-view geometry, and the structured-control subset supplies rendered maps and object annotations. Source-specific preparation and quality validation connect these supervision levels to the three training stages. The flows illustrate processing and supervision relationships rather than measured data proportions.}
\label{fig:pipeline}
\end{figure}

HelloWorld combines broad visual generalization with driving-specific control and synchronized observation generation, but these capabilities rely on data with different levels of supervision. General and fleet video provide large-scale appearance and motion coverage, synchronized camera rigs introduce calibrated multi-view geometry, and a smaller annotated subset provides HD maps and 3D object layouts for explicit scene control. Figure~\ref{fig:pipeline} traces how these supervision levels feed the three training stages through source-specific preparation and quality checks. We describe the sources, conversion and quality control, semantic annotation and balancing, pose normalization, and data governance below.

\subsection{Data Sources and Supervision Availability}
\label{sec:data-sources}

The data inventory distinguishes training sources from the external evaluation benchmark:

\begin{enumerate}[leftmargin=2em]
  \item \textbf{Fleet data}: real 7-camera vehicle logs (front-wide, front-tele, left/right-front, left/right-back, rear, 10Hz, mixed FOVs, fixed extrinsics) with per-frame metric ego pose and scene tags. An annotated subset carries HD map and 3D bounding-box labels. The curated pool snapshot contains 1,915,569 quality-gated 40s clips, from which the balanced selection of Section~\ref{sec:data-balance} retains 497,773.
  \item \textbf{General-purpose corpora}: DL3DV, RealEstate10K, SpatialVID, Sekai (real-walking and game subsets), OmniWorld-Game, ScanNet, and MatrixCity. These corpora provide video with recoverable but \emph{scale-free} camera trajectories and no driving-specific labels.
  \item \textbf{nuScenes}~\cite{caesar2020nuscenes}: the public benchmark used for external comparison (Section~\ref{sec:exp-nuscenes}).
\end{enumerate}

Table~\ref{tab:data} quantifies the Stage-1 mixture: fleet data contribute 35.1\% of the 29-frame training windows, with the remainder drawn from general-purpose video. The key observation is an asymmetry: \emph{camera pose is available or recoverable for all training sources, whereas synchronized calibration and structured scene annotations are available only for progressively smaller subsets of fleet data}. Consequently, pose-conditioned learning can use the broadest mixture of fleet and general-purpose video, while multi-view specialization and structured control rely on increasingly selective driving supervision. This supervision hierarchy motivates the progressive learning strategy described in Section~\ref{sec:method}: broad pose-conditioned modeling first, followed by synchronized multi-view specialization and structured scene-control adaptation.

\begin{table}[t]
\centering
\caption{Stage-1 training asset inventory (shares are pre-flatten window proportions, not effective sampler exposure): 29-frame training windows per source (754,629 sequences, 3,662,212 windows in total). Fleet windows are counted after caption-QA filtering of the 1,510,454 pose-balanced windows of Section~\ref{sec:data-balance}. With \SevenViewFlatten{} each fleet window yields 7 single-view samples.}
\label{tab:data}
\small
\begin{tabular}{lrrccc}
\toprule
Source & Windows (w29) & Share & Views & Metric & HDMap/BBox \\
\midrule
Fleet (7-camera)      & 1,283,914 & 35.1\% & 7 & \cmark & subset \\
SpatialVID            &   991,073 & 27.1\% & 1 & \xmark & \xmark \\
Sekai (real-walking)  &   746,509 & 20.4\% & 1 & \xmark & \xmark \\
DL3DV                 &   211,432 &  5.8\% & 1 & \xmark & \xmark \\
ScanNet               &   185,548 &  5.1\% & 1 & \xmark & \xmark \\
RealEstate10K         &   127,536 &  3.5\% & 1 & \xmark & \xmark \\
Sekai (game)          &    71,930 &  2.0\% & 1 & \xmark & \xmark \\
OmniWorld-Game        &    36,837 &  1.0\% & 1 & \xmark & \xmark \\
MatrixCity            &     7,433 &  0.2\% & 1 & \xmark & \xmark \\
\bottomrule
\end{tabular}
\end{table}

\paragraph{Nested supervision and progressive learning.}
Figure~\ref{fig:pipeline} organizes the training data by available supervision. Let $\mathcal{D}_{\mathrm{pose}}$ denote observations with camera trajectories, $\mathcal{D}_{\mathrm{mv}}$ the subset with synchronized and calibrated camera views, and $\mathcal{D}_{\mathrm{ctrl}}$ the subset that additionally carries structured scene annotations. At the level of source observations, these requirements define
\begin{equation}
\mathcal{D}_{\mathrm{ctrl}}\subset\mathcal{D}_{\mathrm{mv}}\subset\mathcal{D}_{\mathrm{pose}}.
\end{equation}
The branches are therefore overlapping supervision pools rather than disjoint partitions of raw data. A synchronized fleet observation can contribute independent camera streams to Stage~1, a grouped surround-view example to Stage~2, and, when scene labels are available, a controlled example to Stage~3. This organization allows broad video coverage to support appearance and motion learning before more selective geometric and layout supervision is introduced.

\subsection{Conversion Pipeline}
\label{sec:data-pipeline}

The preparation paths in Figure~\ref{fig:pipeline} enforce the observation requirements of each supervision pool. Pose-supervised video is partitioned into temporal windows and paired with captions and camera trajectories. Synchronized multi-view data additionally requires consistent camera calibration, ordering, and timestamps. Structured-control data further requires validated scene annotations rendered into the corresponding camera views. The illustrated 29-frame window is the standard training unit, with longer windows described in Section~\ref{sec:data-balance}.

Quality validation follows these requirements. Duplicate removal, visual quality filtering, and pose and motion validation support the common video interface. Multi-view and structured-control examples additionally require synchronization and overlap validation so their streams describe compatible observations of the same scene. Control annotation checks and rendering alignment ensure that the supplied geometry refers to the associated RGB frames. These checks establish sample validity, while the balancing procedure in Section~\ref{sec:data-balance} addresses the frequency of valid scenarios and maneuvers.

For fleet data, a common intermediate representation retains synchronized RGB, calibration, poses, and available scene annotations. HD maps and boxes are rasterized into per-camera control videos depicting lane lines, curbs, road marks, traffic signs and lights, and category-colored 3D boxes. Cross-frame lane-divider fusion reduces annotation flicker, and image-boundary clipping retains the visible portions of projected elements. The resulting segments preserve camera ordering and temporal alignment for pose encoding and condition retrieval during training.

\subsection{Captioning and Semantic Annotation}
\label{sec:data-caption}

Each segment carries VLM-generated captions in six stored variants (short/medium/long $\times$ with/without ego-behavior description), produced by qwen3.6-plus, plus structured VRI tags and scene tags (road environment, turn bin, weather, time of day). Captioning is performed per segment on 3 sampled frames at 720p, with the six caption variants and a 7-dimension scene-tag set produced in a single VLM call. Captions overlapping previously annotated clips are reused. The released Stage-1 training configuration consumes only the three \texttt{caption\_wo\_behavior\_\{short,medium,long\}} fields. The other variants remain data assets rather than training inputs. Training samples select the annotation window nearest to the segment center, avoiding stale captions that we have observed to cause semantic drift later in long generated videos. In synchronized multi-view training, front-wide receives the complete caption while the other cameras receive only a view-specific prefix. This prevents one scene description from being replicated as if it described every field of view and avoids duplicating the same directional description across views.

\subsection{Selection, Balancing, and Quality Gates}
\label{sec:data-balance}

Driving data is heavily biased toward straight, uneventful cruising: in the raw fleet pool, 82.8\% of clips are tagged \emph{straight road}, 69.9\% \emph{daytime}, 84.1\% \emph{medium speed}, while curves, night, rain, intersections, and pedestrian interactions each fall below a few percent. We rebalance in \emph{two orthogonal layers}: a clip-level tag balance decides \emph{which clips} enter the pool (appearance/semantic diversity), and a segment-level pose balance decides \emph{which windows within each clip} are trained on (geometric/maneuver ratios). The two layers are deliberately separated because clip-level tags are 40s-window unions. In the pool, 94.7\% of high-curvature-turn clips also carry the \emph{straight} tag (a 40s clip is typically 25s approach + 8s turn + 7s exit). Consequently, clip-level balance alone would not translate into window-level balance.

\paragraph{Layer 1: clip-level tag balancing.}
Starting from a quality-gated pool of 1,915,569 clips (snapshot 2026-06-11), low-image-quality clips (fog on lens or glare, 53,441) are removed, leaving 1,862,128. The 445 raw scene tags are folded into a balancing vocabulary of 29 categories / 198 nodes (deepest-node deduplication per category, pipeline-QC tags excluded). Selection then applies a three-tier policy:
\begin{itemize}[leftmargin=2em]
  \item \emph{Hard-lock true long tail}: all clips carrying any of the 114 tags with $n_t \le 10\mathrm{k}$ (tunnels, construction, roundabouts, hard braking, standing water, animals, \ldots) are retained at 100\%.
  \item \emph{Hard-lock safety/interaction whitelist}: 14 mid-frequency whitelist tags ($10\mathrm{k} < n_t \le 100\mathrm{k}$: high-curvature turns, unprotected left turns, rain, cut-ins, U-turns, ramps, \ldots) are retained at 100\%. Four high-frequency whitelist tags (night, fog, low/mid-curvature turns, dawn/dusk) are exempt from capping and survive via co-occurrence.
  \item \emph{Cap-and-trim the head}: remaining candidates are greedily removed in ascending order of a rarity weight $w_i = \max_{t\in T_i} \left(N/\max(n_t, 50)\right)^{0.5}$, stratified over 23,405 (city, vehicle$\times$day) strata so that redundant clips from the same collection session are removed first, until head-tag counts reach their caps.
\end{itemize}
The result is a selection of 497,773 clips (26.7\% of the pool): every rare tag is preserved verbatim, long-tail \emph{share} rises uniformly by $3.74\times$ (e.g., high-curvature turns 4.9\%$\rightarrow$18.4\%, unprotected left turns 4.6\%$\rightarrow$17.1\%, rain 3.7\%$\rightarrow$13.9\%), while head tags lose 72\% of absolute volume. Residual head-share shaping (e.g., pushing the effective straight-road share to a target) is done by per-tag repeat weights in the training cache, which is reversible and tunable, rather than by destroying locked long-tail clips.

\paragraph{Layer 2: segment-level pose balancing.}
Within the selected clips (493,500 unique sequences), non-overlapping 29-frame ($\approx$2.9s) and 61-frame ($\approx$6.1s) windows are cut under a per-frame validity mask requiring all 7 cameras, metric pose, and sane timestamps for every frame in the window. A sampled audit found that 63\% of clips contain mid-clip invalid frames, which fixed head/tail margins would miss. Each window is characterized purely from pose: cumulative yaw turn angle (5 bins), median speed (5 buckets, using the median to retain sensitivity to hard stops), and maximum longitudinal acceleration. Quotas allocate 30\% to straight windows (uniform over the 5 speed buckets) and 25/20/15/10\% to increasing turn-angle bins, with the largest bin (${\ge}40^\circ$) taken exhaustively as the bottleneck. Windows with $\max|a| \ge 3\,\mathrm{m/s^2}$ ($\approx$0.5\% of supply, the empirically rare regime) bypass quotas entirely. From 5,436,368 candidate 29-frame windows this yields 1,510,454 balanced training windows (and 1,121,947 61-frame windows), with straight-speed buckets exactly equalized at 88,919 each.

Before training, a Data QA contract is enforced per segment: correct camera ordering across all views, box-drift bounds, and caption field alignment.

\subsection{Pose Representation and Scale Unification}
\label{sec:data-pose}

For fleet data, per-frame ground-truth metric camera pose is composed as $\mathrm{vcs2boot} \circ \mathrm{camera2vcs}$. For general corpora, trajectories are recovered up to scale. A fixed $\mathrm{OpenCV\ camera}\rightarrow\mathrm{vehicle}$ rotation bridges the OpenCV convention ($x$ right, $y$ down, $z$ forward) and the vehicle-control convention ($x$ forward, $y$ left, $z$ up), so commanded yaw and pitch retain their physical meaning. Unless stated otherwise, conditioning uses frame-to-frame relative transforms rather than expressing every frame only against the clip's first frame.

Each transform is converted online into a dense six-channel Pl\"ucker raymap.
For camera center $o$ and unit ray direction $d$, each pixel stores the
ray moment $m=o\times d$ followed by the direction $d$.
For non-metric sequences, translation is normalized by the median magnitude
of frame-to-frame displacement over the complete sequence, with degenerate
static sequences falling back to unit scale.
The same translation normalization is applied consistently during training
and inference.
This canonicalization preserves within-sequence motion ratios while reducing
arbitrary scale variation across reconstructed corpora.

To retain the distinction between metric fleet trajectories and
scale-free reconstructed trajectories, each pose sequence is additionally
associated with a binary metric-scale indicator.
Together with the normalized camera trajectory, this forms the common
pose-conditioning contract that allows metric driving logs and
scale-ambiguous general video to coexist in the same training mixture.
The model-side Pl\"ucker encoding, metric-scale conditioning,
and pose injection are described in Section~\ref{sec:method-stage1}.

\subsection{Data Governance}
\label{sec:data-governance}

Fleet data undergoes face and license-plate anonymization. Third-party corpora and pretrained weights remain subject to their respective licenses. Their use does not confer redistribution rights. Training and evaluation sequences are strictly disjoint. Geographically sensitive information is reported only as auditable aggregate statistics.

\paragraph{LiDAR representation and paired generation data.}
The LiDAR tokenizer dataset comprises 50,000 29-frame segments (1,450,000 frames), separate from the RGB foundation inventory. It uses main LiDAR sensors 0--3 and excludes blind-spot sensors 4--7. Full $360^\circ$ range observations are re-binned from 3600 to 1792 columns. The Plan-B representation uses a minimum range of 1 m and an ego-body box $x\in[-1.1,4.0]$, $y\in[-1.1,1.1]$ m. Return validity and geometric compression are described in Section~\ref{sec:app-lidar-implementation}. 
VAE training observes LiDAR alone; conditional DiT training additionally requires synchronized RGB and calibrated sensor geometry. Rectifier training uses paired VAE reconstructions and measured LiDAR, with validation windows withheld from rectifier training. Reconstruction validation, the 100-clip generation evaluation, and the eight-road-clip structural comparison are separate populations; scores are not pooled across them. Generated-input application examples have no matched physical target scan.

\paragraph{Calibration and condition provenance.}
The V5 adapter reads camera calibration from external calibration files when camera metadata is absent, and inverts vehicle-to-camera extrinsics into the camera-to-vehicle contract. Layout-only preprocessing can instead consume self-contained maps, per-frame objects, ego poses and seven-camera calibration without fetching source imagery. Training and deployment must use the same object classes, lane styles and traffic-light rendering semantics. Calibration, rendering semantics and adapter versions form part of each experiment's condition manifest.

%% file: sections/method.tex
\section{Method}
\label{sec:method}

\begin{figure}[t]
    \centering
    \includegraphics[width=\linewidth]{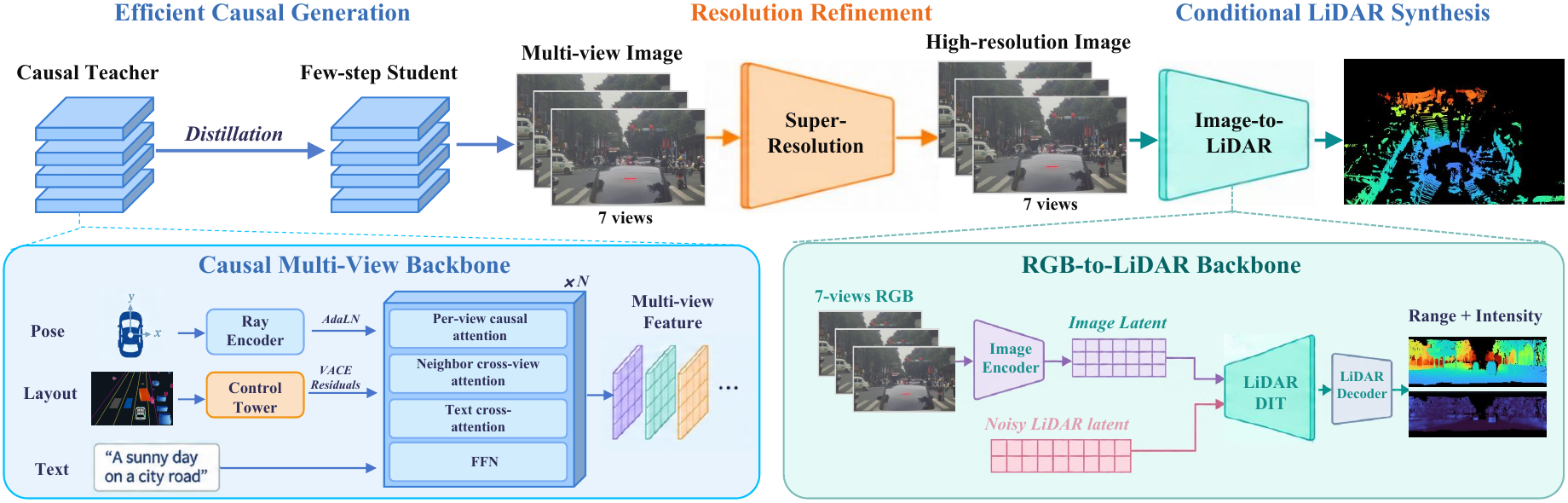}
    \caption{\textbf{Overview of the HelloWorld generation framework.}
    A causal teacher is distilled into a few-step student for seven-view RGB generation.
    A separate super-resolution module refines the generated views before conditional
    LiDAR synthesis. (a) The causal multi-view backbone incorporates ego-pose geometry
    through a ray encoder and AdaLN, scene layout through residuals from a control tower,
    and text through cross-attention, while cross-view attention exchanges information between
    neighboring cameras. (b) The RGB-to-LiDAR backbone conditions a LiDAR diffusion
    on encoded multi-view RGB latents while denoising LiDAR latents, which
    are decoded into range and intensity.}
    \label{fig:arch}
\end{figure}

HelloWorld progressively transforms broad video priors into a controllable driving world model for data generation and interactive simulation. The RGB core first learns pose-conditioned visual dynamics from heterogeneous video and is then specialized using synchronized multi-view observations and structured scene supervision. After capability learning, we align the model with repeated sequential generation by exposing it to imperfect and self-generated observation histories, and finally distill the multi-step generator into a few-step student for efficient inference. Figure~\ref{fig:arch} locates these mechanisms in the final architecture, from pose and scene conditioning in the causal RGB backbone to the distilled RGB generator and RGB-conditioned LiDAR branch. We first define the observation-generation task and common generative backbone, then describe broad pose-conditioned learning, structured driving specialization, sequential simulation alignment, and few-step generation. The extension from RGB observations to conditional LiDAR synthesis is presented separately at the end of this section.

\subsection{Overview and Problem Formulation}

\paragraph{Observation-generation task.}
Let $Y_k=\{Y_k^{(v)}\}_{v=1}^{V}$ denote the RGB observations generated for temporal chunk $k$, where $V$ is the number of camera views. Let $C_k^{\mathrm{sup}}$ collect the externally supplied conditions, including text, camera poses, and structured scene controls, and let $S_{k-1}$ denote the committed observation history before generating the current chunk. HelloWorld models the conditional observation transition
\begin{equation}
Y_k \sim p_\theta\left(\cdot \mid S_{k-1}, C_k^{\mathrm{sup}}\right),
\qquad
S_k=\mathcal{U}(S_{k-1},Y_k,C_k^{\mathrm{sup}}),
\label{eq}
\end{equation}
where $\mathcal{U}$ denotes history commitment and context management after a chunk has been generated. This formulation separates observation generation from vehicle dynamics: ego trajectories and scene configurations are prescribed conditions, while HelloWorld predicts the corresponding sensor observations. The model does not infer low-level vehicle actions or propagate an explicit dynamics model from control commands.

\paragraph{Chunk-sequential generation.}
Video is encoded by a causal video VAE and partitioned into temporal latent chunks. Frames within the current chunk are generated jointly, while previously completed chunks form fixed visual context for subsequent predictions. The chunk therefore serves as the common unit for latent generation, history commitment, causal attention, and cache management. The causal claim in this work concerns observation history: future visual observations are inaccessible when predicting the current chunk. Supplied condition sequences may cover a larger temporal window, and strictly online condition causality would additionally require restricting every conditioning pathway to information available at the current simulation step.

\paragraph{Training-stage terminology.}
The method is organized below by capability rather than by training chronology. For consistency with the experiments, however, we retain three checkpoint names. \textbf{Stage~1} denotes the model after broad pose-conditioned learning, \textbf{Stage~2} denotes the model after synchronized multi-view specialization, and \textbf{Stage~3} denotes the model after structured scene-control adaptation. Subsequent sequential-alignment and distillation stages operate on the resulting controllable multi-view model.

\subsection{Generative Backbone and Factorized Conditioning}

\paragraph{Latent rectified-flow backbone.}
HelloWorld performs generation in the latent space of a causal video VAE using a rectified-flow objective.
Given clean video latents $x_0$ and Gaussian noise $\epsilon\sim\mathcal{N}(0,I)$, we construct
\begin{equation}
x_\sigma=(1-\sigma)x_0+\sigma\epsilon,
\qquad \sigma\in[0,1],
\end{equation}
and train a velocity network $v_\theta$ to predict the constant transport direction:
\begin{equation}
\mathcal{L}_{\mathrm{RF}}
=
\mathbb{E}_{x_0,\epsilon,\sigma}
\left[
\left\|
v_\theta(x_\sigma,\sigma,c)-(\epsilon-x_0)
\right\|_2^2
\right],
\label{eq:rf}
\end{equation}
where $c$ contains the available text, pose, view, and scene-layout conditions.
Operating in latent space reduces the spatial and temporal cost of video generation while retaining the common generative interface used throughout all RGB training stages.

\paragraph{Block-causal multi-view DiT.}
The denoising backbone uses full spatiotemporal attention within the chunk currently being generated and causal access across temporal chunks.
This factorization allows all latent tokens of the current observation to be jointly refined while preventing future observations from modifying already committed history.
Block causality therefore defines the temporal interface of the model rather than a separate prediction objective.
At multi-view stages, temporal self-attention remains organized per camera stream, while dedicated cross-view pathways exchange information among synchronized cameras at the same latent time.

\paragraph{Factorized conditioning.}
We separate conditioning according to the physical quantity that each signal constrains.
\emph{Camera identity} specifies the observation view through a learned camera embedding.
\emph{Ego pose} specifies observer geometry and is encoded from dense Pl\"ucker raymaps before modulating the backbone through adaptive normalization.
\emph{Scene layout}, represented by rendered HD maps and 3D boxes, is processed by a dedicated control tower that injects residual features at multiple backbone depths.
Text provides complementary semantic information through cross-attention.
This factorization avoids forcing geometrically distinct controls into a single token pathway and allows observer motion and scene configuration to be manipulated independently.

\subsection{Causal Observation Modeling}
\label{sec:method-causal}

\paragraph{Block-causal latent modeling.}
To align the temporal information flow during training with sequential generation,
HelloWorld adopts block-causal observation modeling from the beginning of driving adaptation.
The video latent sequence is partitioned into temporal chunks
$\{x^{(1)},\ldots,x^{(K)}\}$.
Tokens within the same chunk interact bidirectionally, allowing each local observation to be modeled jointly,
whereas attention across chunks is causal:
a query in chunk $k$ can access observation tokens from chunks $j\leq k$ but not from future chunks $j>k$.
The corresponding attention mask is
\begin{equation}
M_{ij}
=
\begin{cases}
1, & \mathrm{chunk}(j)\leq\mathrm{chunk}(i),\\
0, & \mathrm{otherwise}.
\end{cases}
\label{eq:block-causal-mask}
\end{equation}
This observation-access pattern is retained when synchronized multi-view interaction and structured scene control are introduced in later specialization stages.

\paragraph{Chunk-wise causal training.}
The block-causal mask determines which observations can be accessed,
while chunk-wise rectified-flow times control the quality of the available history.
For each chunk $k$, we construct
\begin{equation}
x_{\sigma_k}^{(k)}
=
(1-\sigma_k)x_0^{(k)}
+
\sigma_k\epsilon^{(k)},
\qquad
\epsilon^{(k)}\sim\mathcal{N}(0,I),
\label{eq:chunk-rf}
\end{equation}
where the noise level $\sigma_k$ may vary across temporal chunks.
The model predicts the corresponding velocity targets under the same block-causal observation mask:
\begin{equation}
\mathcal{L}_{\mathrm{causal\text{-}RF}}
=
\mathbb{E}
\left[
\sum_{k\in\mathcal{K}_{\mathrm{tgt}}}
\left\|
v_\theta^{(k)}
-
\left(
\epsilon^{(k)}-x_0^{(k)}
\right)
\right\|_2^2
\right].
\label{eq:causal-rf}
\end{equation}
By varying the noise configuration across chunks,
the same causal training interface exposes the predictor to histories of different quality.
We use independently sampled, progressively cleaner, and clean-context configurations depending on the training phase.

\paragraph{Context corruption and \SelfRollout{}.}
Chunk-wise noise training varies the quality of the available history,
but standard causal training still conditions predominantly on observations derived from recorded data.
During sequential generation, later predictions increasingly depend on observations generated by the model itself,
creating a distribution gap between recorded and generated histories.
We address this discrepancy at two levels.
First, context corruption perturbs preceding latent chunks with noise, appearance degradation, and spatial distortions,
broadening the local neighborhood of histories encountered during training.
Second, \SelfRollout{} directly exposes the model to its own prediction distribution.
It first generates a variable-length history with gradients disabled and then evaluates the rectified-flow objective for subsequent chunks while conditioning on that generated history.
Rollout generation and optimization are therefore separated, avoiding backpropagation through the complete generated trajectory.
Context corruption primarily models local degradation around recorded histories,
whereas \SelfRollout{} additionally captures structured errors induced by the model's own generation process.

\paragraph{Causal execution and cache management.}
At inference, HelloWorld follows the same chunk-sequential observation interface.
A generated chunk is not exposed to subsequent predictions until its denoising trajectory reaches the clean endpoint.
The completed chunk is then committed to the history state and written into a bounded KV cache,
so future predictions condition only on completed observations rather than intermediate denoising states.

To maintain consistent relative positions as the active temporal window moves,
cached keys are stored without positional rotation,
and positional offsets are reapplied according to their positions in the current window.
This separates persistent observation content from its temporary location within the active cache.

\subsection{Broad Pose-Conditioned World Modeling}
\label{sec:method-stage1}
The first specialization step aims to preserve the breadth of heterogeneous video while introducing an explicit camera-motion interface.
Because camera trajectories are either directly available or recoverable for substantially more video than synchronized multi-view or scene-layout annotations, pose-conditioned learning can exploit the broadest supervision pool.
General-purpose video contributes diverse appearance, object, and motion statistics, while fleet video introduces driving-specific visual dynamics and metric motion.
Stage~1 therefore learns a shared pose-conditioned RGB generator before more selective driving supervision is introduced.

\paragraph{\SevenViewFlatten{}.}
Synchronized fleet clips contain multiple camera streams, whereas most general video contains only a single moving camera.
To place these sources under the same learning interface, we flatten each fleet camera stream into an independent pose-conditioned training sample during Stage~1.
This preserves all fleet observations while allowing them to be mixed directly with general video under a common single-stream objective.
Cross-view correspondence is intentionally deferred to later specialization, where synchronized observations are explicitly grouped again.

\paragraph{Pose representation.}
Each camera trajectory is converted into dense Pl\"ucker raymaps aligned with the video latent grid.
For camera center $o$ and unit ray direction $d$, each pixel stores the ray moment $m=o\times d$ together with $d$.
Fleet trajectories retain metric translation, whereas reconstructed trajectories from general video are normalized by their sequence-level motion scale.
We append the binary \MetricScaleFlag{} as an additional dense pose channel to distinguish metric and scale-free regimes within the same pose encoder.
Its newly introduced input weights are initialized to zero so that adding the flag does not perturb the pretrained pathway at initialization.
A separately learned null-pose representation distinguishes missing pose conditioning from a physically stationary camera.
The encoded pose features are temporally aligned with the latent video features and injected into the backbone through adaptive layer normalization.

\subsection{Structured Driving Specialization}
\label{sec:method-structured}
Broad pose-conditioned learning provides general visual priors and camera-motion control, but driving simulation additionally benefits from structured supervision that ties multiple views and scene-level conditions to the same underlying world state.
We therefore specialize the Stage~1 model in two steps.
Synchronized camera observations first introduce explicit cross-view interaction, after which HD maps and 3D boxes provide a separate interface for manipulating scene configuration.

\subsubsection{Synchronized Multi-View Modeling}
\label{sec:method-stage2}

\paragraph{\ZeroInitTransfer{}.} 
Cross-view interaction is introduced through a residual attention branch inserted after per-view self-attention.
Each camera communicates with its calibrated neighboring cameras at the same latent time, allowing correspondence to emerge without assuming pixel-level alignment.
Query, key, value, and normalization parameters are initialized from the corresponding self-attention layer, while the output projection of the new branch is initialized to zero.
For hidden features $h$, the added branch has the form
\begin{equation}
h'
=
h+
W_o\,\mathrm{Attn}_{\mathrm{neighbor}}(h),
\qquad
W_o=0
\quad\Longrightarrow\quad
h'=h.
\label{eq:zeroinit-transfer}
\end{equation}
The inserted branch is therefore initially inactive and learns cross-view interaction progressively during optimization.
Restricting communication to neighboring cameras encodes the rig topology and avoids dense interaction among every view pair.

\paragraph{Multi-view adaptation.}
Zero initialization guarantees an identity mapping only for the newly inserted residual branch when its output bias is also zero;
it does not imply exact equivalence of the complete architecture once camera embeddings and multi-view organization are introduced.
We therefore fine-tune the full Stage~2 network, allowing appearance, pose response, and cross-view communication to adapt jointly.
The Stage~2 checkpoint used in our experiments corresponds to the model after this synchronized multi-view specialization and before structured scene control is introduced.

\subsubsection{Explicit Scene Control}
\label{sec:method-stage3}
Multi-view specialization constrains how the scene is observed, but counterfactual generation additionally requires an explicit interface for modifying the scene itself.
We distinguish \emph{observer motion}, specified by ego pose, from \emph{scene configuration}, specified by HD maps and 3D object boxes.
Keeping these factors separate enables interventions on camera motion and scene layout through independent conditioning pathways.

\paragraph{Structured-control tower.}
HD maps and 3D boxes are rendered into camera-aligned control videos and encoded into latent features.
A VACE-style control tower transforms these features into residual signals that are injected at multiple depths of the RGB backbone.
Cross-view branches within the control pathway allow layout information to propagate across synchronized cameras, while a learnable scalar gate modulates the overall contribution of structured control.
Pose conditioning remains in the geometric modulation pathway and is not replaced by the layout representation.
Consequently, either pose or layout conditioning can be disabled without modifying the model architecture.

\paragraph{Progressive control adaptation.}
We introduce the structured-control pathway in two phases.
We first freeze the established multi-view backbone and optimize only the newly added control tower on examples with active scene-layout conditioning.
This allows the control pathway to acquire a meaningful influence before interacting with the pretrained visual representation.
We then fine-tune the complete model using a mixture of joint, pose-only, and control-only examples.
The mixture exposes the same network to different subsets of conditions and encourages it to retain the corresponding interfaces after joint adaptation.
When ego pose is changed while world-coordinate map and object geometry are held fixed, the camera-projected controls are re-rendered for the new viewpoint.

\subsection{Efficient Few-Step Generation}
\label{sec:method-distill}

Repeated use of the world model makes sampling cost part of the system design.
The multi-step rectified-flow model provides a strong teacher but requires multiple network evaluations for every generated chunk.
We therefore distill the same observation and control interface into a few-step student.
The student is first initialized using discretized consistency training,
then optimized with continuous-time consistency under recorded history and self-forcing distribution matching under its own generated samples.

\subsubsection{Consistency Initialization and Training}
\label{sec:method-consistency}

\paragraph{Packed teacher forcing.}
For each view, clean and noisy latent sequences are concatenated as $[x_0\mid x_\sigma]$.
Clean queries follow the standard block-causal mask.
A noisy query belonging to chunk $k$ may attend to clean chunks strictly before $k$ and to its own noisy chunk, but not to the corresponding clean target or to future clean chunks.
Conditions follow the same view and temporal ordering, and the consistency objective is evaluated only on noisy target tokens.
This packed formulation allows all target chunks to be trained in a single forward pass while preserving the history available under teacher forcing.

\paragraph{Discretized consistency initialization.}
We first initialize the few-step student with discretized consistency training.
Adjacent noise levels are sampled independently for each target chunk, and a frozen guided teacher advances the noisy state by one Euler step under the teacher flow.
The student is trained so that predictions from the original and teacher-advanced states map to a consistent clean estimate, with the latter branch stop-gradient.
This initialization compresses local segments of the teacher trajectory before continuous-time consistency and distribution matching are introduced.

\paragraph{Continuous-time consistency.}
Continuous-time consistency extends the same objective beyond neighboring discretization levels.
Let $v_\theta$ denote the student velocity, $v_T$ the guided teacher velocity, and $M$ the noisy-target mask.
The directional derivative of the student along the teacher flow is
\begin{equation}
D_Tv_\theta
=
J_{(x,\sigma)}v_\theta
\,[M v_T,M].
\end{equation}
We construct the surrogate direction
\begin{equation}
g
=
M\bigl(
v_T-v_\theta-r\sigma D_Tv_\theta
\bigr),
\qquad
\widehat g
=
\frac{g}{\lVert g\rVert_2+\varepsilon_{\mathrm{s}}},
\end{equation}
and optimize
\begin{equation}
\mathcal{L}_{\mathrm{sCM}}
=
\mathbb{E}
\left[
\left\|
M\left(
v_\theta-\operatorname{sg}(v_\theta)-\operatorname{sg}(\widehat g)
\right)
\right\|_2^2
\right],
\label{eq:scm}
\end{equation}
where $\operatorname{sg}$ denotes stop-gradient and $r$ gradually introduces the tangent term during warmup.
Clean history is held fixed while the consistency constraint is applied to the current noisy prediction.
The directional derivative is evaluated with an exact Jacobian-vector product.

\subsubsection{Self-Forcing Distribution Matching}
\label{sec:method-dmd}
Consistency training compresses the teacher trajectory under observed history, but the deployed student generates both its outputs and subsequent context through its own few-step execution.
We therefore additionally optimize the student on its induced sample distribution.
The student first produces a sample $G$ by chunk-sequential few-step generation through the same KV-cache interface used at inference.
The generated sample is then re-noised as
\begin{equation}
x_\sigma=(1-\sigma)G+\sigma\epsilon.
\end{equation}
A frozen rectified-flow teacher and a trainable fake-score network predict corresponding clean estimates $\widehat x_T$ and $\widehat x_F$.
The fake-score network is initialized from the teacher and trained with a separate optimizer on the student's generated distribution.

The difference between teacher and fake-score predictions defines a normalized correction direction,
\begin{equation}
\Delta
=
\frac{
\widehat x_F-\widehat x_T
}{
\max\!\left(
\operatorname{mean}_{M}|G-\widehat x_T|,
\varepsilon_{\mathrm{d}}
\right)
},
\end{equation}
and the student receives the surrogate distribution-matching objective
\begin{equation}
\mathcal{L}_{\mathrm{DMD}}
=
\mathbb{E}
\left[
\left\|
M\left(
G-\operatorname{sg}(G-\Delta)
\right)
\right\|_2^2
\right].
\label{eq:dmd}
\end{equation}
The combined student objective is
\begin{equation}
\mathcal{L}_{\mathrm{student}}
=
\lambda_{\mathrm{c}}\mathcal{L}_{\mathrm{sCM}}
+
\lambda_{\mathrm{d}}\mathcal{L}_{\mathrm{DMD}}.
\end{equation}
Consistency training begins before distribution matching is activated, providing a stable few-step initialization before optimization on the student's self-generated distribution.

\subsubsection{Few-Step Sequential Inference}
\label{sec:method-fewstep-inference}

Few-step inference follows a descending rectified-flow noise schedule.
At each denoising step, the student predicts a clean latent estimate and re-noises it to the next scheduled level.
After the final step, the resulting chunk is committed to the KV cache and becomes context for the subsequent prediction.
The distilled model retains the same text, pose, scene-control, and cross-view interfaces as the teacher, so acceleration does not require a separate deployment architecture.
The same procedure is used for synchronized multi-view generation, with cross-view communication evaluated at every student step.

\subsection{Multi-Sensor Extension: Conditional LiDAR Generation}
\label{sec:method-lidar}

\FloatBarrier
\input{sections/lidar_implementation}
\FloatBarrier

%% file: sections/lidar_implementation.tex
\label{sec:app-lidar-implementation}
We extend RGB observations to LiDAR through two complementary tasks: learning a geometry-preserving video representation and generating LiDAR conditioned on multi-camera RGB. The stages share output-space geometric supervision, while only the LiDAR stream is synthesized during conditional generation.

\subsubsection{Geometry-preserving LiDAR VAE}
\label{sec:method-lidar-vae}
We adapt the Wan 2.1 causal video VAE~\cite{wan2025}, initialized from the tokenizer distributed with Cosmos-Predict2.5, to LiDAR sequences using weights separate from the RGB tokenizer. The three-channel input and reconstruction target are $[\bar d,\bar d,2m-1]$: the first two channels duplicate normalized range, and the third encodes binary return validity $m$. There is no intensity channel. Representing missing returns with a fixed depth value forces regression to learn whether a surface exists and how far away it is through the same signal. Instead, the validity channel records measured returns independently of the continuous depth canvas. This separates missing-return prediction from depth reconstruction and avoids treating interpolated holes as observed surfaces.

Validity alone cannot eliminate flying points: a ray with a genuine return can still be reconstructed between foreground and background. To reduce flying-point artifacts at depth discontinuities, we augment range reconstruction and validity classification with complementary Noman and Haar losses. \emph{Noman (no-man's-land)} identifies depth gaps at foreground--background boundaries in the reference and penalizes predictions inside their empty interior, excluding locations where the reference itself contains an intermediate surface. It discourages floating points between real surfaces, while range reconstruction determines which surface to recover. \emph{Haar} matches multiscale high-frequency Haar-wavelet coefficients of predicted and reference depth maps. These coefficients describe local depth changes, so matching them preserves sharp boundaries and penalizes misplaced edges without indiscriminately smoothing genuine detail. Together, Noman suppresses unsupported intermediate depths, while Haar constrains boundary transitions to reduce geometric artifacts associated with blurred or displaced edges. The complete objective is
\begin{equation}
\begin{aligned}
\mathcal L_{\mathrm{VAE}}={}&\mathcal L_{\mathrm{range}}
+\lambda_v\mathcal L_{\mathrm{valid}}
+\lambda_n\mathcal L_{\mathrm{noman}}
+\lambda_h\mathcal L_{\mathrm{Haar}}\\
&+\lambda_s\mathcal L_{\mathrm{normal}}
+\lambda_p\mathcal L_{\mathrm{perceptual}}
+\lambda_{\mathrm{KL}}\mathcal L_{\mathrm{KL}}.
\end{aligned}
\label{eq:lidar-vae-objective}
\end{equation}
Range reconstruction supervises the two depth channels, with greater weight on measured returns; noman, Haar, normal and depth-perceptual terms supervise their reconstructed geometry. Validity uses binary cross-entropy on the third channel's raw logits against $m$, while KL regularizes the latent distribution. 

\subsubsection{RGB-conditioned LiDAR diffusion}
\label{sec:method-lidar-dit}
A multiview diffusion transformer couples clean camera latents with noisy LiDAR latents. RGB observations remain fixed during sampling, while attention exchanges information across views, modalities and time to generate the LiDAR sequence. Native-grid packed attention concatenates separately patchified camera and LiDAR grids, avoiding the spatial padding required by a shared canvas. Ray features encode azimuth and camera coverage, alongside view identifiers, to distinguish sensor directions. Following Unified Latent Anchoring~\cite{zhao2026unidrivedreamer}, we align LiDAR latent statistics to the RGB-pretrained scale and invert this transform before decoding.

\paragraph{Geometry supervision through a frozen decoder.}
Latent prediction error does not directly capture whether decoded points lie on plausible surfaces. We therefore supervise the generator through the frozen LiDAR decoder, transferring the four geometric criteria used in representation learning to conditional generation:
\begin{equation}
\mathcal L_{\mathrm{gen}}
=\mathcal L_{\mathrm{RF}}
+w(\sigma)\!\sum_{k\in\{\mathrm{range},\mathrm{valid},\mathrm{noman},\mathrm{Haar}\}}
\alpha_k\,\mathcal L_k\!\left(D_\phi(A^{-1}(\hat z_0)),x\right).
\label{eq:lidar-generation-objective}
\end{equation}
Here $\mathcal L_{\mathrm{RF}}$ is the rectified-flow loss on LiDAR latents, $\hat z_0$ is the generator's clean-latent estimate, $A^{-1}$ reverses latent alignment, and $x$ is the target range/validity map. The weights $\alpha_k$ balance geometric terms, while $w(\sigma)$ controls their contribution across noise levels. Decoder parameters $\phi$ remain fixed, but gradients pass through decoding to update the DiT. Thus the decoder cannot adapt to absorb generation errors: the generator must produce latents whose decoded geometry better matches the target. Auxiliary supervision is used only during training and adds no loss evaluation at inference.

\paragraph{Radial refinement.}
\label{sec:method-lidar-rrn}
Following the 3D residual-regression approach of L3DR~\cite{liu2026l3dr}, a separately trained radial rectification network (RRN) uses point-cloud neighborhoods to adjust decoded points along their original sensor rays. Its temporal variant also incorporates neighboring predictions after ego-motion compensation; it is an offline refinement, not a causal streaming component. RRN adjusts existing returns rather than creating missing ones and requires no GT point cloud at inference. The interface can refine both VAE reconstructions and generated LiDAR, with results evaluated separately for each pipeline.

%% file: sections/experiments.tex
\section{Experiments}
\label{sec:experiments}

We evaluate HelloWorld in terms of generation quality, geometric
consistency, and controllability.
We first examine single-view trajectory-conditioned video generation at Stage 1,
followed by multi-view generation at Stage 2 and
Stage 3.
Additional experiments evaluate LiDAR generation, few-step
distillation, and controllability under different conditioning inputs.

\providecommand{\pendingmetric}{\textnormal{xx}}
\providecommand{\notapplicable}{\textnormal{N/A}}
\providecommand{\storycell}[2]{\fbox{\parbox[c][1.0cm][c]{#1}{\centering\scriptsize #2\\[3pt]\textcolor{gray}{xx}}}}

\subsection{Single-View Video Generation}
\label{sec:single_view_generation}

\paragraph{Experimental setup.}
We compare HelloWorld with HY-WorldPlay and Lingbot World 2.0
on the 29-frame and 100-frame benchmarks.
The evaluation measures whether generated videos follow the
prescribed camera trajectories while maintaining perceptual quality.

The short-video benchmark contains 100 clips, comprising
60 general-domain clips from DL3DV, Sekai, SpatialVid,
RealEstate10K, and OmniWorld, and 40 real-world driving clips.
Clips are selected through deterministic, scene-stratified sampling.
The driving subset covers highways, turning maneuvers, nighttime,
adverse weather, parking and residential areas, and daytime
urban roads.

We estimate camera trajectories from generated videos using a fixed
Pi3X estimator and compare them with the dataset reference trajectories.
Estimated rotation matrices are projected onto $\mathrm{SO}(3)$
before evaluation.
We report first-frame-relative geodesic rotation RMSE in degrees
(\textbf{Rot}) and translation RMSE after Umeyama $\mathrm{Sim}(3)$
alignment (\textbf{Trans}).
Both metrics are computed per clip and then averaged across clips.
On the real-world on-road subset, we additionally report HelloWorld's
metric-scale translation RMSE (\textbf{Trans}$^{*}$), which aligns
only the initial coordinate frame without fitting a translation scale.
All pose errors are lower-is-better.

For perceptual evaluation, we report eight VBench dimensions:
subject consistency, background consistency, motion smoothness,
dynamic degree, aesthetic quality, imaging quality, and
image-to-video subject and background consistency.
The two image-to-video dimensions measure consistency with the
reference image.
The reported Mean is the unweighted arithmetic mean of these eight
raw scores, rather than the full VBench benchmark score.

\paragraph{Trajectory adherence.}
Tables~\ref{tab:pose_29f} and~\ref{tab:pose_100f} show that HelloWorld
achieves the lowest rotation and Sim(3)-aligned translation errors
among the compared methods on both benchmarks.
On the 29-frame benchmark, HelloWorld achieves a rotation error of
$2.0620^\circ$ and a translation error of $0.2570$,
reducing the corresponding errors of Lingbot World 2.0 by
$36.2\%$ and $9.9\%$, respectively.
On the 100-frame benchmark, HelloWorld obtains
$2.7056^\circ$ rotation error and $1.3997$ translation error,
corresponding to reductions of $50.1\%$ and $6.6\%$.
These results demonstrate more accurate adherence to the prescribed
camera trajectories under the shared evaluation pipeline.

HelloWorld's metric-scale translation errors on the real-world
driving cases are $0.6830$\,m and $5.5857$\,m on the two benchmarks.
These measurements complement the scale-aligned errors by retaining
sensitivity to displacement scale.
Since corresponding baseline measurements are not reported,
they do not establish comparative superiority in metric-scale
translation accuracy.

\input{tables/sv-29f-pose}
\input{tables/sv-100f-pose}

\paragraph{Perceptual quality.}
Tables~\ref{tab:vbench_29f} and~\ref{tab:vbench_100f} show that
HelloWorld's improved trajectory adherence coexists with competitive,
but not uniformly superior, perceptual scores.
HelloWorld achieves eight-dimension means of $0.8328$ and $0.8229$,
compared with $0.8411$ and $0.8416$ for Lingbot World 2.0.
HelloWorld obtains higher motion smoothness on both benchmarks and
higher dynamic degree on the 29-frame benchmark.
On the 100-frame benchmark, it also achieves slightly higher
subject and background consistency.

HY-WorldPlay obtains strong consistency and motion-smoothness scores,
but substantially lower dynamic-degree scores of $0.2800$ and
$0.1400$.
This combination highlights the need to interpret consistency jointly
with motion and trajectory adherence.
HelloWorld maintains a dynamic-degree score of $0.9200$ on both
benchmarks while achieving the best reported pose errors.
Its lower imaging-quality scores relative to Lingbot World 2.0
identify visual fidelity as a remaining limitation.

\input{tables/sv-29f-vbench}
\input{tables/sv-100f-vbench}

\subsection{Multi-View Video Generation}
\label{sec:multi_view_generation}

\paragraph{Experimental setup.}
We evaluate two training stages of HelloWorld on synchronized
multi-camera generation.
The mid-trained model uses pose conditioning without additional
scene-control inputs, whereas the post-trained model jointly uses
pose and scene-control inputs.
For each model, we evaluate generation with and without first-frame
image conditioning, denoted by \emph{w/ ff} and \emph{w/o ff}.
Removing first-frame conditioning preserves the other conditioning
inputs associated with each model.
We compare against Cosmos transfer 2.5 in the first-frame-conditioned
setting.

We use the pose metrics defined in
Section~\ref{sec:single_view_generation}.
For VBench, we report the six dimensions that do not require a
reference image: subject consistency, background consistency,
motion smoothness, dynamic degree, aesthetic quality, and imaging
quality. Their unweighted arithmetic mean is reported as Mean.

\paragraph{Cross-view consistency metric.}
We measure cross-view geometric consistency using
Cross-camera Sampson Error (\textbf{CSE}).
For synchronized images from cameras $a$ and $b$, we extract
correspondences using a frozen outdoor LoFTR matcher.
The fundamental matrix $F_{ab}$ is computed from clip-specific
camera intrinsics and extrinsics, with intrinsics adjusted to the
evaluation resolution.
For homogeneous matched points $\mathbf{x}_a$ and $\mathbf{x}_b$,
the pixel-valued Sampson distance is
\begin{equation}
d =
\frac{
    \left|\mathbf{x}_b^\top F_{ab}\mathbf{x}_a\right|
}{
    \sqrt{
    \|(F_{ab}\mathbf{x}_a)_{1:2}\|_2^2
    +
    \|(F_{ab}^{\top}\mathbf{x}_b)_{1:2}\|_2^2
    }
}.
\end{equation}

We restrict correspondences to calibrated overlap regions with five-pixel boundary erosion and retain matches with confidence at least 0.5. For each camera-pair/frame combination, we compute the confidence-weighted mean of distances truncated at eight pixels. Combinations with fewer than 20 valid matches receive an error of eight pixels rather than being discarded. Scores are averaged with equal weights over frames, camera pairs, and clips. 
We additionally report $\Delta\mathrm{CSE} = \mathrm{CSE}_{\mathrm{gen}}-\mathrm{CSE}_{\mathrm{GT}}$, using real videos evaluated with the same pipeline as the reference.

CSE evaluates epipolar agreement rather than complete 3D consistency:
depth discrepancies that preserve the epipolar constraint may remain
undetected.

\paragraph{Trajectory control.}
Table~\ref{tab:mv_pose} shows that the first-frame-conditioned post-trained model achieves the lowest rotation and Sim(3)-aligned translation errors among the evaluated settings. Compared with Cosmos transfer 2.5, it reduces rotation error from $5.534^\circ$ to $2.027^\circ$ and translation error from $1.880$ to $1.352$, improvements of $63.4\%$ and $28.1\%$,
respectively. Metric-scale translation error decreases from $8.886$\,m to $4.904$\,m, a $44.8\%$ reduction.

Relative to the first-frame-conditioned mid-trained model, the post-trained model reduces rotation and aligned translation errors by $38.9\%$ and $11.3\%$. However, metric-scale translation error increases from $4.486$\,m to $4.904$\,m. The same pattern holds without first-frame conditioning:
rotation and aligned translation errors improve, while metric-scale translation error increases from $7.972$\,m to $8.485$\,m. Thus, improved orientation and scale-aligned trajectory accuracy do not imply uniformly improved metric-scale displacement accuracy.

\begin{table}[t]
\centering
\footnotesize
\caption{Pose comparison against ground truth on multiview benchmark.}
\label{tab:mv_pose}
\begin{tabular}{lrrr}
\toprule
Setting & Rot ($^\circ$)$\downarrow$ & Trans$\downarrow$ & Trans$^{*}\downarrow$ \\
\midrule
Cosmos Transfer 2.5 (w/ ff) & 5.534 & 1.880 & 8.886 \\
\midrule
Stage 2 (w/ ff)    & 3.315 & 1.525 & \textbf{4.486} \\
Stage 2 (w/o ff) & 5.163 & 2.189 & 7.972 \\
Stage 3 (w/ ff)    & \textbf{2.027} & \textbf{1.352} & 4.904 \\
Stage 3 (w/o ff) & 3.334 & 1.844 & 8.485 \\
\bottomrule
\end{tabular}
\end{table}

\paragraph{Cross-view consistency.}
Table~\ref{tab:mv_cse} shows improved cross-view epipolar consistency
for the post-trained model under both initialization settings.
With first-frame conditioning, CSE decreases from $6.192$ to
$4.180$, a $32.5\%$ reduction relative to the mid-trained model.
Without first-frame conditioning, it decreases from $7.509$ to
$4.892$, a $34.9\%$ reduction.

Compared with Cosmos transfer 2.5, the first-frame-conditioned
post-trained model reduces CSE from $6.549$ to $4.180$,
an improvement of $36.2\%$.
Its excess error over the real-video reference decreases from
$4.819$ to $2.450$, a $49.2\%$ reduction.
Even without first-frame conditioning, the post-trained model
obtains lower CSE than the first-frame-conditioned Cosmos baseline.
All generated settings nevertheless remain above the real-video
reference of $1.730$, indicating a remaining gap in cross-view
geometric consistency.

\begin{table}[t]
\centering
\footnotesize
\caption{Multiview CSE.}
\label{tab:mv_cse}
\begin{tabular}{lrr}
\toprule
Setting & CSE$\downarrow$ & $\Delta$CSE$\downarrow$ \\
\midrule
GT video            & 1.730 & 0.000 \\
Cosmos Transfer 2.5 (w/ ff) & 6.549 & +4.819 \\
\midrule
Stage 2 (w/ ff) & 6.192 & +4.462 \\
Stage 2 (w/o ff) & 7.509 & +5.779 \\
Stage 3 (w/ ff) & 4.180 & +2.450 \\
Stage 3 (w/o ff) & 4.892 & +3.162 \\
\bottomrule
\end{tabular}
\end{table}

\paragraph{Perceptual quality.}
Table~\ref{tab:mv_vbench} shows that all four HelloWorld settings
obtain higher six-dimension means than Cosmos transfer 2.5.
The post-trained model without first-frame conditioning achieves
the highest mean of $0.7901$, compared with $0.7722$ for Cosmos.
Under matched first-frame conditioning, the post-trained model
improves subject consistency, background consistency, and imaging
quality over Cosmos, while aesthetic quality and dynamic degree
are lower and motion smoothness is nearly unchanged.
These results indicate a mixed perceptual-quality trade-off
alongside the improvements in trajectory control and cross-view
consistency.

\begin{table}[t]
\centering
\footnotesize
\caption{VBench comparison on the multiview benchmark.}
\label{tab:mv_vbench}
\begin{tabular}{lrrrrrrr}
\toprule
Setting & Subject & Background & Motion & Dynamic & Aesthetic & Imaging & Mean \\
\midrule
Cosmos Transfer 2.5 (w/ ff) & 0.8916 & 0.9376 & 0.9878 & 0.9555 & 0.4366 & 0.4239 & 0.7722 \\
Stage 2 (w/ ff)    & 0.8999 & 0.9324 & 0.9843 & 0.9800 & 0.4374 & 0.4953 & 0.7882 \\
Stage 2 (w/o ff) & 0.8716 & 0.9199 & 0.9822 & 0.9700 & 0.4466 & 0.4844 & 0.7791 \\
Stage 3 (w/ ff)    & 0.9089 & 0.9462 & 0.9876 & 0.9500 & 0.4138 & 0.4456 & 0.7754 \\
Stage 3 (w/o ff) & 0.9111 & 0.9472 & 0.9788 & 0.9700 & 0.4259 & 0.5074 & \textbf{0.7901} \\
\bottomrule
\end{tabular}
\end{table}

\paragraph{Effect of first-frame conditioning.}
First-frame conditioning improves all three pose metrics and CSE
for both training stages.
For the post-trained model, it reduces rotation error from
$3.334^\circ$ to $2.027^\circ$, aligned translation error from
$1.844$ to $1.352$, metric-scale translation error from
$8.485$\,m to $4.904$\,m, and CSE from $4.892$ to $4.180$.
These results support the role of image initialization in
constraining generated geometry.

Its effect on perceptual quality is less uniform.
First-frame conditioning increases the mid-trained model's
VBench mean from $0.7791$ to $0.7882$, but decreases the
post-trained model's mean from $0.7901$ to $0.7754$.
The latter difference includes lower aesthetic and imaging-quality
scores despite improved geometric accuracy.
This discrepancy demonstrates that the perceptual mean alone
does not capture trajectory adherence or cross-view consistency.

\paragraph{Public driving benchmark.}
\label{sec:exp-nuscenes}
Table~\ref{tab:nuscenes} compares driving generation at short and long
temporal horizons using FID and FVD. FID measures frame-level visual
fidelity, while FVD evaluates the distributional similarity of generated
and real videos in a spatiotemporal feature space. Lower values indicate
better distributional fidelity for both metrics. The capability columns
identify autoregressive generation, multi-view output, video generation,
and whether dense supervision is required, distinguishing methods with
different output formats and supervision requirements.

For short-horizon generation, HelloWorld obtains an FID of 7.75 and an
FVD of 41.08. Among the autoregressive multi-view methods listed in the
table, FAR-Drive provides a reference with an FID of 11.92 and an FVD
of 82.78. Relative to FAR-Drive, HelloWorld reduces FID by 35.0\% and
FVD by 50.4\%. Considering both metrics helps distinguish improvements
in individual-frame appearance from changes in overall video quality.
Image-generation baselines are included for FID comparison only, since
FVD is not applicable to their image-only outputs.

For long-horizon generation, HelloWorld achieves an FID of 17.81 and
an FVD of 88.10, compared with 20.91/94.84 for MagicDrive-V2 and
13.82/92.99 for HorizonDrive. Relative to HorizonDrive, HelloWorld has
a 28.9\% higher FID and a 5.3\% lower FVD, indicating a trade-off between
frame-level and video-level distributional quality. 

\input{tables/nuscenes-short-long}

\FloatBarrier
\subsection{Controllability Analysis}
\label{sec:exp-control}
\begin{figure}[htbp]
    \centering
    \includegraphics[width=1.0\linewidth]{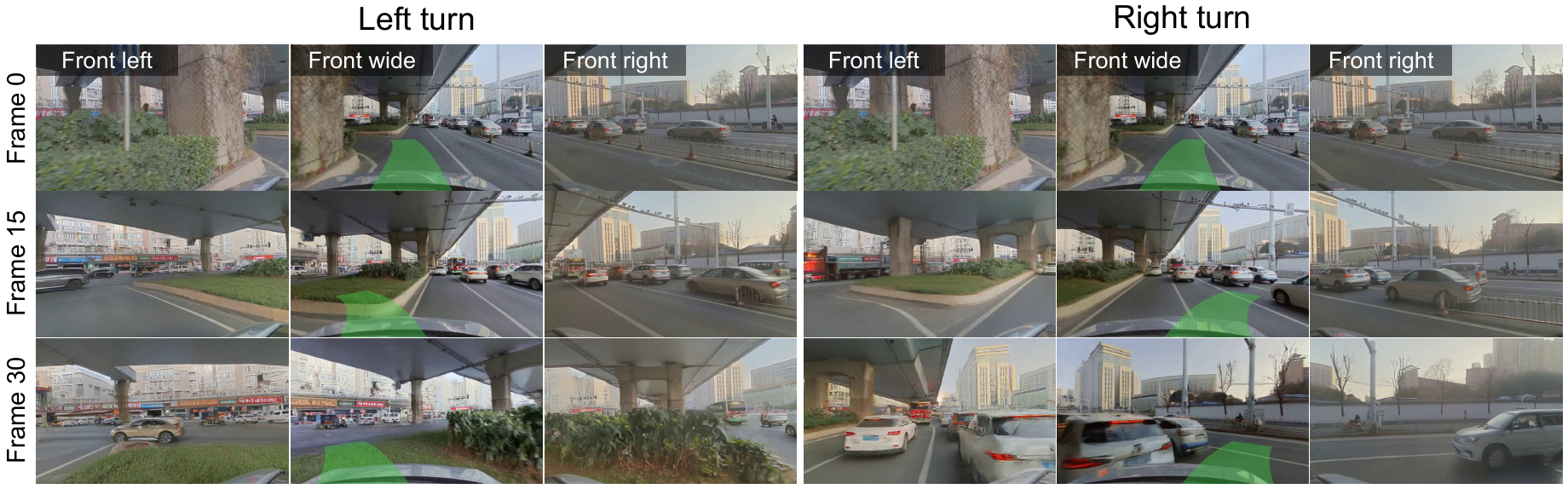}
    \caption{Counterfactual pose control results. Given the same scene context, our model generates plausible future observations conditioned on counterfactual left-turn and right-turn ego-pose trajectories, demonstrating flexible and geometrically coherent control over ego motion.}
    \label{fig:pose-control}
\end{figure}

\paragraph{Pose Control.} We assess pose controllability by conditioning the model on counterfactual future ego trajectories. Given the same scene prefix, we specify alternative pose sequences for left-turn and right-turn maneuvers that differ from the nominal future trajectory of the original sequence. As illustrated in \Cref{fig:pose-control}, the generated videos exhibit viewpoint transitions that are consistent with the prescribed pose conditions, while preserving realistic scene geometry and traffic dynamics. Importantly, these trajectories are not standard continuations of the observed motion, but counterfactual interventions on ego motion. The results therefore show that our model can go beyond deterministic future prediction and instead synthesize geometrically coherent counterfactual futures, enabling explicit simulation of alternative driving behaviors.

\begin{figure}[htbp]
    \centering
    \includegraphics[width=1.0\linewidth]{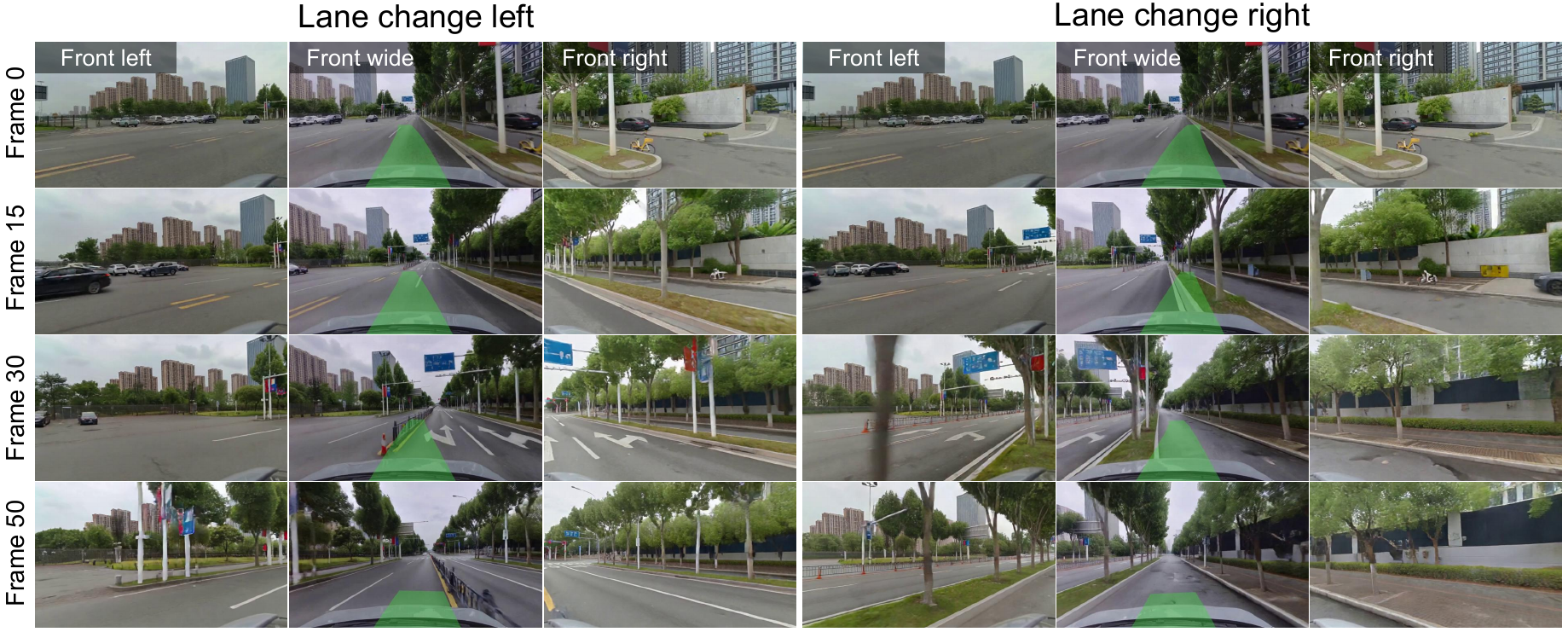}
    \caption{Unconventional lane-change control. We condition the model on two unconventional lane-change trajectories: a rightward lane change that crosses the roadside green belt into the non-motorized lane, and a leftward lane change that crosses the barrier into the opposite-direction lane. The generated results remain consistent with the prescribed pose conditions while preserving coherent scene structure.}
    \label{fig:lane-change}
\end{figure}

In addition to regular left-turn and right-turn controls, we further evaluate two extreme counterfactual lane-change trajectories. In particular, the rightward trajectory crosses the roadside green belt into the non-motorized lane, whereas the leftward trajectory crosses the road barrier into the opposite-direction lane. The generated results in \Cref{fig:lane-change} show that the model can still follow these unconventional pose interventions and produce corresponding viewpoint changes while preserving coherent scene structure. This further verifies the flexibility of the proposed pose control under highly unusual ego-motion conditions.

\begin{figure}[htbp]
    \centering
    \includegraphics[width=1.0\linewidth]{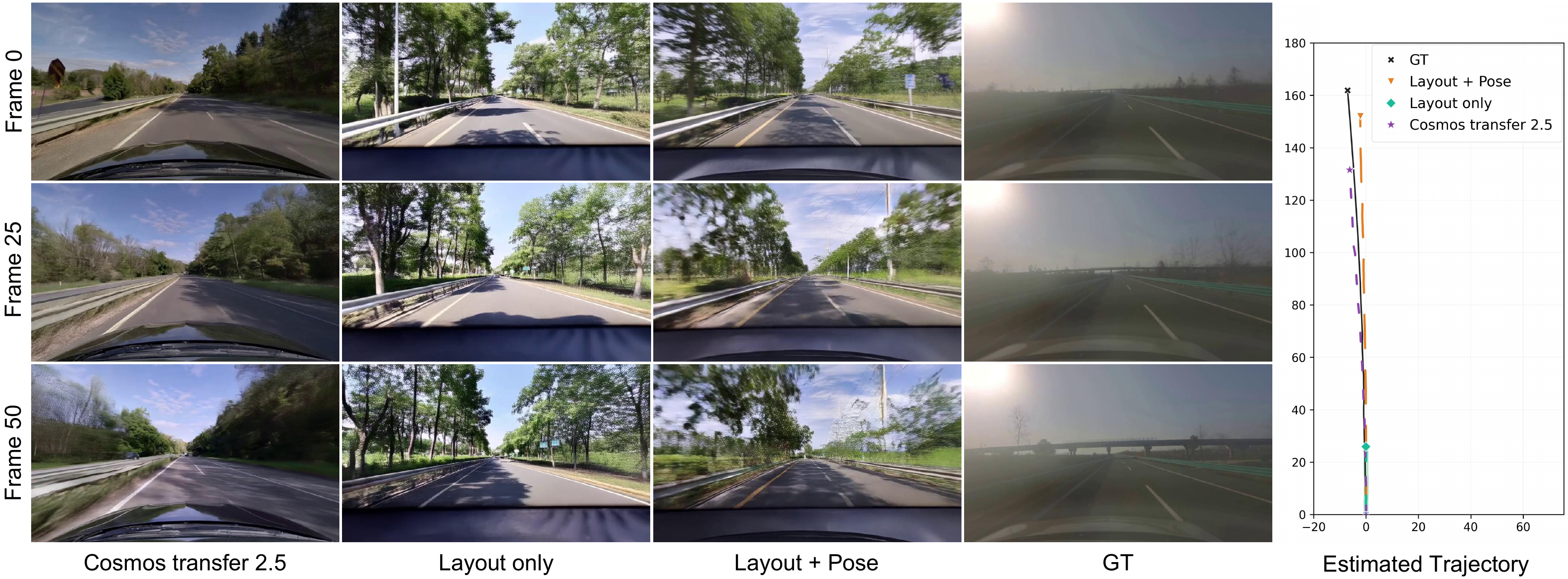}
    \caption{Comparison of scale control under different conditioning inputs. Our method supports both control video only and control video + pose. Compared with Cosmos transfer 2.5 and the control-video-only variant, adding pose leads to more accurate geometry, better scale consistency, and trajectories closer to the ground truth. Camera poses are estimated using Pi3X.}
    \label{fig:pose-layout}
\end{figure}

\paragraph{Scale Control with Pose Conditioning.}
We study whether explicit pose input can improve geometric consistency and scale controllability in video generation. Our framework supports two input modes: control video only and control video together with pose. We compare these two variants with Cosmos transfer 2.5, which does not take pose as input. As illustrated in \Cref{fig:pose-layout}, the control-video-only setting already captures the rough structure of the target scene, showing that the control video provides useful layout guidance. However, its estimated trajectory still departs from the ground truth, indicating limited control over ego-motion scale. By contrast, when pose is additionally provided, the generated sequence exhibits substantially better alignment with the ground-truth observations, both visually and in the recovered trajectory. In particular, the Layout + Pose trajectory remains consistently closer to the ground truth than the other two baselines, demonstrating that explicit pose conditioning effectively reduces scale drift and improves geometric fidelity during generation.

\begin{figure}[htbp]
    \centering
    \includegraphics[width=1.0\linewidth]{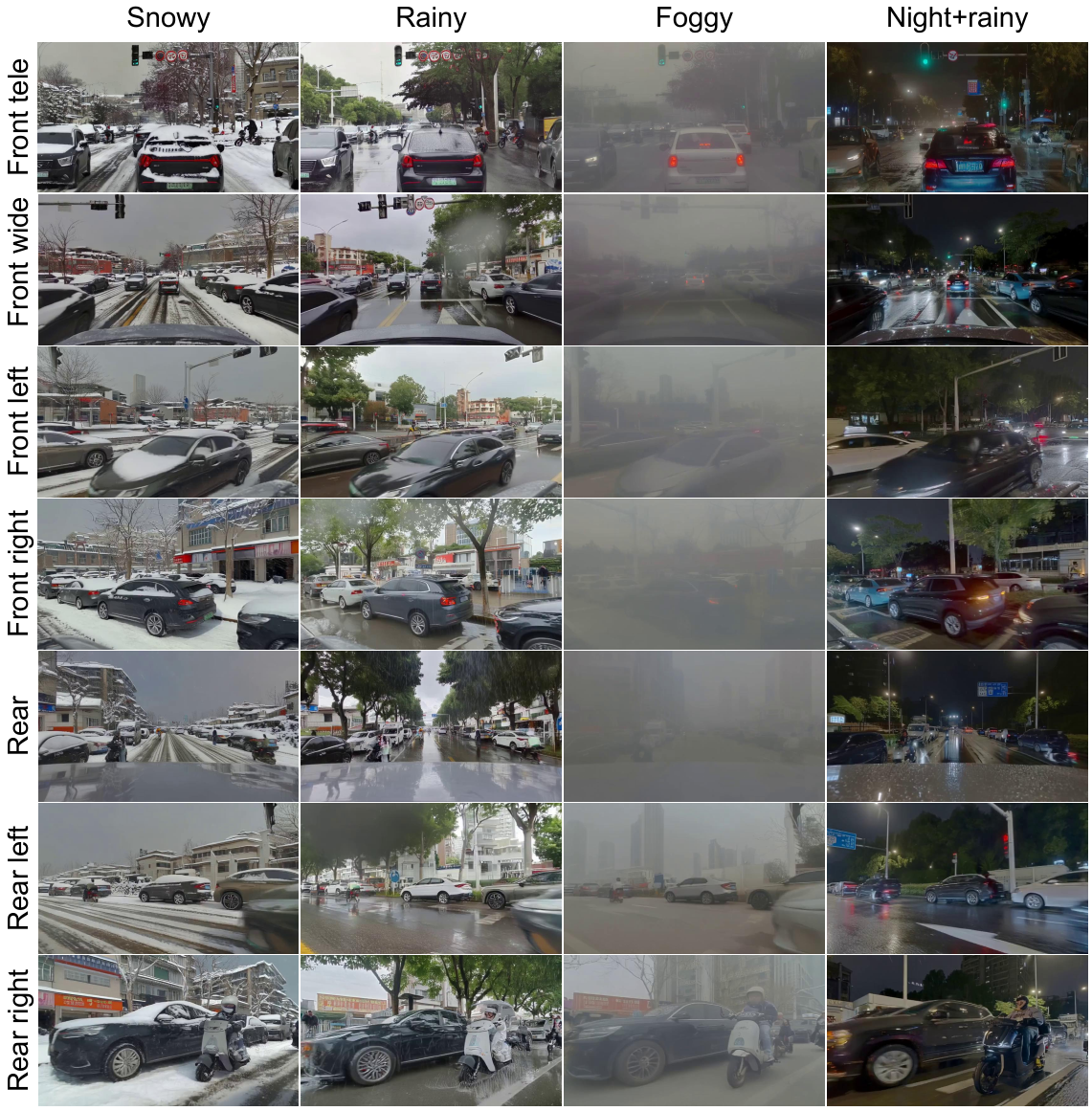}
    \caption{Given the same underlying driving scenes, our model generates diverse weather conditions, including snowy, rainy, foggy, night-and-rainy, while preserving consistent scene structure and traffic semantics.}
    \label{fig:weather}
\end{figure}

\begin{figure}[htbp]
    \centering
    \includegraphics[width=1.0\linewidth]{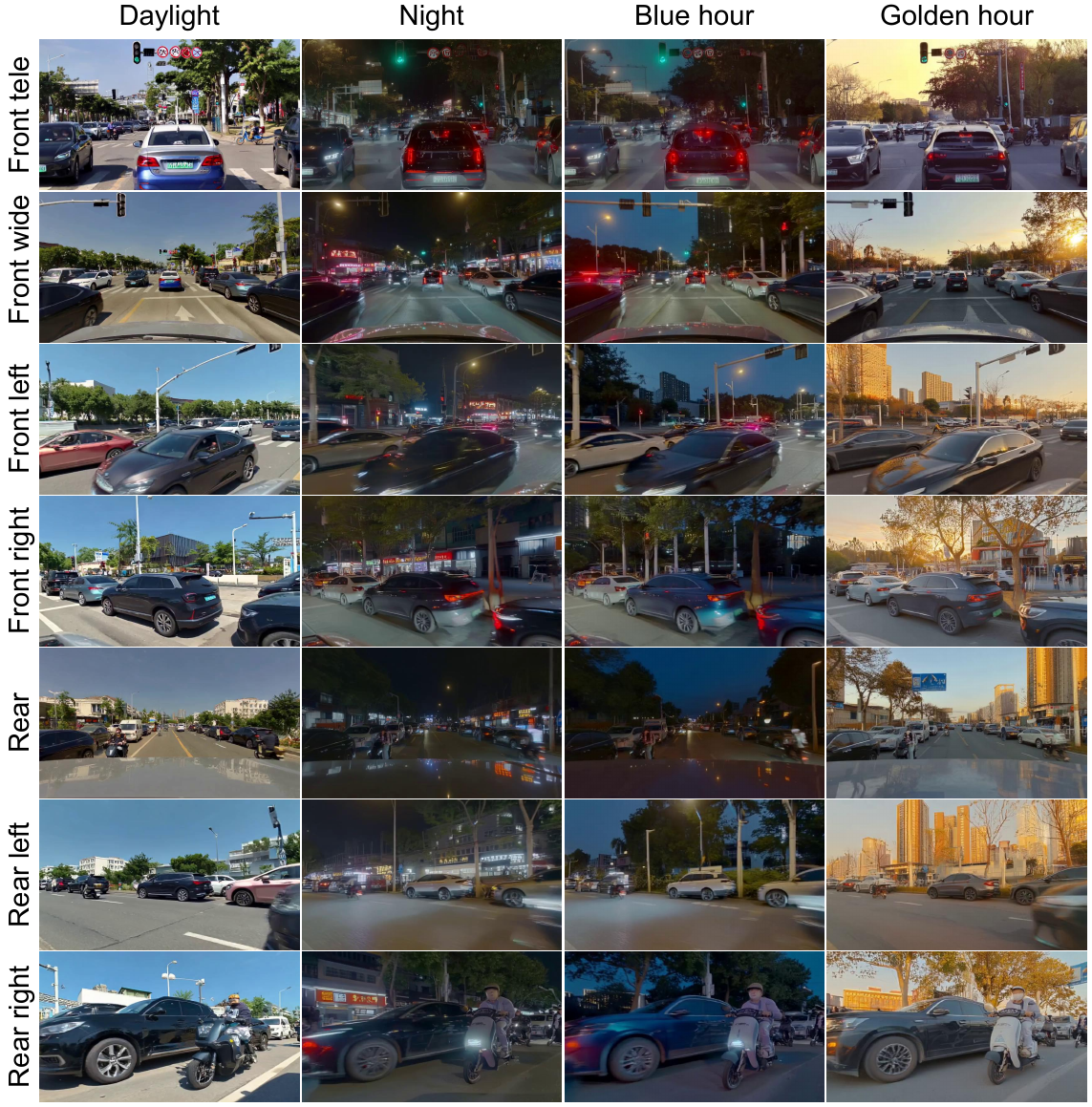}
    \caption{Given the same underlying driving scenes, our model generates diverse weather conditions, including daylight, night, golden-hour, and blue-hour scenarios, while preserving consistent scene structure and traffic semantics.}
    \label{fig:lighting}
\end{figure}

\paragraph{Environment Control.}
We further evaluate the model's controllability over diverse environmental conditions, including weather and illumination. As shown in \Cref{fig:weather,fig:lighting}, given similar driving contexts, the model can synthesize visually consistent scenes under a wide range of conditions, including snowy, rainy, night, night-and-rainy, golden-hour, and blue-hour scenarios. The generated results exhibit condition-specific appearance changes, such as snow coverage, wet road surfaces and reflections, low-light illumination, and characteristic color tones at different times of day, while largely preserving the underlying road structure, traffic layout, and scene semantics. Notably, the model also supports compositional control, as demonstrated by the night-and-rainy condition, where both nighttime illumination and rainy-weather effects are simultaneously reflected in the generated observations. These results demonstrate that the model can generate diverse and plausible driving environments while maintaining scene-level consistency.

\begin{figure}[t]
    \centering
    \includegraphics[width=1.0\linewidth]{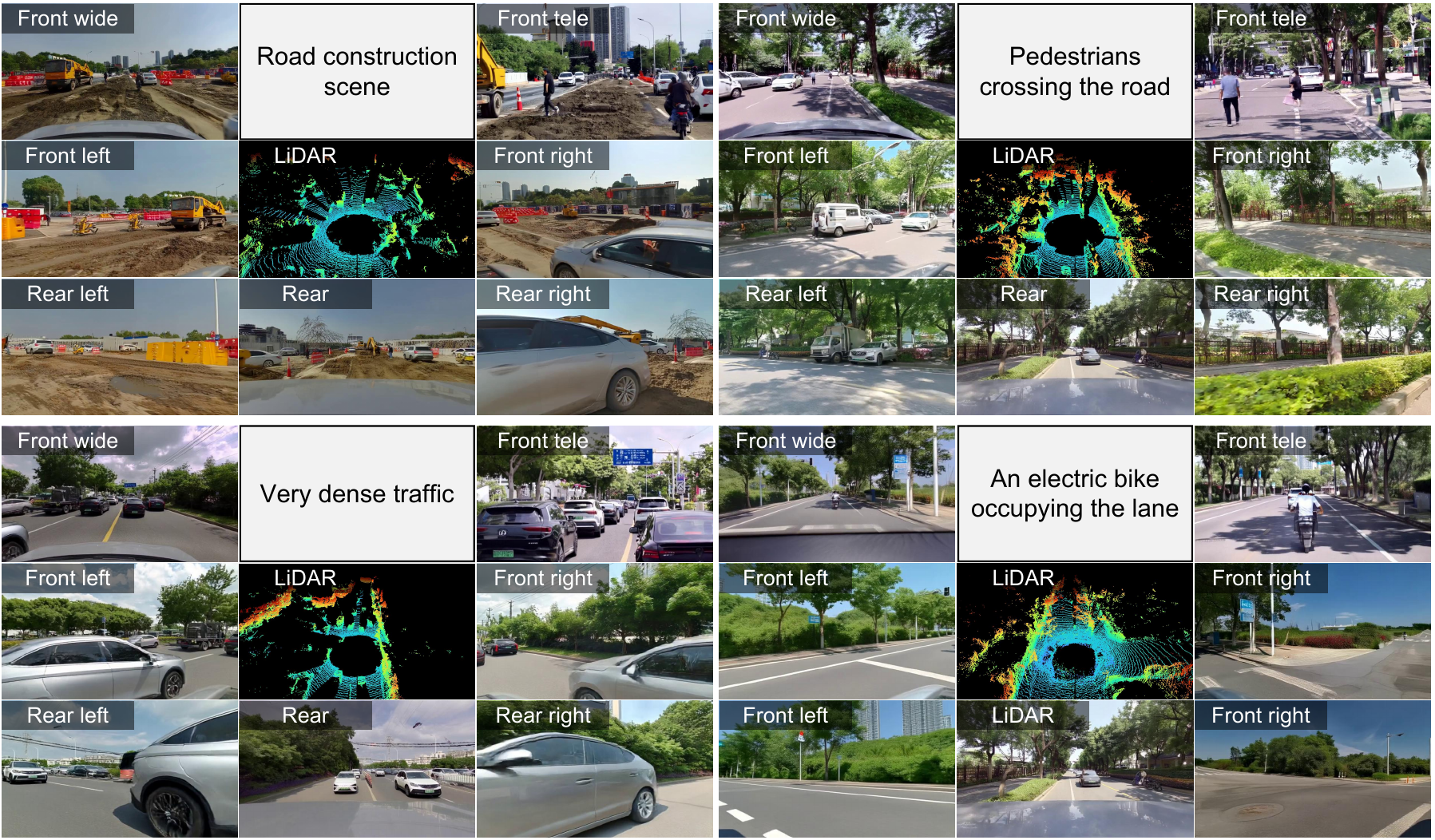}
    \caption{Special scene control results. Our model generates plausible long-tail traffic scenarios conditioned on special scene descriptions, demonstrating controllable synthesis of rare and safety-critical events while maintaining coherent scene structure and traffic context.}
    \label{fig:special-scenes}
\end{figure}

\paragraph{Special Scene Control.}
We further evaluate the model on special traffic scenes, which correspond to rare or long-tail scenarios that are less frequently observed in standard driving data. As shown in \Cref{fig:special-scenes}, the model can generate plausible future observations conditioned on scene-level semantic descriptions of unusual events, such as traffic accidents, abnormal road occupancy, or uncommon traffic participants. Compared with regular driving scenes, these cases require the model to capture more distinctive semantic cues and maintain consistency between the specified event and the surrounding road context. The generated results show that the model is able to synthesize semantically aligned special scenes while preserving coherent scene geometry, traffic layout, and visual realism. This suggests that the model is not limited to frequent driving patterns, but can also support controllable generation of long-tail and safety-critical scenarios, which is important for stress-testing autonomous driving systems.

\FloatBarrier
\subsection{Few-step Distillation}
The post-trained seven-view generator is a 20-step rectified-flow (RF) teacher with classifier-free guidance $\mathrm{CFG}{=}3$.
We distill it into a 4-step student with $\mathrm{CFG}{=}1$ in two stages.
dCM first trains a consistency student from the RF teacher.
DMD then continues from that dCM student: the RF model remains the frozen teacher and fake-score network, and the student is trained with score-matching under the same causal KV rollout used at inference.
At test time both students use the distilled 4-step schedule and a single conditional forward (no CFG branch).

Quality is measured on a frozen 100-clip set of 100-frame videos, with and without the first-frame condition (w/ ff, w/o ff).
Pose (Pi3X) and VBench use the front-wide camera; cross-view consistency (CSE) uses all seven cameras.
Inference cost is measured on eleven 720p corner clips ($1280{\times}720$, 61 frames, 10\,fps, first-frame condition) on 8~GPUs with context-parallel size 8.
The first clip is warmup and is excluded from the means.

Table~\ref{tab:pi3x} and Table~\ref{tab:cse} separate rotation and cross-view error from translation.
With a first frame, dCM increases rotation from $2.027^\circ$ to $2.568^\circ$ and CSE from $4.180$ to $5.201$.
Without a first frame both gaps widen ($3.334^\circ\!\to\!4.619^\circ$, $4.892\!\to\!6.399$).
Translation does not follow rotation: dCM improves $\mathrm{Trans}^*$ in both settings ($4.904\!\to\!2.862$, $8.485\!\to\!5.311$) and improves Trans with a first frame ($1.352\!\to\!1.263$), while Trans without a first frame gets worse ($1.844\!\to\!2.907$).
DMD with a first frame is the best row on Rot, Trans, $\mathrm{Trans}^*$, and CSE.
Without a first frame it still beats the teacher on the same condition ($2.311^\circ$, $1.034$, $3.193$, CSE $4.223$).

\begin{table}[H]
\centering\small
\caption{Multiview generation pose errors of few-step distilled models.}
\label{tab:pi3x}
\small
\begin{tabular}{lccc}
\toprule
Setting & Rot ($^\circ$)$\downarrow$ & Trans$\downarrow$ & Trans$^*$$\downarrow$ \\
\midrule
Baseline (w/ ff)  & 2.027 & 1.352 & 4.904 \\
Baseline (w/o ff) & 3.334 & 1.844 & 8.485 \\
\midrule
dCM (w/ ff)       & 2.568 & 1.263 & 2.862 \\
dCM (w/o ff)      & 4.619 & 2.907 & 5.311 \\
\midrule
DMD (w/ ff)       & \textbf{1.661} & \textbf{0.688} & \textbf{1.885} \\
DMD (w/o ff)      & 2.311 & 1.034 & 3.193 \\
\bottomrule
\end{tabular}
\end{table}

\begin{table}[H]
\centering\small
\caption{Multiview CSE of few-step distilled models.}
\label{tab:cse}
\small
\begin{tabular}{lcc}
\toprule
Setting & CSE$\downarrow$ & $\Delta$CSE$\downarrow$ \\
\midrule
GT video          & 1.730 & 0.000 \\
\midrule
Baseline (w/ ff)  & 4.180 & +2.450 \\
Baseline (w/o ff) & 4.892 & +3.162 \\
\midrule
dCM (w/ ff)       & 5.201 & +3.471 \\
dCM (w/o ff)      & 6.399 & +4.669 \\
\midrule
DMD (w/ ff)       & \textbf{3.510} & \textbf{+1.780} \\
DMD (w/o ff)      & 4.223 & +2.493 \\
\bottomrule
\end{tabular}
\end{table}

Table~\ref{tab:vbench} separates appearance from geometry.
dCM lowers dynamic degree ($0.9500\!\to\!0.8800$, $0.9700\!\to\!0.7600$) and imaging quality ($0.4456\!\to\!0.4281$, $0.5074\!\to\!0.3824$), and the mean falls to $0.7580$ / $0.7316$.
Motion smoothness stays near $0.98$ for every row; the teacher with a first frame remains highest at $0.9876$.
DMD brings the mean back above the teacher ($0.7906$ / $0.7938$ vs.\ $0.7754$ / $0.7901$) and is best on dynamic degree ($0.9900$ / $1.0000$).
Subject and background improve only without a first frame ($0.9209$, $0.9488$).
Aesthetic and imaging without a first frame stay with the teacher ($0.4259$, $0.5074$).

\begin{table}[H]
\centering
\caption{VBench metrics of few-step distilled models on multiview benchmark.}
\footnotesize
\label{tab:vbench}
\begin{tabular}{lccccccc}
\toprule
Setting & Subject & Background & Motion & Dynamic & Aesthetic & Imaging & Mean \\
\midrule
Baseline (w/ ff)  & 0.9089 & 0.9462 & \textbf{0.9876} & 0.9500 & 0.4138 & 0.4456 & 0.7754 \\
Baseline (w/o ff) & 0.9111 & 0.9472 & 0.9788 & 0.9700 & \textbf{0.4259} & \textbf{0.5074} & 0.7901 \\
\midrule
dCM (w/ ff)       & 0.8961 & 0.9409 & 0.9855 & 0.8800 & 0.4176 & 0.4281 & 0.7580 \\
dCM (w/o ff)      & 0.9051 & 0.9469 & 0.9866 & 0.7600 & 0.4086 & 0.3824 & 0.7316 \\
\midrule
DMD (w/ ff)       & 0.9029 & 0.9430 & 0.9829 & 0.9900 & 0.4210 & 0.5039 & 0.7906 \\
DMD (w/o ff)      & \textbf{0.9209} & \textbf{0.9488} & 0.9826 & \textbf{1.0000} & 0.4171 & 0.4935 & \textbf{0.7938} \\
\bottomrule
\end{tabular}%
\end{table}

Table~\ref{tab:distill-cost} compares the RF teacher (20-step, $\mathrm{CFG}{=}3$) with the DMD student (4-step, $\mathrm{CFG}{=}1$) at 720p.
DiT ODE time drops from $305.3\,\mathrm{s}$ to $30.8\,\mathrm{s}$ ($9.91\times$), in line with the tenfold reduction in ODE forwards.
Sampling time falls only from $357.7\,\mathrm{s}$ to $77.8\,\mathrm{s}$ ($4.60\times$), because VAE encoding and decoding take about $32\,\mathrm{s}$ on both sides.
End-to-end time improves by $2.98\times$, with essentially matched memory.
dCM uses the same 4-step, $\mathrm{CFG}{=}1$ sampler as DMD, so we report cost on the DMD student.

\begin{table}[H]
\centering
\caption{Inference cost of the RF teacher and the DMD student.}
\label{tab:distill-cost}
\small
\begin{tabular}{lrrc}
\toprule
Phase & Teacher & DMD student & T/S \\
\midrule
Text encode (s)           & 2.5  & 2.5           & $1.0\times$ \\
VAE encode RGB (s)        & 6.7  & 7.1           & $0.95\times$ \\
VAE encode control (s)    & 13.6 & 13.6          & $1.00\times$ \\
VAE decode (s)            & 11.9 & 11.9          & $1.00\times$ \\
DiT prefix (s)            & 1.1  & 1.7           & $0.68\times$ \\
DiT ODE (s)               & 305.3 & \textbf{30.8} & $\mathbf{9.91\times}$ \\
DiT KV refresh (s)        & 12.1 & 6.0           & $2.02\times$ \\
Cache flush (s)           & 3.7  & 2.3           & -- \\
Sampling wall (s)         & 357.7 & \textbf{77.8} & $\mathbf{4.60\times}$ \\
Video write (s)           & 68.1 & 64.5          & -- \\
End-to-end (s)            & 426.7 & \textbf{143.2} & $2.98\times$ \\
Peak allocated (GiB)      & 61.6 & 61.9          & -- \\
nvidia-smi peak (GiB)     & 73.9 & 73.3          & -- \\
\bottomrule
\end{tabular}
\end{table}

In short: dCM reaches a 4-step sampler but increases rotation and CSE; DMD keeps that sampler and improves pose and CSE beyond the teacher on the matched condition, with a slightly higher VBench mean, matched memory, a $9.9\times$ faster DiT ODE, and $4.60\times$ faster sampling.

\FloatBarrier
\subsection{LiDAR Reconstruction and Conditional Generation}
\label{sec:exp-extensions}
\input{sections/lidar_experiments}

\subsection{Super-Resolution Refinement}
\begin{figure}[t]
    \centering
    \includegraphics[width=1.0\linewidth]{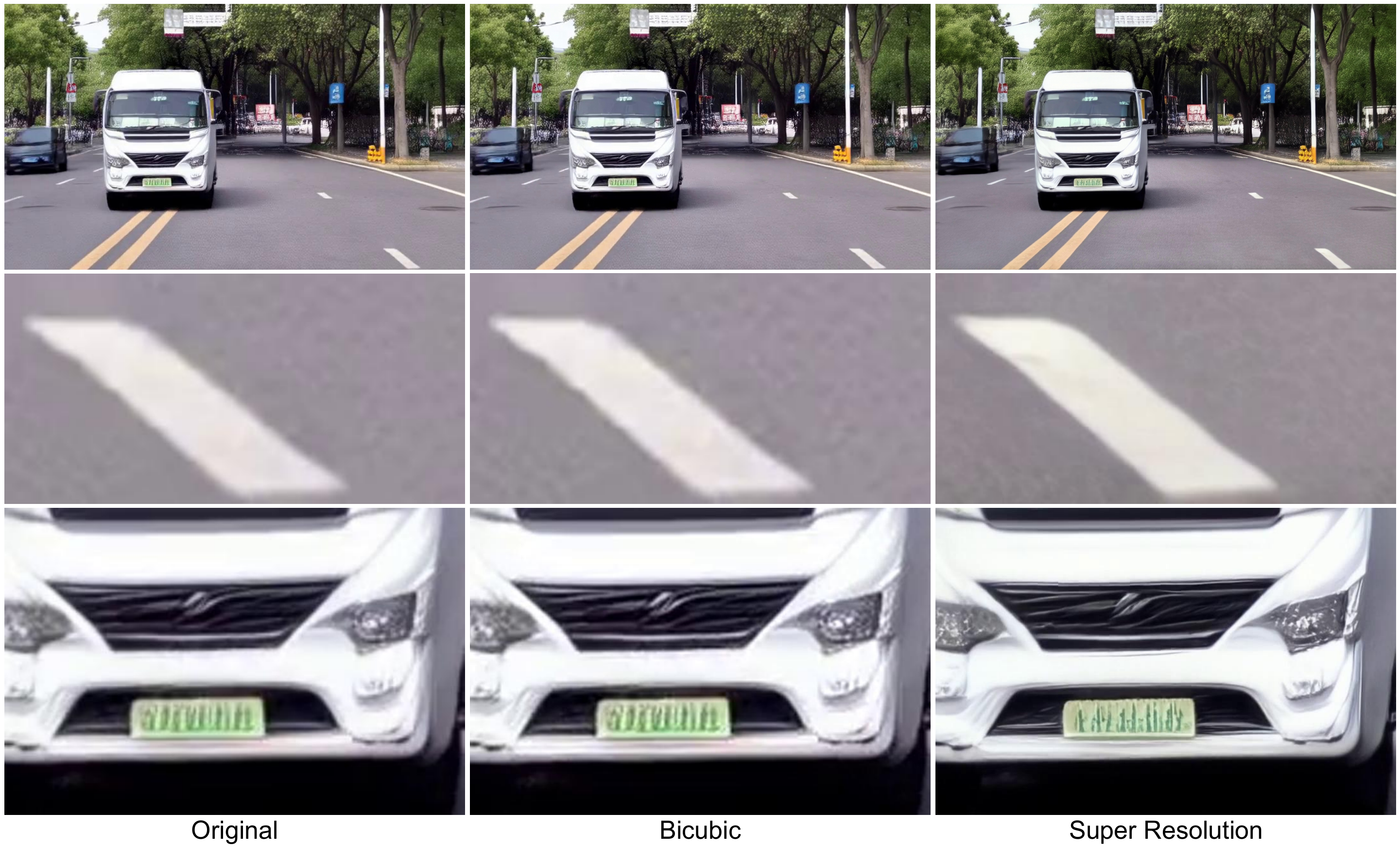}
    \caption{Super-resolution enhancement. Our super-resolution module produces sharper boundaries and richer local details than bicubic upsampling, as highlighted by the zoomed-in lane-marking and vehicle regions.}
    \label{fig:super-resulution}
\end{figure}

To further improve the visual fidelity of the generated videos, we employ a super-resolution module on the generated frames. Compared with bicubic interpolation, the proposed super-resolution stage better restores high-frequency details and sharp structural boundaries. As illustrated in \Cref{fig:super-resulution}, fine-scale elements such as lane markings, vehicle fronts, headlights, and license-plate regions become noticeably clearer after super-resolution. Importantly, the enhancement preserves the global scene layout and object structure while improving local appearance quality, making the generated videos more suitable for high-resolution visualization and downstream analysis.

\FloatBarrier

%% file: tables/sv-29f-pose.tex
\begin{table*}[t]
\centering
\footnotesize
\caption{Pose comparison against ground truth on the 29-frame single view benchmark.
Rot and Trans use Sim(3) alignment.
Trans$^{*}$ denotes metric-scale translation error in meters,
reported only for HelloWorld on real-world on-road vehicle cases.
Lower is better for all metrics.}
\label{tab:pose_29f}
\begin{tabular}{lccc}
\toprule
Method & Rot $\downarrow$ & Trans $\downarrow$ & Trans$^{*}$ $\downarrow$ \\
\midrule
HY-WorldPlay & 8.6305 & 1.0669 & -- \\
Lingbot World 2.0 & 3.2310 & 0.2853 & -- \\
HelloWorld & \textbf{2.0620} & \textbf{0.2570} & 0.6830 \\
\bottomrule
\end{tabular}
\end{table*}

%% file: tables/sv-100f-pose.tex
\begin{table*}[t]
\centering
\footnotesize
\caption{Pose comparison against ground truth on the 100-frame single view benchmark.
Rot and Trans use Sim(3) alignment.
Trans$^{*}$ denotes metric-scale translation error in meters,
reported only for HelloWorld on real-world on-road vehicle cases.
Lower is better for all metrics.}
\label{tab:pose_100f}
\begin{tabular}{lccc}
\toprule
Method & Rot $\downarrow$ & Trans $\downarrow$ & Trans$^{*}$ $\downarrow$ \\
\midrule
HY-WorldPlay & 15.0835 & 5.4504 & -- \\
Lingbot World 2.0 & 5.4169 & 1.4988 & -- \\
HelloWorld & \textbf{2.7056} & \textbf{1.3997} & 5.5857 \\
\bottomrule
\end{tabular}
\end{table*}

%% file: tables/sv-29f-vbench.tex
\begin{table*}[t]
\centering
\footnotesize
\caption{VBench comparison on the 29-frame single view benchmark.
SC: subject consistency; BC: background consistency;
MS: motion smoothness; DD: dynamic degree;
AQ: aesthetic quality; IQ: imaging quality;
I2V-S/B: image-to-video subject/background consistency.
Mean is the unweighted arithmetic mean of the eight raw scores.}
\label{tab:vbench_29f}
\begin{tabular}{lccccccccc}
\toprule
Method & SC & BC & MS & DD & AQ & IQ & I2V-S & I2V-B & Mean \\
\midrule
HY-WorldPlay
& 0.9669 & 0.9543 & 0.9933 & 0.2800
& 0.4762 & 0.5774 & 0.9829 & 0.9847 & 0.7770 \\
Lingbot World 2.0
& 0.9150 & 0.9424 & 0.9760 & 0.8900
& 0.4819 & 0.5933 & 0.9623 & 0.9676 & \textbf{0.8411} \\
HelloWorld
& 0.9148 & 0.9364 & 0.9807 & 0.9200
& 0.4521 & 0.5392 & 0.9563 & 0.9632 & 0.8328 \\
\bottomrule
\end{tabular}%
\end{table*}

%% file: tables/sv-100f-vbench.tex
\begin{table*}[t]
\centering
\footnotesize
\caption{VBench comparison on the 100-frame single view benchmark.}
\label{tab:vbench_100f}
\begin{tabular}{lccccccccc}
\toprule
Method & SC & BC & MS & DD & AQ & IQ & I2V-S & I2V-B & Mean \\
\midrule
HY-WorldPlay
& 0.9354 & 0.9325 & 0.9951 & 0.1400
& 0.4685 & 0.5584 & 0.9844 & 0.9855 & 0.7500 \\
Lingbot World 2.0
& 0.8841 & 0.9270 & 0.9814 & 0.9900
& 0.4489 & 0.5816 & 0.9573 & 0.9624 & \textbf{0.8416} \\
HelloWorld
& 0.8876 & 0.9288 & 0.9842 & 0.9200
& 0.4453 & 0.4976 & 0.9571 & 0.9627 & 0.8229 \\
\bottomrule
\end{tabular}%
\end{table*}

%% file: tables/nuscenes-short-long.tex
\begin{table}[t]
\centering
\small
\setlength{\tabcolsep}{5pt}
\renewcommand{\arraystretch}{1.13}
\caption{Driving generation on nuScenes~\cite{caesar2020nuscenes}
at short and long horizons. AR: autoregressive generation;
MV: multi-view; DSF: dense-supervision-free.
Lower FID and FVD are better.}
\label{tab:nuscenes}
\begin{tabularx}{\linewidth}{@{}Xccccrr@{}}
\toprule
Method & AR & MV & Video & DSF & FID $\downarrow$ & FVD $\downarrow$ \\
\midrule
\multicolumn{7}{@{}l}{\textit{Short-horizon generation}} \\
\addlinespace[3pt]
DrivingGPT~\cite{chen2025drivinggpt}
& \xmark & \xmark & \cmark & \cmark & 12.78 & 142.61 \\
DrivingWorld~\cite{hu2024drivingworld}
& \xmark & \xmark & \cmark & \cmark & 7.40 & 90.90 \\
Vista~\cite{gao2024vista}
& \xmark & \xmark & \cmark & \cmark & 6.90 & 89.40 \\
Epona~\cite{zhang2025epona}
& \xmark & \xmark & \cmark & \cmark & 7.50 & 82.80 \\
\midrule
BEVControl~\cite{yang2023bevcontrol}
& \xmark & \cmark & \xmark & \cmark & 24.85 & -- \\
BEVGen~\cite{swerdlow2024street}
& \xmark & \cmark & \xmark & \cmark & 25.54 & -- \\
MagicDrive~\cite{gao2024magicdrive}
& \xmark & \cmark & \xmark & \cmark & 16.20 & -- \\
\midrule
UniScene~\cite{li2025uniscene}
& \xmark & \cmark & \cmark & \xmark & 6.45 & 71.94 \\
DiST-4D~\cite{guo2025dist}
& \xmark & \cmark & \cmark & \xmark & 7.40 & 25.55 \\
OmniNWM~\cite{li2025omninwm}
& \xmark & \cmark & \cmark & \xmark & 5.45 & 23.63 \\
\midrule
MagicDrive~\cite{gao2024magicdrive}
& \xmark & \cmark & \cmark & \cmark & 18.75 & 218.12 \\
GenAD~\cite{yang2024genad}
& \xmark & \cmark & \cmark & \cmark & 15.40 & 184.00 \\
Panacea~\cite{wen2024panacea}
& \xmark & \cmark & \cmark & \cmark & 16.96 & 139.00 \\
Drive-WM~\cite{wang2024driving}
& \xmark & \cmark & \cmark & \cmark & 15.80 & 122.70 \\
DriveDreamer-2~\cite{zhao2025drivedreamer}
& \xmark & \cmark & \cmark & \cmark & 25.00 & 105.10 \\
\midrule
FAR-Drive~\cite{li2026far}
& \cmark & \cmark & \cmark & \cmark & 11.92 & 82.78 \\
\rowcolor{black!6}
\textbf{HelloWorld (Ours)}
& \cmark & \cmark & \cmark & \cmark & 7.75 & 41.08 \\
\midrule[0.8pt]
\multicolumn{7}{@{}l}{\textit{Long-horizon generation}} \\
\addlinespace[3pt]
\multicolumn{5}{@{}l}{MagicDrive-V2~\cite{gao2025magicdrive}}
& 20.91 & 94.84 \\
\multicolumn{5}{@{}l}{HorizonDrive~\cite{zhang2026horizondrive}}
& 13.82 & 92.99 \\
\rowcolor{black!6}
\multicolumn{5}{@{}l}{\textbf{HelloWorld (Ours)}}
& 17.81 & 88.10 \\
\bottomrule
\end{tabularx}
\end{table}

%% file: sections/lidar_experiments.tex
\input{sections/lidar_structural_metrics}
\subsubsection{Comparison with Cosmos1}
\label{sec:exp-lidar-roads}
We compare the fleet-adapted Cosmos1 three-camera, single-frame generator with our auxiliary-supervised generator, before and after temporal RRN, on the same 100 recorded clips (2,900 frames). All methods use one shared reference and are scored within the union of the front-wide, left-rear and right-rear camera frusta. Our model retains seven-camera video conditioning; Cosmos1 generates each frame from three cameras. Equalizing the output scoring region does not equalize input information, training budget or architecture.

\begin{table}[H]
\centering\small
\caption{Common three-camera-region evaluation on 100 clips, frames 0--28. All five metrics use the same GT-derived output region; MAE uses pixels valid in all three predictions (82.7671\% of regional valid GT). Static-Align uses shared reference-ICP transforms. Distances are in meters and alignment scores are fractions.}
\label{tab:lidar-road-structure}
\begin{tabularx}{\linewidth}{@{}Xrrr@{}}
\toprule
Metric & Cosmos1 & Ours & + Temporal RRN \\
\midrule
Range MAE (m) $\downarrow$ & 6.1724 & 2.2199 & \textbf{2.1629} \\
Chamfer Distance (m) $\downarrow$ & 2.5709 & \textbf{1.0134} & 1.0426 \\
Edge Pred$\rightarrow$GT (m) $\downarrow$ & 1.5802 & 0.7976 & \textbf{0.7115} \\
Ray-Align $\downarrow$ & 0.3776 & 0.4245 & \textbf{0.2757} \\
Static-Align $\uparrow$ & 0.4503 & 0.6928 & \textbf{0.8463} \\
\bottomrule
\end{tabularx}
\end{table}

Our temporally rectified generator improves all five reported measures relative to Cosmos1 under this output-region protocol. Cosmos1 nevertheless has less radial streaking than our unrectified generator according to Ray-Align, despite higher geometric errors and weaker temporal agreement. Within our pipeline, temporal RRN reduces MAE and predicted-boundary error and improves structural agreement, but increases Chamfer from 1.0134 to 1.0426\,m; rectification therefore does not improve every geometric measure. All 2,800 reference-ICP pairs pass the fitness threshold of 0.6, with none discarded. 

\subsubsection{LiDAR VAE reconstruction and rectification}
\label{sec:exp-lidar-tokenizer}
We assess the first training stage independently of conditional generation. Table~\ref{tab:lidar-final-vae} evaluates one frozen video VAE on 100 clips of 29 frames, without rectification, with single-frame RRN, and with motion-compensated temporal RRN. All three arms use the same windows, reference data and evaluation protocol. Range MAE uses the intersection of predicted and reference return validity with native range gating (\texttt{common-valid(pred\_validity)}), pooled over valid pixels; it is not an error over the interpolated target canvas. Point-cloud and structural metrics are computed from each branch's gated range maps, independently of the choice of depth-error mask.

\begin{table}[H]
\centering\small
\caption{LiDAR VAE reconstruction on 100 clips (2,900 frames). Range MAE uses common-valid returns; temporal scores use recorded poses. Distances are in meters and alignment scores are fractions.}
\label{tab:lidar-final-vae}
\begin{tabularx}{\linewidth}{@{}Xccc@{}}
\toprule
Metric & VAE & + Single-frame RRN & + Temporal RRN\\
\midrule
Range MAE (m) $\downarrow$ & 0.3981 & 0.3839 & \textbf{0.3775}\\
Chamfer Distance (m) $\downarrow$~\citep{fan2017pointset} & 0.3978 & 0.3815 & \textbf{0.3695}\\
Edge Pred$\rightarrow$GT (m) $\downarrow$~\citep{xu2025pixelperfect} & 0.2997 & 0.2331 & \textbf{0.2106}\\
Ray-Align $\downarrow$ & 0.2967 & 0.2610 & \textbf{0.2589}\\
Static-Align $\uparrow$ & 0.8215 & 0.8510 & \textbf{0.8974}\\
Warp Chamfer (m) $\downarrow$ & 0.4668 & 0.4422 & \textbf{0.3668}\\
\bottomrule
\end{tabularx}
\end{table}

Temporal RRN reduces range MAE by 5.2\%, Chamfer distance by 7.1\%, and edge prediction-to-reference distance by 29.7\% relative to the unrectified VAE. Static-Align rises from 82.15\% to 89.74\% (+7.59 percentage points), while pose-compensated Warp Chamfer falls from 0.4668 to 0.3668\,m. These gains do not imply improved return coverage.
RRN leaves the validity head unchanged, although corrected depths can cross the range gates. The single-frame and temporal models are separately trained recipes, so their difference is not a controlled ablation of temporal input alone.

\subsubsection{RGB-conditioned LiDAR generation}
\paragraph{Conditional generation with frozen-decoder supervision.}
\label{sec:exp-lidar-conditional}
Our final DiT results use the selected auxiliary-supervised seven-camera generator, evaluated on 100 clips of 29 frames with 35 sampling steps. Table~\ref{tab:lidar-conditional-rrn} compares the same saved generator outputs without rectification, with single-frame RRN, and with temporal RRN. This generator uses three-channel azimuth/coverage ray features and retains its original frozen tokenizer, distinct from the final reconstruction tokenizer, preserving the latent/decoder pairing used in training. Both rectifiers are matched to this generation pipeline rather than substituted from the final VAE reconstruction experiment. The four decoded auxiliary terms are described in Section~\ref{sec:app-lidar-implementation}. This comparison evaluates postprocessing of the selected auxiliary-supervised model; it does not isolate auxiliary supervision from RF-only training.

\begin{table}[H]
\centering\small
\caption{Full-panorama RGB-to-LiDAR generation on 100 clips (2,900 frames), without the three-camera-region restriction of Table~\ref{tab:lidar-road-structure}. All arms use the same saved generator outputs and reference data; rectification does not rerun the DiT. Distances are in meters and alignment scores are fractions.}
\label{tab:lidar-conditional-rrn}
\begin{tabularx}{\linewidth}{@{}Xrrr@{}}
\toprule
Metric & Generator & + Single-frame RRN & + Temporal RRN \\
\midrule
Range MAE (m) $\downarrow$ & 2.2797 & 2.2532 & \textbf{2.2194} \\
Chamfer Distance (m) $\downarrow$ & \textbf{1.0738} & 1.1038 & 1.0920 \\
Edge Pred$\rightarrow$GT (m) $\downarrow$ & 0.8814 & 0.8339 & \textbf{0.7990} \\
Ray-Align $\downarrow$ & 0.4366 & 0.3238 & \textbf{0.2822} \\
Static-Align $\uparrow$ & 0.6741 & 0.7081 & \textbf{0.8310} \\
\bottomrule
\end{tabularx}
\end{table}

Ray-Align and Static-Align in this table cover all 29 frames (indices 0--28), comprising 2,900 ray-alignment frames and 2,800 adjacent frame pairs across 100 clips. All ICP pairs pass the minimum fitness threshold of 0.6, with none discarded. Static-Align uses shared transforms estimated from GT LiDAR for all three arms; scores are averaged within each clip and then across clips. These reference-motion-assisted scores use panoramic outputs rather than the three-camera region above. The MAE validity support and point filtering also differ, so values should not be compared directly across the two tables.

Temporal RRN reduces range MAE and edge prediction-to-reference distance. Ray-Align falls from 0.4366 to 0.2822 relative to the generator, while Static-Align rises from 67.41\% to 83.10\% (+15.69 percentage points). Chamfer improves over single-frame RRN but remains worse than the generator, so these gains do not establish uniformly improved geometry. 
These are measured generation results, not transferred VAE reconstruction scores.

\FloatBarrier
\subsubsection{Qualitative results}
\begin{figure}[p]
\centering
\includegraphics[width=\linewidth,height=.85\textheight,keepaspectratio]{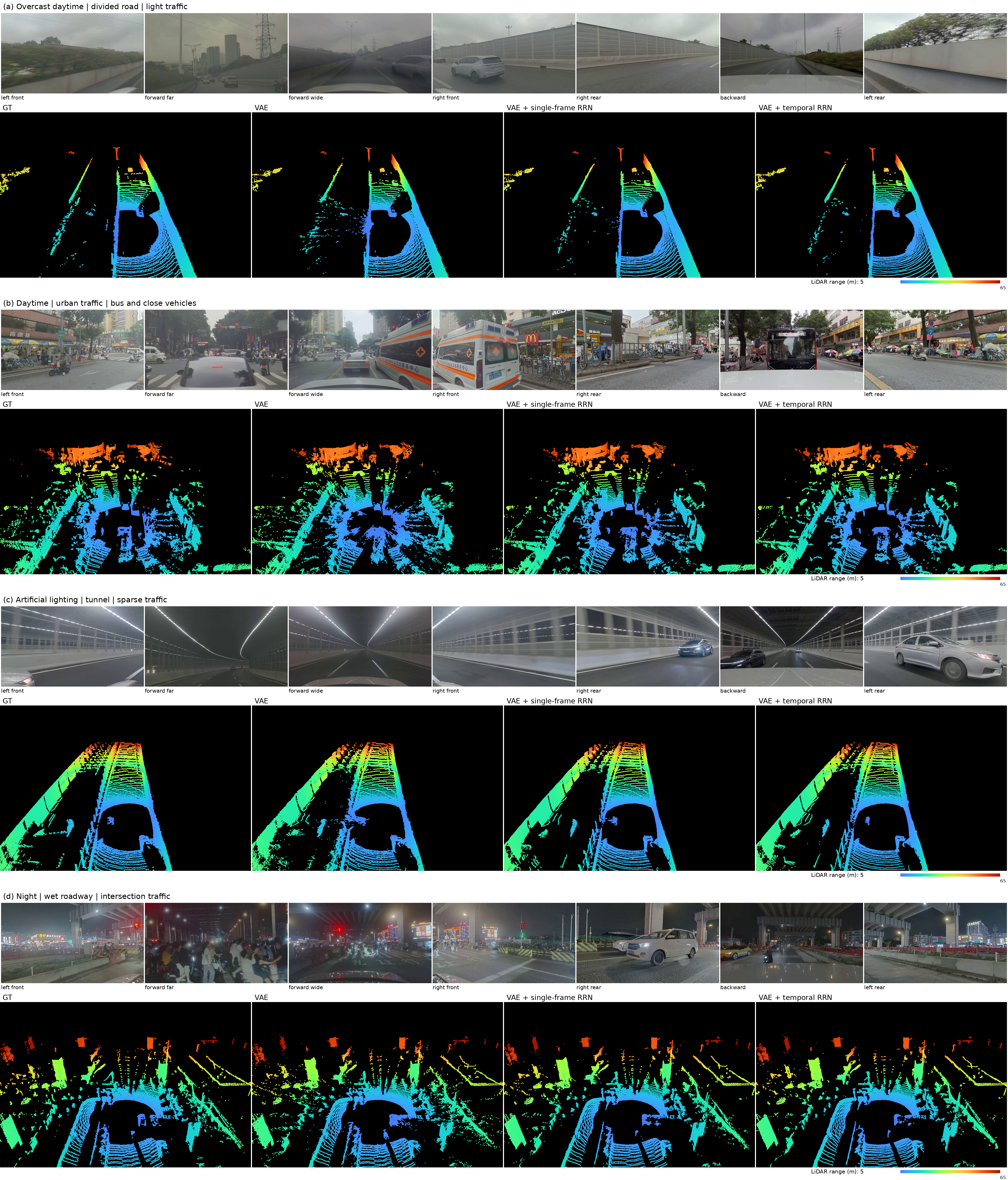}
\caption{LiDAR VAE reconstruction across four recorded scenes: (a) overcast divided road, (b) daytime urban traffic, (c) illuminated tunnel, and (d) a wet roadway at night. For each scene, seven synchronized RGB views are followed by point clouds of GT, VAE reconstruction, VAE + single-frame RRN, and VAE + temporal RRN, from left to right. All examples use frame 14 and the same VAE. Point colors encode range with a shared 5--65\,m color scale; this scale is not the evaluation range gate. RGB is shown only for scene context.}
\label{fig:lidar-vae-four-scenes}
\end{figure}

Figure~\ref{fig:lidar-vae-four-scenes} presents paired reconstructions in four scenes selected for scene diversity rather than reconstruction scores. The camera rows provide visual context only; the VAE reconstructs LiDAR input and is not conditioned on RGB. Each point-cloud row compares the same frame and viewing direction across GT, VAE, single-frame RRN and temporal RRN. Still images illustrate geometry but do not establish temporal stability by themselves.

\begin{figure}[p]
\centering
\begin{minipage}[t]{.485\linewidth}
\centering
{\small (a) Temporal RRN, $t=2.8$\,s\par}\vspace{3pt}
\includegraphics[width=\linewidth,height=.405\textheight,keepaspectratio]{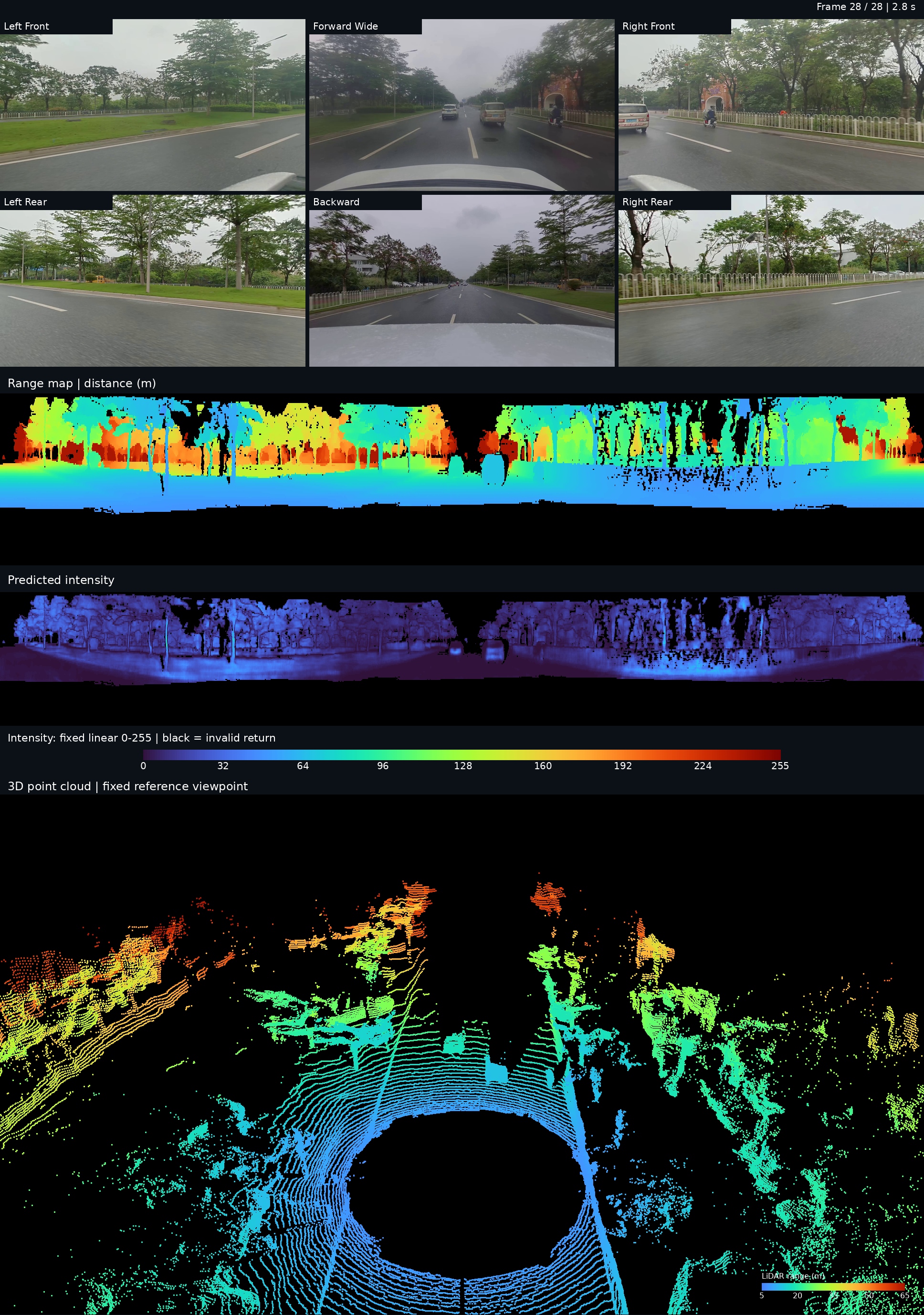}
\end{minipage}\hfill
\begin{minipage}[t]{.485\linewidth}
\centering
{\small (b) Temporal RRN, $t=1.4$\,s\par}\vspace{3pt}
\includegraphics[width=\linewidth,height=.405\textheight,keepaspectratio]{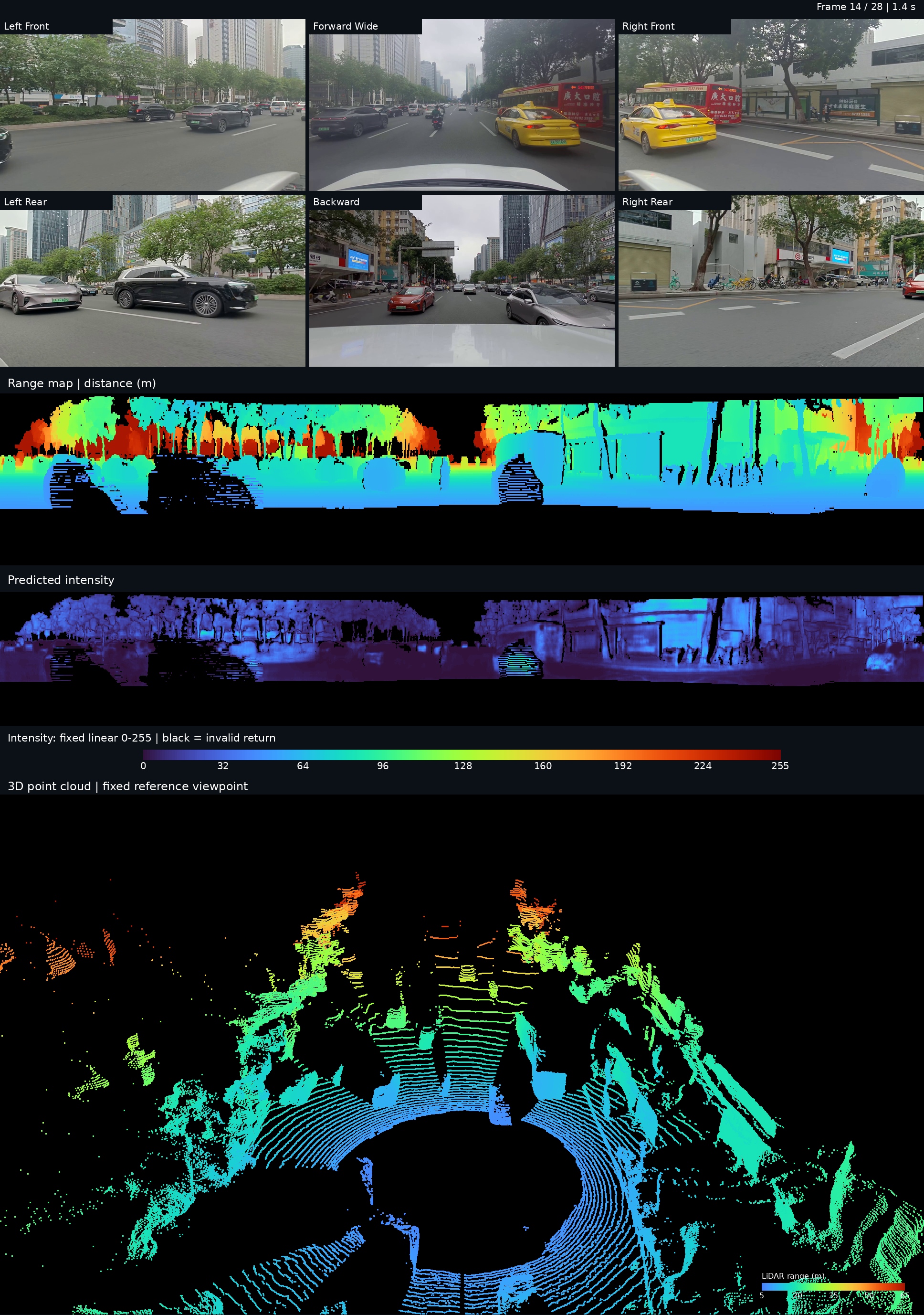}
\end{minipage}
\par\vspace{7pt}
\begin{minipage}[t]{.485\linewidth}
\centering
{\small (c) Temporal RRN, $t=1.4$\,s\par}\vspace{3pt}
\includegraphics[width=\linewidth,height=.405\textheight,keepaspectratio]{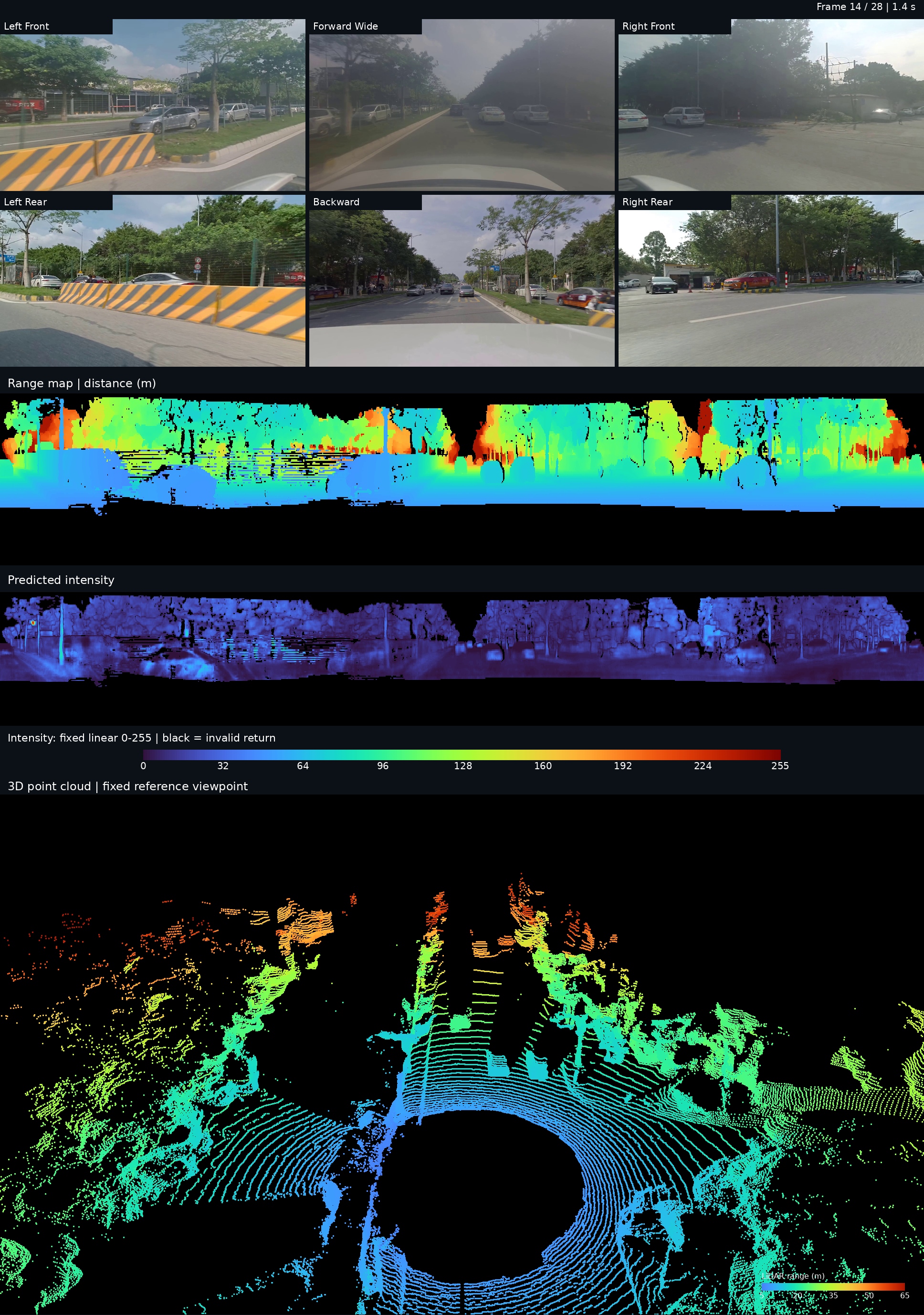}
\end{minipage}\hfill
\begin{minipage}[t]{.485\linewidth}
\centering
{\small (d) Temporal RRN, $t=2.0$\,s\par}\vspace{3pt}
\includegraphics[width=\linewidth,height=.405\textheight,keepaspectratio]{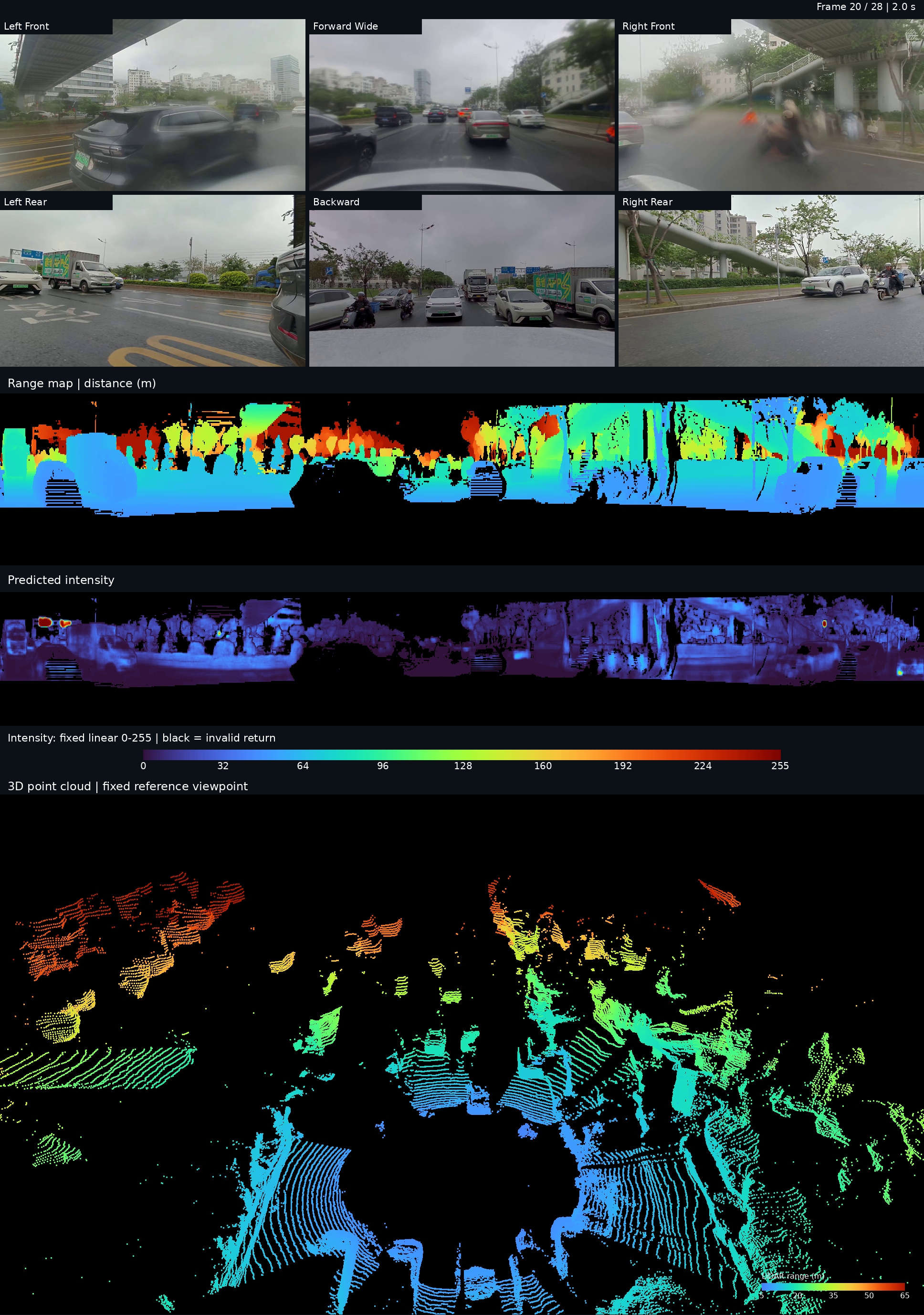}
\end{minipage}
\caption{RGB-to-LiDAR generation with temporal RRN on four recorded-road scenes. Each panel shows six displayed camera views, predicted range, auxiliary intensity, and a range-colored point cloud. Seven views condition generation; forward-far is omitted only from display. All four panels use the same temporal rectifier. Intensity is predicted by a separate head, not GT. These are generated and rectified point clouds, not VAE reconstructions.}
\label{fig:lidar-rgb2lidar-example}
\end{figure}

Figure~\ref{fig:lidar-rgb2lidar-example} shows four recorded-road examples from the selected generator with synchronized camera observations, generated range images, and point clouds rectified by temporal RRN. All four examples use the same temporal RRN checkpoint as the temporal arm of Table~\ref{tab:lidar-conditional-rrn}, but are qualitative road examples rather than the 100-clip aggregate evaluation itself. The auxiliary intensity panels come from a separate prediction head, not the range/validity tokenizer.

\begin{figure}[htbp]
\centering
\begin{tabular}{@{}cc@{}}
\small (a) Clear morning, $t=2.0$\,s & \small (b) Light fog, $t=2.0$\,s\\
\includegraphics[width=.48\linewidth]{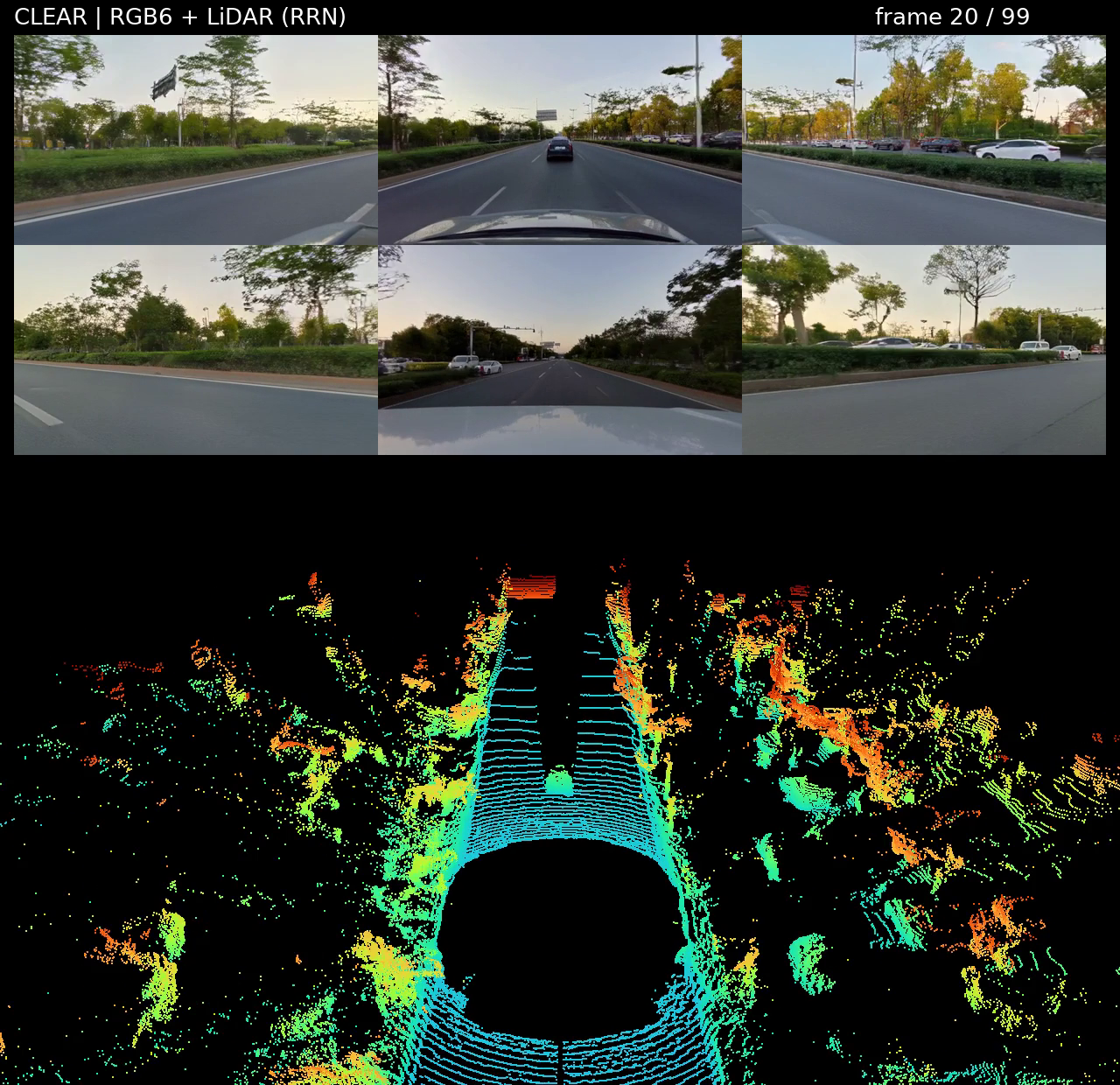}&
\includegraphics[width=.48\linewidth]{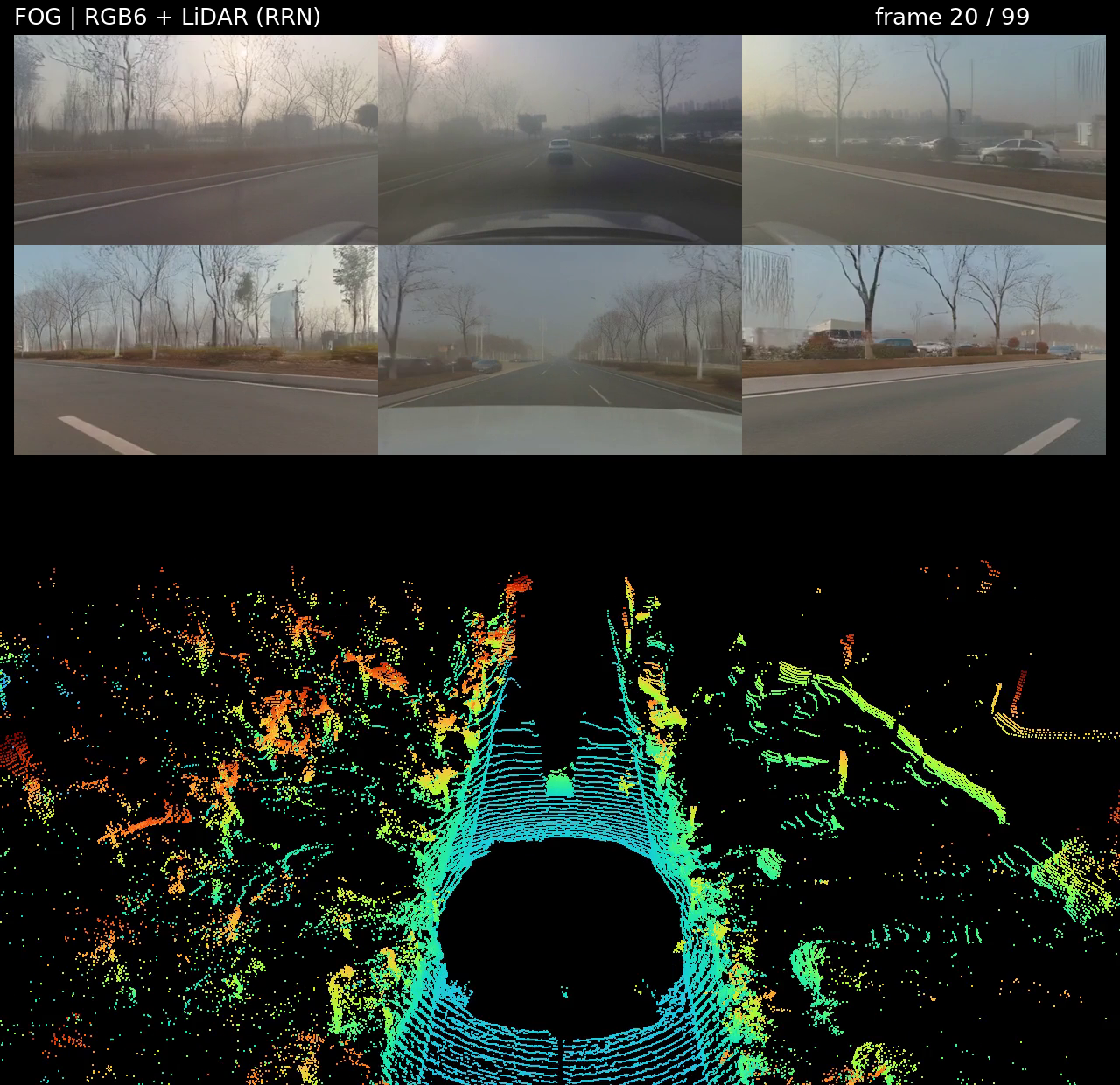}
\end{tabular}
\caption{LiDAR synthesis conditioned on generated RGB: selected clear-morning and light-fog frames at $t=2.0$\,s. Each panel pairs the displayed camera video mosaic with its synchronized rectified point cloud. These examples illustrate use beyond recorded RGB, not measured physical simulation of fog or a controlled weather ablation. Selected stills do not establish long-horizon consistency or temporal-RRN performance.}
\label{fig:lidar-generated-weather}
\end{figure}

\FloatBarrier
\paragraph{LiDAR synthesis from generated videos.}
\label{sec:app-lidar}
Figure~\ref{fig:lidar-generated-weather} shows clear-morning and light-fog examples conditioned on generated RGB, both at $t=2.0$\,s. Each panel pairs a camera mosaic with the synchronized RRN-refined point cloud. The scenes use different generation seeds; these stills demonstrate application beyond recorded RGB, not a controlled weather ablation or proof of long-horizon consistency. No paired target scan is available for the altered scenes, so the recorded-input scores above are not assigned to these examples.

\FloatBarrier

%% file: sections/lidar_structural_metrics.tex
\paragraph{Reconstruction metrics.}
Range MAE measures absolute depth error on rays valid in both prediction and reference. Chamfer distance (CD) sums the two directional mean nearest-neighbor distances, using unsquared Euclidean distances in meters~\citep{fan2017pointset}. Edge Pred$\rightarrow$GT applies the prediction-to-reference distance to a GT-derived boundary band, adapting edge-aware evaluation~\citep{xu2025pixelperfect}: range jumps above $\max(2\,\mathrm m,0.1r_{\mathrm{near}})$ are dilated by two pixels before intersecting the scoring region. Warp Chamfer applies CD to consecutive predicted clouds after ego-motion compensation. Lower is better for these distances; common-valid MAE and one-way edge distance do not measure missing-return coverage.

\paragraph{Ray-Align and Static-Align.}
We introduce two complementary structural diagnostics:
\begin{equation}
\mathrm{RayAlign}=\frac{1}{|\mathcal N|}\sum_{i\in\mathcal N}
\mathbf 1\{|n_i^\top u_i|<0.2\},\qquad
\mathrm{StaticAlign}_t=\frac{1}{|\mathcal Q_t|}\sum_{i\in\mathcal Q_t}
\mathbf 1\{|r(Tp_{t,i})-\hat r_{t+1}(\pi(Tp_{t,i}))|<0.5\,\mathrm m\}.
\end{equation}
For Ray-Align, $u_i$ is the unit sensor ray and $n_i$ is the unit normal from right/down point neighbors. The set $\mathcal N$ retains valid triangles with both neighbor distances below 3\,m and nondegenerate normals. Lower scores indicate fewer surfaces stretched along sensor rays, although real grazing surfaces also contribute. For Static-Align, $T$ compensates ego motion, $r$ measures range, and $\pi$ projects into the next range image. The set $\mathcal Q_t$ retains in-bounds projections with valid target returns; no occlusion filtering is applied. Higher scores indicate stronger temporal agreement, not necessarily correct geometry or semantic alignment with RGB.

Structural scores are averaged within each clip and then equally across clips. VAE evaluation uses recorded poses; generation evaluation uses the same GT-derived ICP transforms for every method, not independent alignment of predictions. The three-camera comparison restricts both diagnostics to the common GT-derived camera region, whereas the full-panorama results do not. Thus Static-Align is motion-assisted, and neither diagnostic alone establishes camera--LiDAR correspondence.

%% file: sections/applications.tex
\section{Applications}
\label{sec:applications}

The preceding experiments evaluate the individual capabilities of HelloWorld in terms of generation quality,
control fidelity, multi-view consistency, sequential generation, and inference efficiency.
We further examine how these capabilities can be combined in downstream driving applications.
We focus on two representative settings:
\emph{special-scene editing}, where recorded driving contexts are transformed into counterfactual scenarios through controllable generation, \emph{closed-loop simulation}, where HelloWorld serves as an observation generator under iteratively updated ego motion.
These applications illustrate two complementary uses of the same world model:
controllable data generation and interactive observation synthesis beyond recorded driving logs.

\subsection{Special-Scene Editing}
\label{sec:app-special-editing}

We further evaluate HelloWorld on special-scene editing, where rare or safety-critical events are inserted into recorded driving scenes in a controllable manner.
Rather than generating such cases from scratch, we preserve the original road geometry, ego trajectory, and structured scene context, and modify only the semantic or object-level conditions to introduce a target event.
This setting tests whether the model can perform scene-consistent counterfactual editing while maintaining geometric and contextual coherence.

\begin{figure*}[t]
    \centering

    \includegraphics[width=0.49\textwidth]{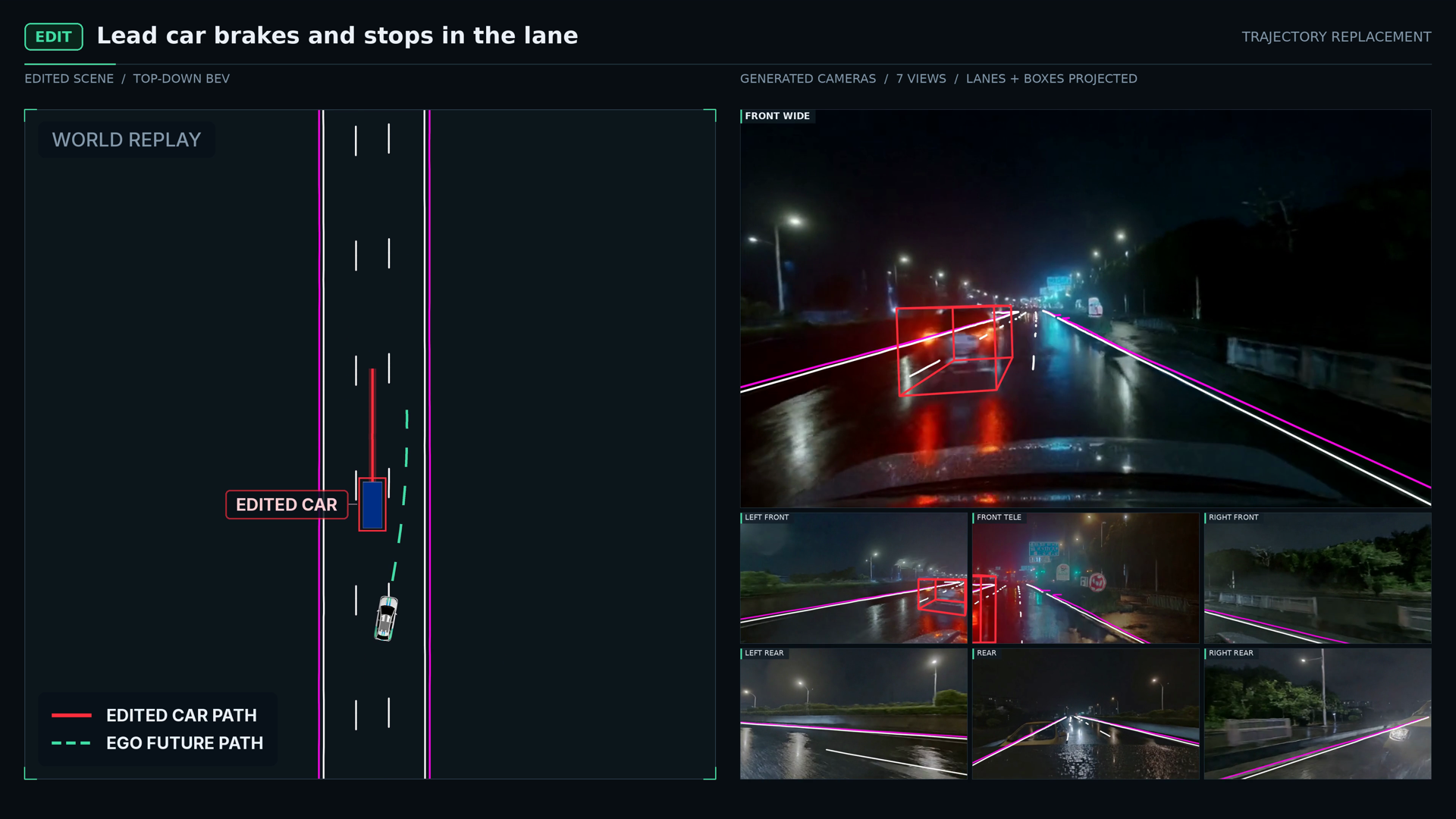}
    \hfill
    \includegraphics[width=0.49\textwidth]{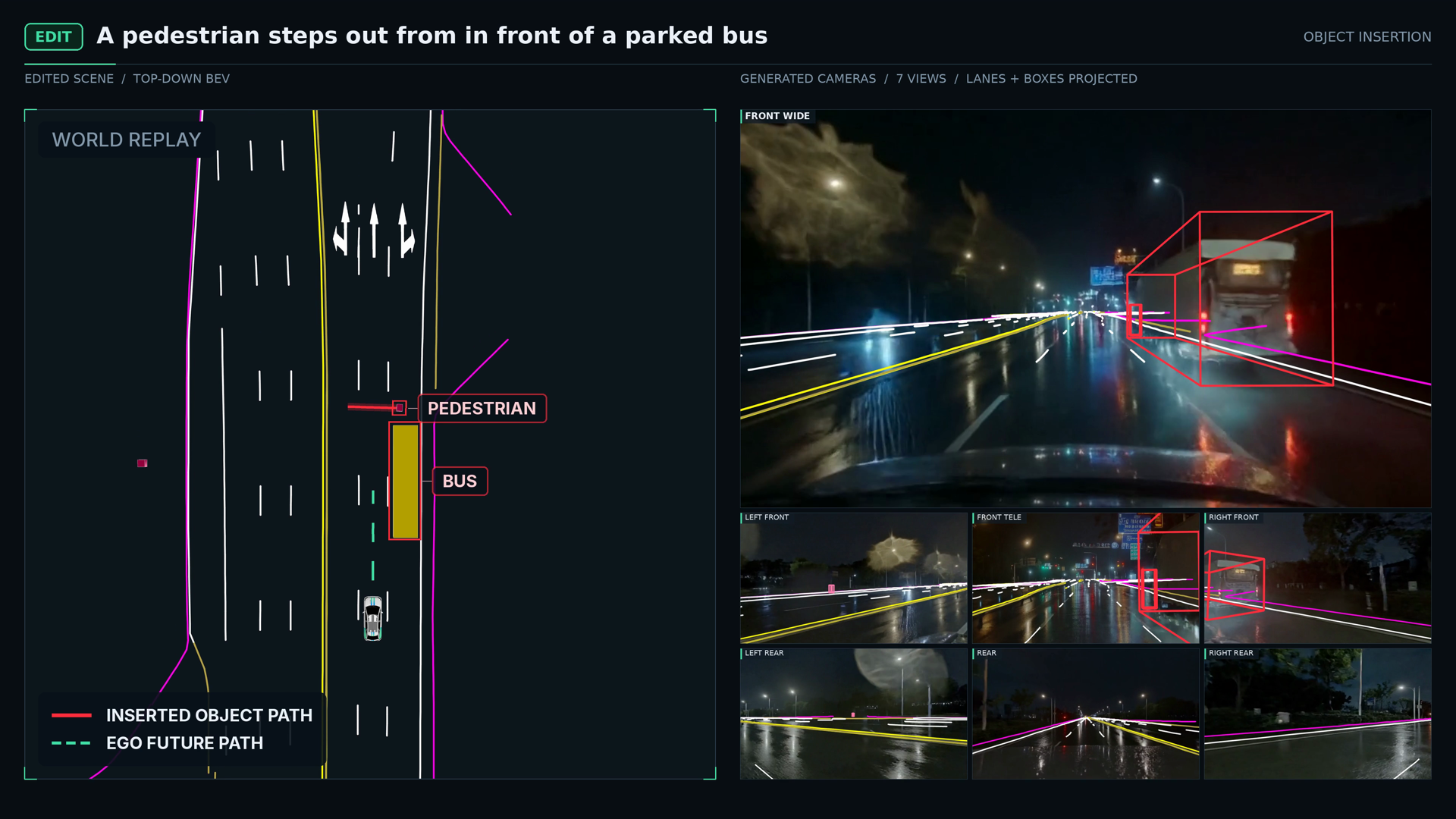}

    \vspace{1.5mm}

    \includegraphics[width=0.49\textwidth]{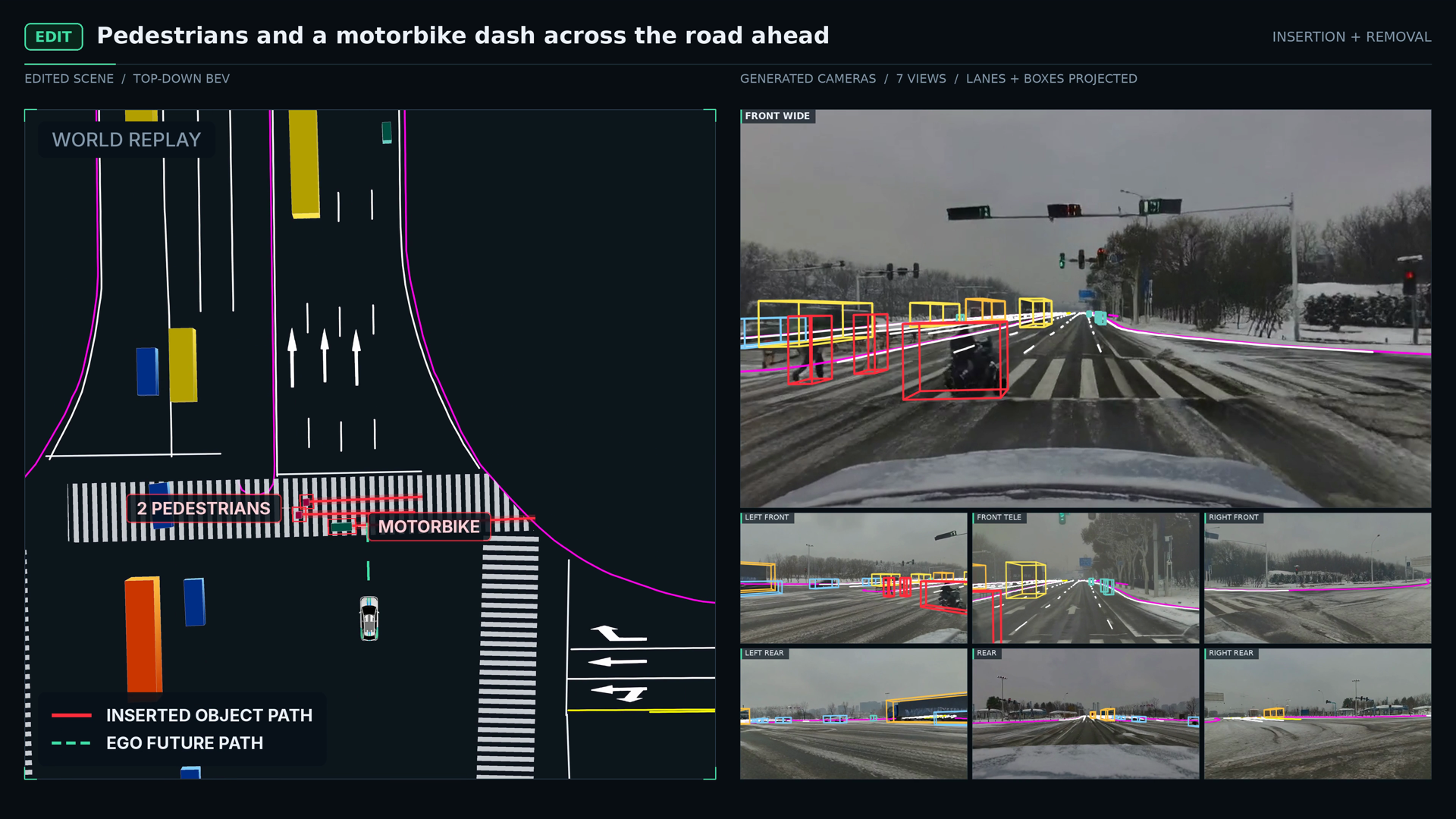}
    \hfill
    \includegraphics[width=0.49\textwidth]{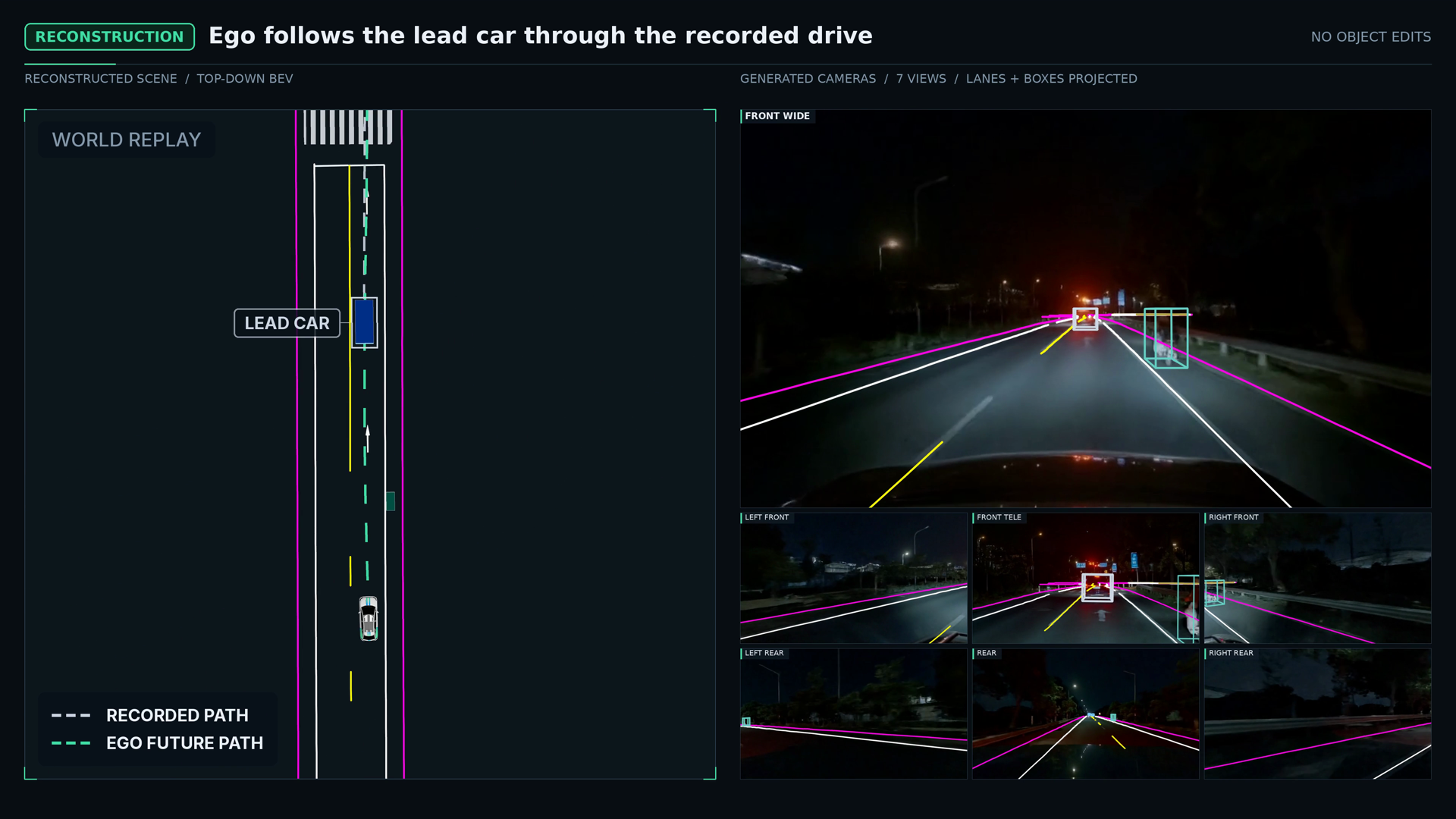}

    \vspace{1.5mm}

    \includegraphics[width=0.49\textwidth]{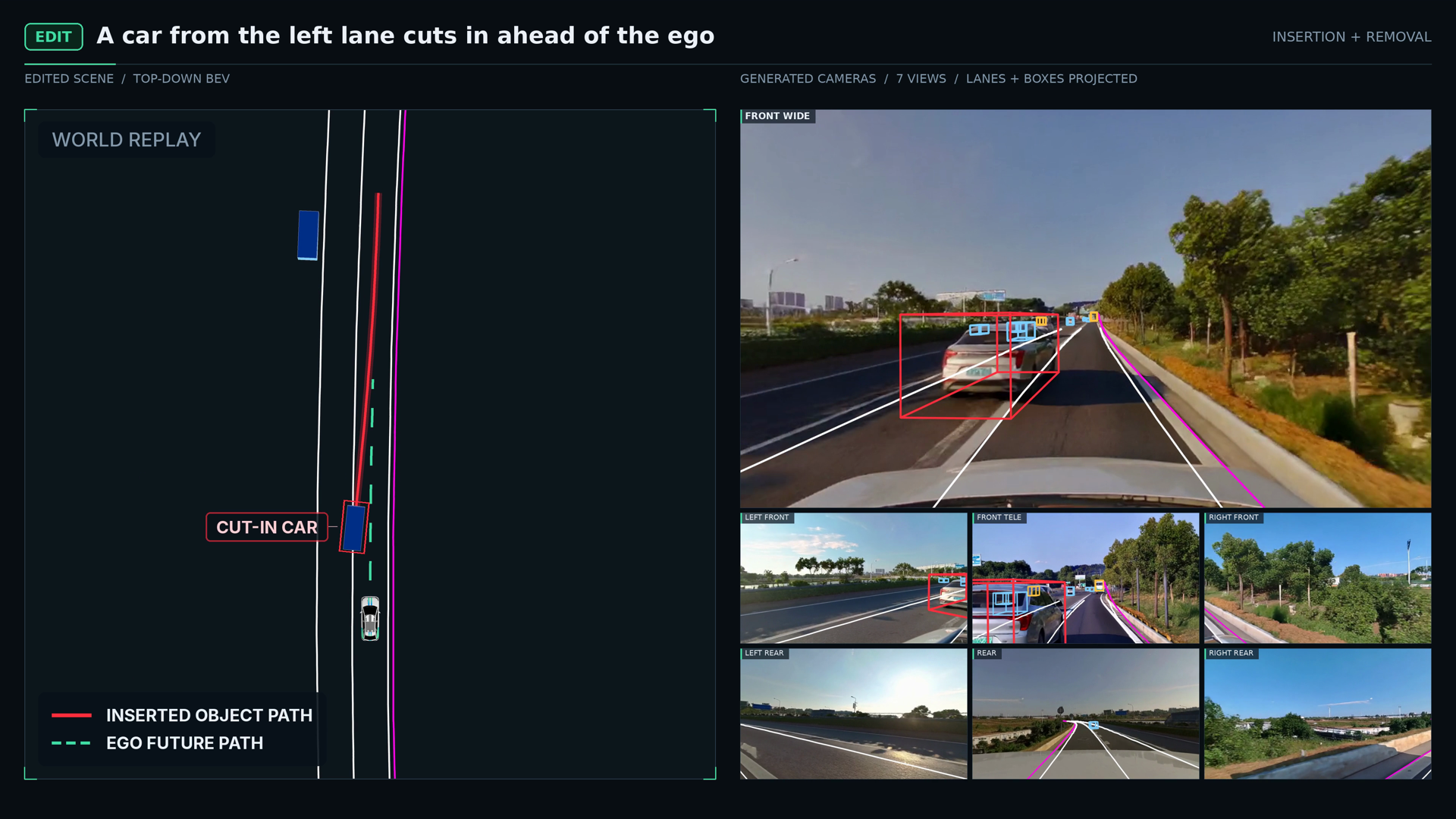}
    \hfill
    \includegraphics[width=0.49\textwidth]{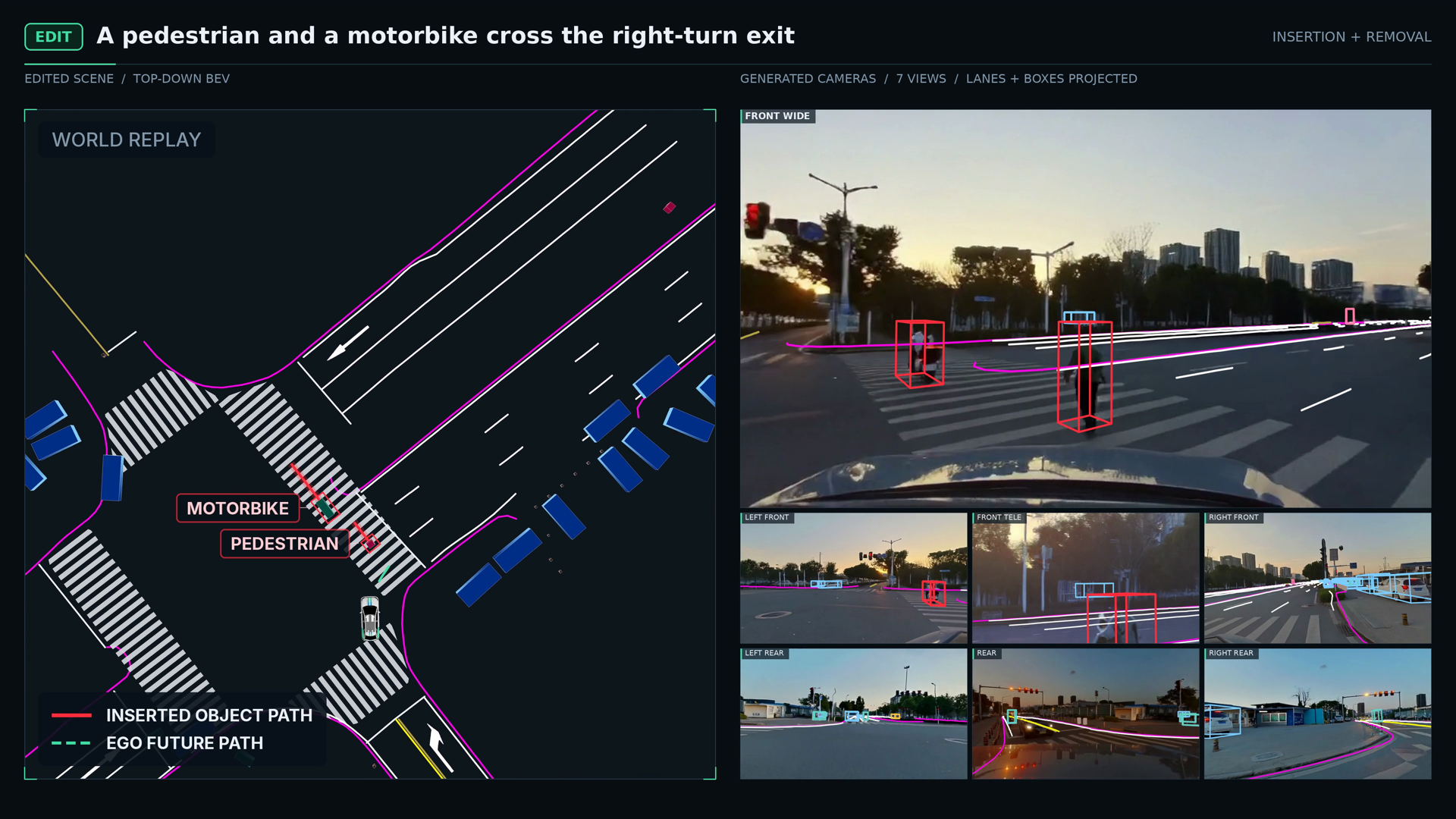}

    \vspace{1.5mm}

    \includegraphics[width=0.49\textwidth]{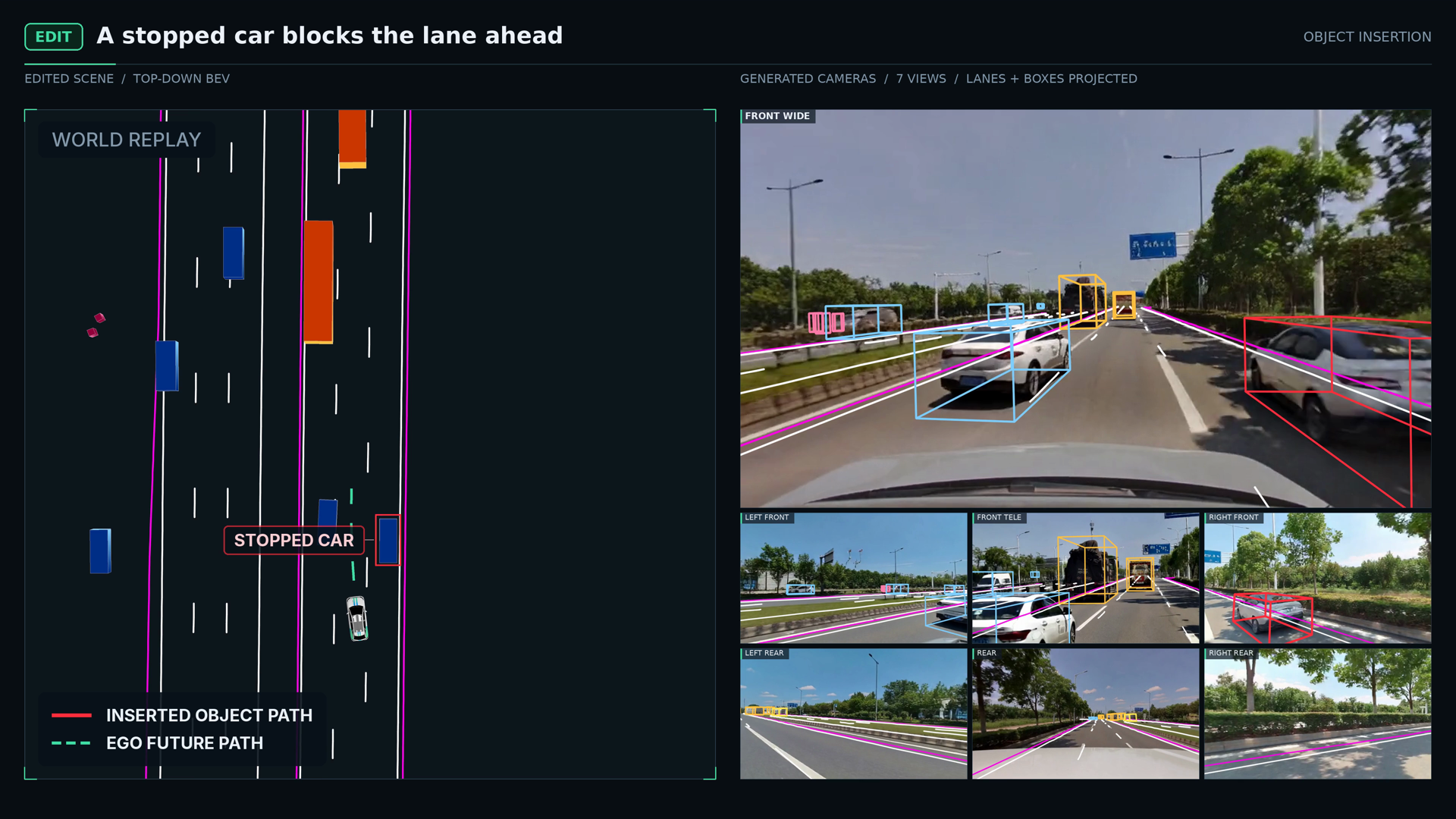}
    \hfill
    \includegraphics[width=0.49\textwidth]{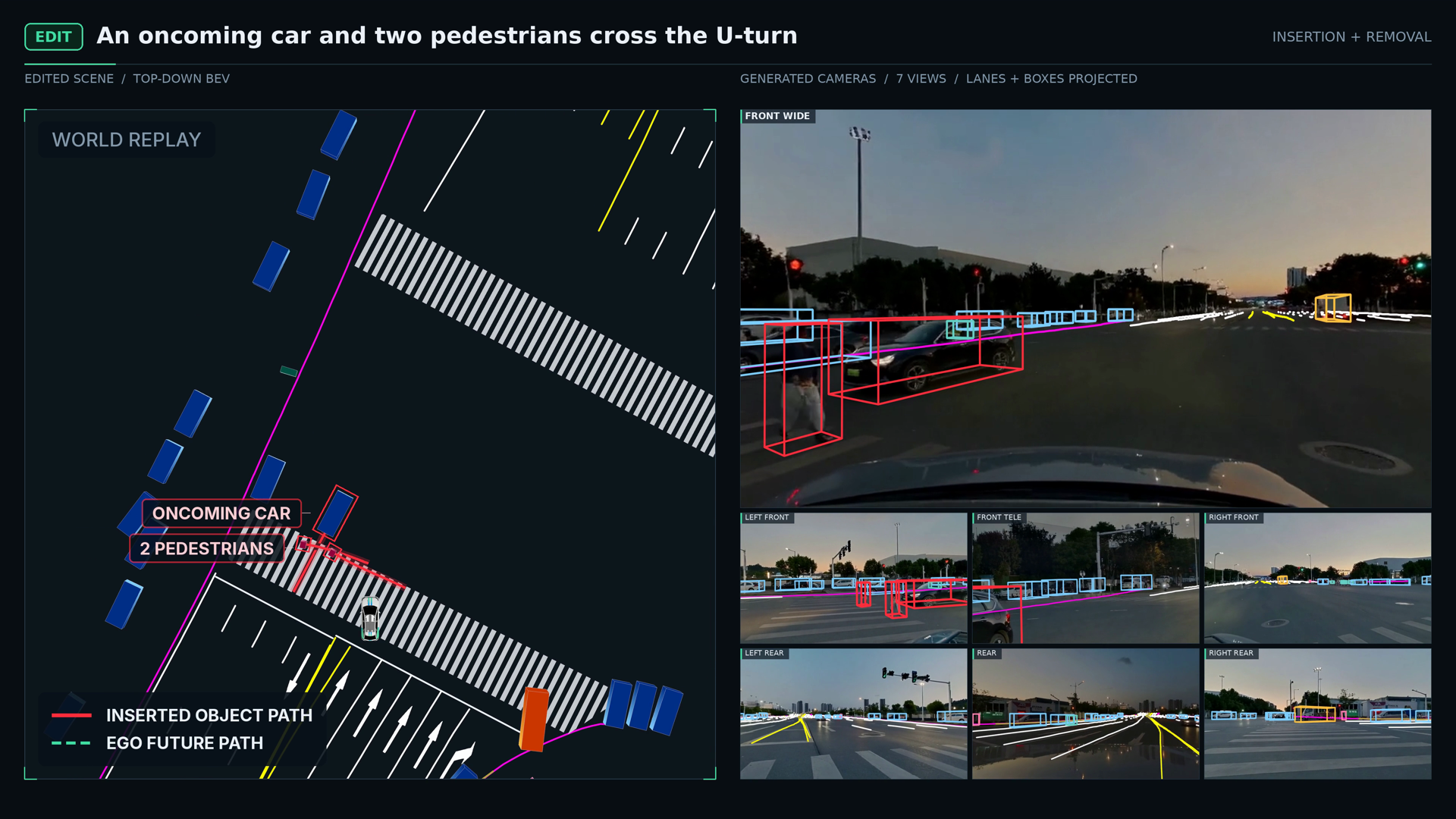}

    \caption{
    Special-scene editing results.
    Starting from recorded driving scenes, HelloWorld preserves the original road geometry, ego trajectory, and surrounding scene context while introducing rare traffic agents or events through edited semantic and object-level conditions.
    Each example shows the edited top-down BEV together with the corresponding synchronized multi-view observations.
    The edited scenarios include vehicle trajectory changes, object insertion, and safety-critical interactions involving pedestrians, motorbikes, and surrounding vehicles, demonstrating scene-consistent counterfactual editing under realistic driving contexts.
    }
    \label{fig:special_scene_editing}
\end{figure*}

As illustrated in Figure~\ref{fig:special_scene_editing}, HelloWorld can insert challenging traffic events such as a pedestrian stepping out from in front of a parked bus, pedestrians and a motorbike crossing the road ahead, and pedestrian--motorbike conflicts near a right-turn exit.
The edited results remain aligned with the original scene layout and multi-view observations, while clearly reflecting the inserted agents and their intended motions.
These examples show that HelloWorld is not limited to replaying common driving patterns, but can also support controllable generation of long-tail scenarios for safety-oriented data augmentation and stress testing.

\subsection{Closed-Loop Simulation}
\label{sec:app-closedloop}
\begin{figure}[t]
    \centering
    \includegraphics[width=1.0\linewidth]{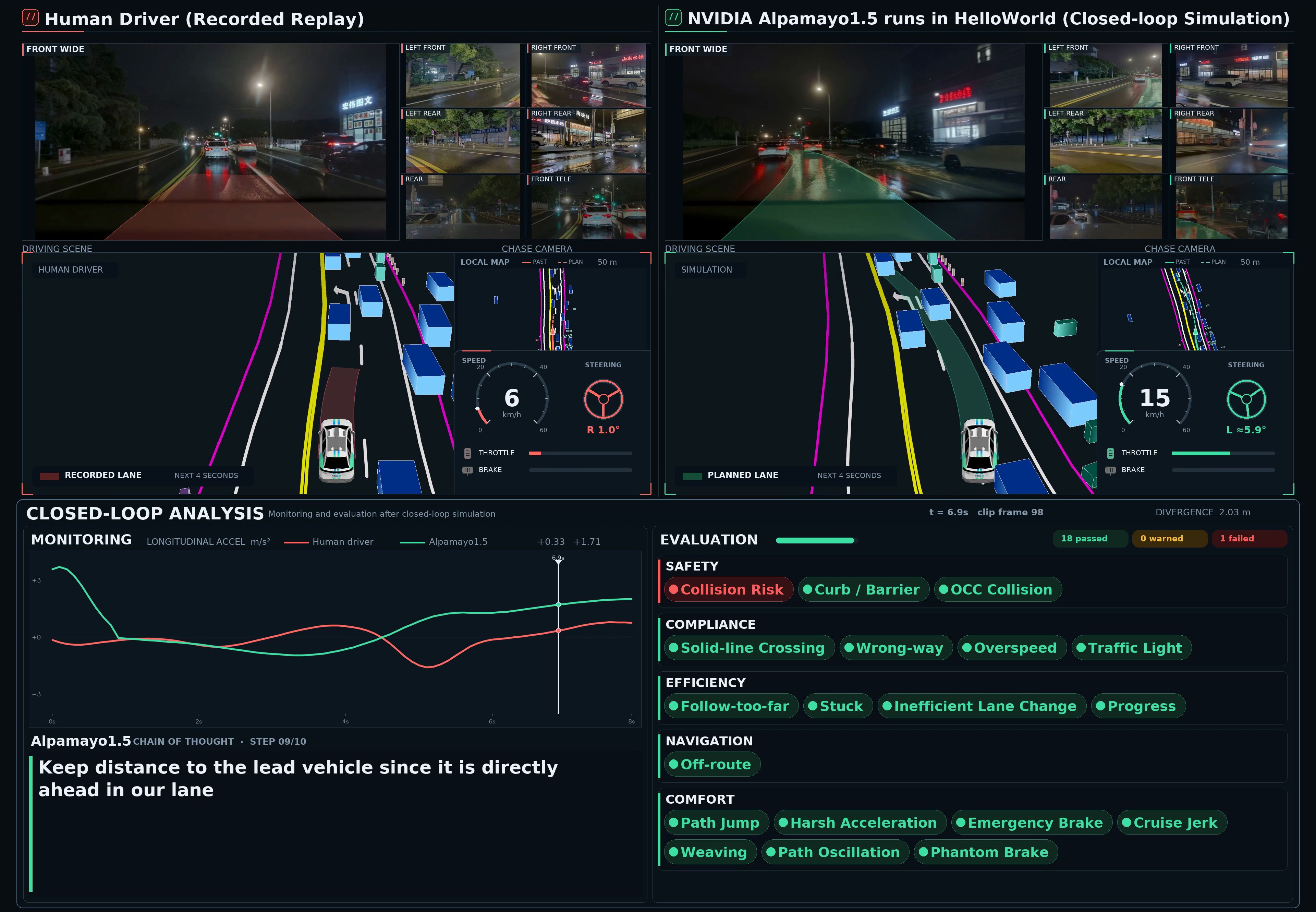}
    \caption{Closed-loop simulation with Alpamayo~1.5 under trajectory adjustment.
    The left side shows the recorded road-test replay, while the right side shows the HelloWorld simulation driven by Alpamayo~1.5.
    At each step, Alpamayo~1.5 predicts an updated ego trajectory from HelloWorld-generated multi-view observations, and HelloWorld synthesizes the corresponding next observations.
    The rollout is executed with the few-step HelloWorld student model, enabling efficient repeated simulation.
    In this example, the simulated trajectory gradually deviates from the recorded path while remaining consistent with the surrounding road structure and scene context.}
    \label{fig:close-sim-1}
\end{figure}

\begin{figure}[t]
    \centering
    \includegraphics[width=1.0\linewidth]{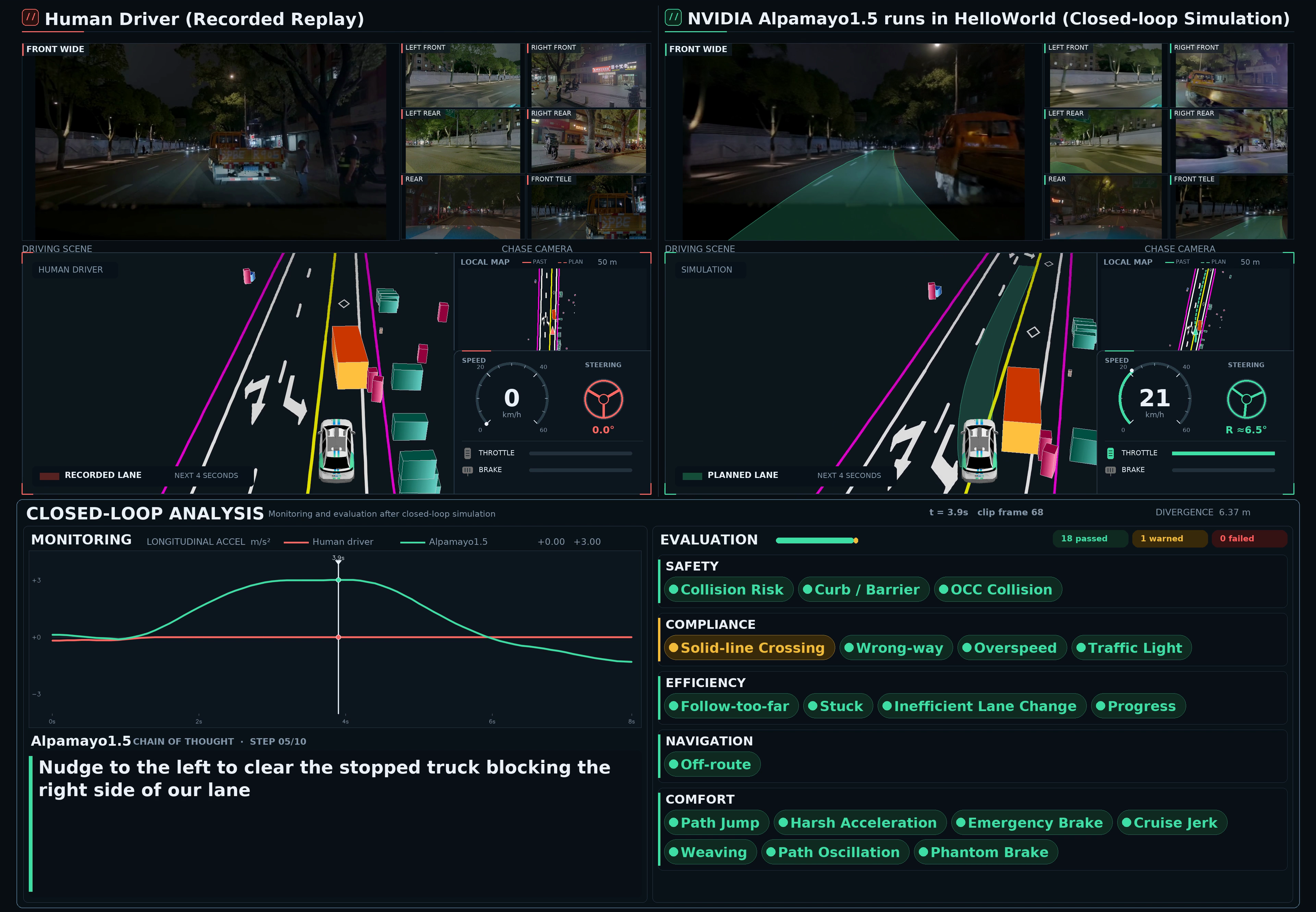}
    \caption{Closed-loop simulation with Alpamayo~1.5 for obstacle avoidance.
    The left side shows the recorded road-test replay, while the right side shows the corresponding closed-loop simulation.
    Alpamayo~1.5 observes the multi-view frames generated by HelloWorld and predicts a lateral avoidance trajectory around a stopped vehicle.
    HelloWorld then generates the subsequent observations conditioned on the updated ego motion.
    The rollout uses the few-step HelloWorld student model and demonstrates interactive world-model simulation under action-model feedback.}
    \label{fig:close-sim-2}
\end{figure}

HelloWorld can serve as the observation-generation component of a closed-loop driving simulation.
In our implementation, we couple HelloWorld with the open-source \textbf{Alpamayo~1.5} action model.
At each interaction step, Alpamayo~1.5 takes the current multi-view observations generated by HelloWorld and predicts the ego trajectory for the next simulation interval.
Conditioned on the committed observation history, the updated ego trajectory, and the available scene conditions,
HelloWorld then generates the corresponding multi-view observations for the next temporal chunk.
These observations are fed back to Alpamayo~1.5 for subsequent trajectory prediction, forming an iterative action--observation loop.
As the predicted trajectory deviates from the recorded trajectory, the simulator correspondingly generates observations along the newly selected ego motion rather than replaying the original driving sequence.
Representative closed-loop rollouts are shown in \Cref{fig:close-sim-1,fig:close-sim-2}.

A central challenge in this setting is that both components progressively operate outside the recorded trajectory:
the action model makes decisions from generated observations, while the world model conditions on its own previously generated history and newly predicted ego motion.
The block-causal formulation and self-generated-context adaptation described in Section~\ref{sec:method-causal}
are designed to support this repeated execution regime.
Moreover, the closed-loop system is run with the \emph{few-step} student model,
which substantially reduces the sampling cost of each interaction step and makes iterative action--observation rollout practical in simulation.

%% file: sections/limitations.tex
\section{Limitations}
\label{sec:limitations}
The model's scope is bounded by its observation interface, available supervision and finite history.

\paragraph{Observation causality and external dynamics.}
The trunk generates observations block-causally, but some control-tower paths use bidirectional attention over the supplied condition sequence. Strict online operation requires causal access in every conditioning pathway. Ego trajectories are prescribed. Low-level vehicle dynamics, policy feedback and reactive traffic behavior are outside the modeled transition. Generated observations do not constitute a validated closed-loop simulator.

\paragraph{Finite memory and conflicting conditions.}
A bounded KV cache limits accessible history and does not provide persistent scene memory for loop closure or revisited streets. Text, pose and layout can also conflict, without a guaranteed priority rule. Long-horizon evaluation must therefore include condition adherence, visibility changes and memory-boundary behavior in addition to visual smoothness.

\paragraph{Compression and observation coverage.}
Few-step inference can alter both per-chunk quality and the distribution of subsequent history. A favorable short-window trade-off does not guarantee equivalent long-horizon behavior. Native RGB resolution is $480\times832$. Image refinement adds a separate quality and latency trade-off. Layout response is limited by annotation coverage, particularly in unlabeled or occluded regions.

\paragraph{Geometric extension.}
The conditional LiDAR generator depends on the quality and calibration of its RGB inputs. A fleet-mean conditioning rig is not equivalent to per-clip calibration. Validity and range compression can lose thin structures, boundaries or rare returns. Recorded-camera and generated-camera conditioning expose different input distributions. Success with recorded RGB does not establish equivalent quality with generated RGB.

%% file: sections/conclusion.tex
\section{Conclusion}
\label{sec:conclusion}
HelloWorld organizes driving observation generation around a common block-causal transition. A supervision-driven curriculum progressively adds calibrated multi-view communication and structured scene control to a pose-conditioned video model. Generated-history alignment addresses the states encountered during repeated execution, while consistency and distribution-matching distillation target few-step sampling of the same transition.

The framework makes capability retention across these transformations a central evaluation question: appearance coverage, cross-view geometry and condition response must be assessed together with horizon and execution cost. Conditional LiDAR synthesis extends the observation interface through explicit return validity, geometric compression and native-grid multiview conditioning. The resulting formulation connects progressive capability learning with efficient scene continuation and counterfactual data synthesis. Persistent memory, fully causal conditioning and policy-coupled dynamics remain directions for extending the observation model.

%% file: sections/authors.tex
\clearpage
\section{HelloWorld Team}
\label{app:authors}

\subsection*{Core Contributors}
Fan Lu, Hanshi Wang, Zijing Wang, Quan Feng, Zhi Wang, Shijie Chen, Xianming Zeng

\subsection*{Contributors}
YuJian Zhang, Jiazhe Wang, Xin Zha, Kai Wang, Zhijie Zhao, Lin Zhu, Tianyi Yang, Yucheng Xu, Tao Ji, Haodong Zhang, ZhiPeng Zhang, Peixi Peng, Guang Chen

\subsection*{Project Leader}
Jianyun Xu

\subsection*{Corresponding Authors}
Jianyun Xu, Xingliang Liu, Lei Yang

%% file: refs.bib
@article{hu2023gaia1,
  title={GAIA-1: A Generative World Model for Autonomous Driving},
  author={Hu, Anthony and Russell, Lloyd and Yeo, Hudson and Murez, Zak and Fedoseev, George and Kendall, Alex and Shotton, Jamie and Corrado, Gianluca},
  journal={arXiv preprint arXiv:2309.17080},
  year={2023}
}

@article{russell2025gaia2,
  title={GAIA-2: A Controllable Multi-View Generative World Model for Autonomous Driving},
  author={Russell, Lloyd and Hu, Anthony and Bertoni, Lorenzo and Fedoseev, George and Shotton, Jamie and Arani, Elahe and Corrado, Gianluca},
  journal={arXiv preprint arXiv:2503.20523},
  year={2025}
}

@article{wang2023drivedreamer,
  title={DriveDreamer: Towards Real-World-Driven World Models for Autonomous Driving},
  author={Wang, Xiaofeng and Zhu, Zheng and Huang, Guan and Chen, Xinze and Zhu, Jiagang and Lu, Jiwen},
  journal={arXiv preprint arXiv:2309.09777},
  year={2023}
}

@inproceedings{gao2023magicdrive,
  title={MagicDrive: Street View Generation with Diverse 3D Geometry Control},
  author={Gao, Ruiyuan and Chen, Kai and Xie, Enze and Hong, Lanqing and Li, Zhenguo and Yeung, Dit-Yan and Xu, Qiang},
  booktitle={ICLR},
  year={2024}
}

@article{gao2024magicdrivev2,
  title={MagicDrive-V2: High-Resolution Long Video Generation for Autonomous Driving with Adaptive Control},
  author={Gao, Ruiyuan and Chen, Kai and Xiao, Bo and Hong, Lanqing and Li, Zhenguo and Xu, Qiang},
  journal={arXiv preprint arXiv:2411.13807},
  year={2024}
}

@inproceedings{wen2024panacea,
  title={Panacea: Panoramic and Controllable Video Generation for Autonomous Driving},
  author={Wen, Yuqing and Zhao, Yucheng and Liu, Yingfei and Jia, Fan and Wang, Yanhui and Luo, Chong and Zhang, Chi and Wang, Tiancai and Sun, Xiaoyan and Zhang, Xiangyu},
  booktitle={CVPR},
  year={2024}
}

@inproceedings{wang2024drivewm,
  title={Driving into the Future: Multiview Visual Forecasting and Planning with World Model for Autonomous Driving},
  author={Wang, Yuqi and He, Jiawei and Fan, Lue and Li, Hongxin and Chen, Yuntao and Zhang, Zhaoxiang},
  booktitle={CVPR},
  year={2024}
}

@inproceedings{li2024uniscene,
  title={UniScene: Unified Occupancy-centric Driving Scene Generation},
  author={Li, Bohan and Guo, Jiazhe and Liu, Hongsi and Zou, Yingshuang and Ding, Yikang and Chen, Xiwu and Zhu, Hu and Tan, Feiyang and others},
  booktitle={CVPR},
  year={2025}
}

@article{jiang2024dive,
  title={DiVE: DiT-based Video Generation with Enhanced Control},
  author={Jiang, Junpeng and Hong, Gangyi and Zhou, Lijun and Ma, Enhui and Hu, Hengtong and others},
  journal={arXiv preprint arXiv:2409.01595},
  year={2024}
}

@inproceedings{gao2024vista,
  title={Vista: A Generalizable Driving World Model with High Fidelity and Versatile Controllability},
  author={Gao, Shenyuan and Yang, Jiazhi and Chen, Li and Chitta, Kashyap and Qiu, Yihang and Geiger, Andreas and Zhang, Jun and Li, Hongyang},
  booktitle={NeurIPS},
  year={2024}
}

@inproceedings{yang2024genad,
  title={Generalized Predictive Model for Autonomous Driving},
  author={Yang, Jiazhi and Gao, Shenyuan and Qiu, Yihang and Chen, Li and Li, Tianyu and Dai, Bo and Chitta, Kashyap and Wu, Penghao and Zeng, Jia and Luo, Ping and others},
  booktitle={CVPR},
  year={2024}
}

@article{hu2024drivingworld,
  title={DrivingWorld: Constructing World Model for Autonomous Driving via Video GPT},
  author={Hu, Xiaotao and Yin, Wei and Jia, Mingkai and Deng, Junyuan and Guo, Xiaoyang and Zhang, Qian and Long, Xiaoxiao and Tan, Ping},
  journal={arXiv preprint arXiv:2412.19505},
  year={2024}
}

@misc{horizondrive2026,
  title={HorizonDrive: Self-Corrective Autoregressive World Model for Long-horizon Driving Simulation},
  author={Zhang, Conglang and Zhan, Yifan and Wang, Qingjie and Ouyang, Zhanpeng and Li, Yu and Yang, Zihao and Guo, Xiaoyang and Ren, Weiqiang and Zhang, Qian and Dong, Zhen and others},
  note={Technical report, 2026},
  year={2026}
}

@article{mila2025,
  title={MiLA: Multi-view Intensive-fidelity Long-term Video Generation World Model for Autonomous Driving},
  author={Wang, Haiguang and Liu, Daqi and Xie, Hongwei and Liu, Haisong and Ma, Enhui and Yu, Kaicheng and Wang, Limin and Wang, Bing},
  journal={arXiv preprint arXiv:2503.15875},
  year={2025}
}

@article{wan2025,
  title={Wan: Open and Advanced Large-Scale Video Generative Models},
  author={{Wan Team}},
  journal={arXiv preprint arXiv:2503.20314},
  year={2025}
}

@article{nvidia2025cosmos,
  title={Cosmos World Foundation Model Platform for Physical AI},
  author={{NVIDIA}},
  journal={arXiv preprint arXiv:2501.03575},
  year={2025}
}

@article{nvidia2025cosmostransfer,
  title={Cosmos-Transfer1: Conditional World Generation with Adaptive Multimodal Control},
  author={{NVIDIA}},
  journal={arXiv preprint arXiv:2503.14492},
  year={2025}
}

@misc{genie3,
  title={Genie 3: A New Frontier for World Models},
  author={{Google DeepMind}},
  note={\url{https://deepmind.google/discover/blog/genie-3-a-new-frontier-for-world-models/}},
  year={2025}
}

@misc{hyworld2025,
  title={HunyuanWorld 1.5 (WorldPlay): A Streaming Interactive World Model},
  author={{Tencent Hunyuan}},
  note={Technical report, December 2025},
  year={2025}
}

@article{lingbotworldinfinity2026,
  title={Infinite Worlds with Versatile Interactions},
  author={Gao, Zelin and Wang, Qiuyu and Zhu, Jiapeng and others},
  journal={arXiv preprint arXiv:2607.07534},
  year={2026}
}

@article{lingbotworld2026,
  title={LingBot-World: An Open Real-Time Interactive World Model},
  author={{LingBot Team}},
  journal={arXiv preprint arXiv:2601.20540},
  year={2026}
}

@misc{matrixgame2,
  title={Matrix-Game 2.0: An Open-Source Real-Time Interactive World Model},
  author={{Skywork AI}},
  note={Technical report},
  year={2025}
}

@inproceedings{chen2024diffusionforcing,
  title={Diffusion Forcing: Next-token Prediction Meets Full-Sequence Diffusion},
  author={Chen, Boyuan and Mart{\'\i} Mons{\'o}, Diego and Du, Yilun and Simchowitz, Max and Tedrake, Russ and Sitzmann, Vincent},
  booktitle={NeurIPS},
  year={2024}
}

@article{huang2025selfforcing,
  title={Self Forcing: Bridging the Train-Test Gap in Autoregressive Video Diffusion},
  author={Huang, Xun and Li, Zhengqi and He, Guande and Zhou, Mingyuan and Shechtman, Eli},
  journal={arXiv preprint arXiv:2506.08009},
  year={2025}
}

@inproceedings{song2023consistency,
  title={Consistency Models},
  author={Song, Yang and Dhariwal, Prafulla and Chen, Mark and Sutskever, Ilya},
  booktitle={ICML},
  year={2023}
}

@article{lu2024scm,
  title={Simplifying, Stabilizing and Scaling Continuous-Time Consistency Models},
  author={Lu, Cheng and Song, Yang},
  journal={arXiv preprint arXiv:2410.11081},
  year={2024}
}

@inproceedings{yin2024dmd,
  title={One-step Diffusion with Distribution Matching Distillation},
  author={Yin, Tianwei and Gharbi, Micha{\"e}l and Zhang, Richard and Shechtman, Eli and Durand, Fr{\'e}do and Freeman, William T and Park, Taesung},
  booktitle={CVPR},
  year={2024}
}

@inproceedings{yin2024dmd2,
  title={Improved Distribution Matching Distillation for Fast Image Synthesis},
  author={Yin, Tianwei and Gharbi, Micha{\"e}l and Park, Taesung and Zhang, Richard and Shechtman, Eli and Durand, Fr{\'e}do and Freeman, William T},
  booktitle={NeurIPS},
  year={2024}
}

@inproceedings{zhang2023controlnet,
  title={Adding Conditional Control to Text-to-Image Diffusion Models},
  author={Zhang, Lvmin and Rao, Anyi and Agrawala, Maneesh},
  booktitle={ICCV},
  year={2023}
}

@article{jiang2025vace,
  title={VACE: All-in-One Video Creation and Editing},
  author={Jiang, Zeyinzi and Han, Zhen and Mao, Chaojie and Zhang, Jingfeng and Pan, Yulin and Liu, Yu},
  journal={arXiv preprint arXiv:2503.07598},
  year={2025}
}

@article{yang2023bevcontrol,
  title={BEVControl: Accurately Controlling Street-view Elements with Multi-perspective Consistency via BEV Sketch Layout},
  author={Yang, Kairui and Ma, Enhui and Peng, Jibin and Guo, Qing and Lin, Di and Yu, Kaicheng},
  journal={arXiv preprint arXiv:2308.01661},
  year={2023}
}

@article{swerdlow2024bevgen,
  title={Street-View Image Generation from a Bird's-Eye View Layout},
  author={Swerdlow, Alexander and Xu, Runsheng and Zhou, Bolei},
  journal={IEEE Robotics and Automation Letters},
  year={2024}
}

@inproceedings{caesar2020nuscenes,
  title={nuScenes: A Multimodal Dataset for Autonomous Driving},
  author={Caesar, Holger and Bankiti, Varun and Lang, Alex H and Vora, Sourabh and Liong, Venice Erin and Xu, Qiang and Krishnan, Anush and Pan, Yu and Baldan, Giancarlo and Beijbom, Oscar},
  booktitle={CVPR},
  year={2020}
}

@inproceedings{liu2026l3dr,
  author = {Liu, Quan and Zhang, Xiaoqin and Shao, Ling and Lu, Shijian},
  title = {{L3DR}: 3D-aware {LiDAR} Diffusion and Rectification},
  booktitle = {Proceedings of the IEEE/CVF Conference on Computer Vision and Pattern Recognition (CVPR)},
  year = {2026},
  month = jun,
  pages = {17153--17163},
  url = {https://openaccess.thecvf.com/content/CVPR2026/html/Liu_L3DR_3D-aware_LiDAR_Diffusion_and_Rectification_CVPR_2026_paper.html}
}

@article{zhao2026unidrivedreamer,
  title={UniDriveDreamer: A Single-Stage Multimodal World Model for Autonomous Driving},
  author={Zhao, Guosheng and Wang, Yaozeng and Wang, Xiaofeng and Zhu, Zheng and others},
  journal={arXiv preprint arXiv:2602.02002},
  year={2026},
  url={https://arxiv.org/abs/2602.02002}
}

@article{wang2026sensor2sensor,
  title={Sensor2Sensor: Cross-Embodiment Sensor Conversion for Autonomous Driving},
  author={Wang, Jiahao and Sun, Bo and Bai, Yijing and Casser, Vincent and others},
  journal={arXiv preprint arXiv:2605.22809},
  year={2026},
  url={https://arxiv.org/abs/2605.22809}
}

@inproceedings{fan2017pointset,
  title={A Point Set Generation Network for {3D} Object Reconstruction From a Single Image},
  author={Fan, Haoqiang and Su, Hao and Guibas, Leonidas J.},
  booktitle={Proceedings of the IEEE Conference on Computer Vision and Pattern Recognition},
  pages={605--613},
  year={2017},
  url={https://openaccess.thecvf.com/content_cvpr_2017/html/Fan_A_Point_Set_CVPR_2017_paper.html}
}

@article{xu2025pixelperfect,
  title={Pixel-Perfect Depth with Semantics-Prompted Diffusion Transformers},
  author={Xu, Gangwei and Lin, Haotong and Luo, Hongcheng and Wang, Xianqi and Yao, Jingfeng and Zhu, Lianghui and Pu, Yuechuan and Chi, Cheng and Sun, Haiyang and Wang, Bing and Chen, Guang and Ye, Hangjun and Peng, Sida and Yang, Xin},
  journal={arXiv preprint arXiv:2510.07316},
  year={2025},
  url={https://arxiv.org/abs/2510.07316}
}

@inproceedings{chen2025drivinggpt,
  title={Drivinggpt: Unifying driving world modeling and planning with multi-modal autoregressive transformers},
  author={Chen, Yuntao and Wang, Yuqi and Zhang, Zhaoxiang},
  booktitle={Proceedings of the IEEE/CVF International Conference on Computer Vision},
  pages={26890--26900},
  year={2025}
}

@inproceedings{zhang2025epona,
  title={Epona: Autoregressive diffusion world model for autonomous driving},
  author={Zhang, Kaiwen and Tang, Zhenyu and Hu, Xiaotao and Pan, Xingang and Guo, Xiaoyang and Liu, Yuan and Huang, Jingwei and Yuan, Li and Zhang, Qian and Long, Xiao-Xiao and others},
  booktitle={Proceedings of the IEEE/CVF International Conference on Computer Vision},
  pages={27220--27230},
  year={2025}
}

@article{swerdlow2024street,
  title={Street-view image generation from a bird's-eye view layout},
  author={Swerdlow, Alexander and Xu, Runsheng and Zhou, Bolei},
  journal={IEEE Robotics and Automation Letters},
  volume={9},
  number={4},
  pages={3578--3585},
  year={2024},
  publisher={IEEE}
}

@inproceedings{gao2024magicdrive,
  title={Magicdrive: Street view generation with diverse 3d geometry control},
  author={Gao, Ruiyuan and Chen, Kai and Xie, Enze and Hong, Lanqing and Li, Zhenguo and Yeung, Dit-Yan and Xu, Qiang},
  booktitle={International Conference on Learning Representations},
  volume={2024},
  pages={22841--22860},
  year={2024}
}

@inproceedings{li2025uniscene,
  title={Uniscene: Unified occupancy-centric driving scene generation},
  author={Li, Bohan and Guo, Jiazhe and Liu, Hongsi and Zou, Yingshuang and Ding, Yikang and Chen, Xiwu and Zhu, Hu and Tan, Feiyang and Zhang, Chi and Wang, Tiancai and others},
  booktitle={Proceedings of the computer vision and pattern recognition conference},
  pages={11971--11981},
  year={2025}
}

@inproceedings{guo2025dist,
  title={Dist-4D: Disentangled spatiotemporal diffusion with metric depth for 4d driving scene generation},
  author={Guo, Jiazhe and Ding, Yikang and Chen, Xiwu and Chen, Shuo and Li, Bohan and Zou, Yingshuang and Lyu, Xiaoyang and Tan, Feiyang and Qi, Xiaojuan and Li, Zhiheng and others},
  booktitle={Proceedings of the IEEE/CVF International Conference on Computer Vision},
  pages={27231--27241},
  year={2025}
}

@article{li2025omninwm,
  title={OmniNWM: Omniscient Driving Navigation World Models},
  author={Li, Bohan and Ma, Zhuang and Du, Dalong and Peng, Baorui and Liang, Zhujin and Liu, Zhenqiang and Ma, Chao and Jin, Yueming and Zhao, Hao and Zeng, Wenjun and others},
  journal={arXiv preprint arXiv:2510.18313},
  year={2025}
}

@inproceedings{wang2024driving,
  title={Driving into the future: Multiview visual forecasting and planning with world model for autonomous driving},
  author={Wang, Yuqi and He, Jiawei and Fan, Lue and Li, Hongxin and Chen, Yuntao and Zhang, Zhaoxiang},
  booktitle={Proceedings of the IEEE/CVF Conference on Computer Vision and Pattern Recognition},
  pages={14749--14759},
  year={2024}
}

@article{zhao2025drivedreamer,
  title={DriveDreamer-2: {LLM}-enhanced world models for diverse driving video generation},
  author={Zhao, Guosheng and Wang, Xiaofeng and Zhu, Zheng and Chen, Xinze and Huang, Guan and Bao, Xiaoyi and Wang, Xingang},
  journal={Proceedings of the AAAI Conference on Artificial Intelligence},
  volume={39},
  number={10},
  pages={10412--10420},
  year={2025},
  doi={10.1609/aaai.v39i10.33130}
}

@article{li2026far,
  title={FAR-Drive: Frame-AutoRegressive Video Generation in Closed-Loop Autonomous Driving},
  author={Li, Yaoru and Landi, Federico and Godi, Marco and Jin, Xin and Fu, Ruiju and Ma, Yufei and Sun, Muyang and Si, Heyu and Guo, Qi},
  journal={arXiv preprint arXiv:2603.14938},
  year={2026}
}

@inproceedings{gao2025magicdrive,
  title={MagicDrive-V2: High-resolution long video generation for autonomous driving with adaptive control},
  author={Gao, Ruiyuan and Chen, Kai and Xiao, Bo and Hong, Lanqing and Li, Zhenguo and Xu, Qiang},
  booktitle={Proceedings of the IEEE/CVF International Conference on Computer Vision},
  pages={28135--28144},
  year={2025}
}

@article{zhang2026horizondrive,
  title={HorizonDrive: Self-Corrective Autoregressive World Model for Long-horizon Driving Simulation},
  author={Zhang, Conglang and Zhan, Yifan and Wang, Qingjie and Ouyang, Zhanpeng and Li, Yu and Yang, Zihao and Guo, Xiaoyang and Ren, Weiqiang and Zhang, Qian and Dong, Zhen and others},
  journal={arXiv preprint arXiv:2605.11596},
  year={2026}
}
